# Machine Learning and System Identification for Estimation in Physical Systems

Fredrik Bagge Carlson

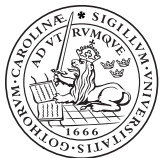

Department of Automatic Control

*To Farah and the next step*

# Abstract

In this thesis, we draw inspiration from both classical system identification and modern machine learning in order to solve estimation problems for real-world, physical systems. The main approach to estimation and learning adopted is optimization based. Concepts such as regularization will be utilized for encoding of prior knowledge and basis-function expansions will be used to add nonlinear modeling power while keeping data requirements practical.

The thesis covers a wide range of applications, many inspired by applications within robotics, but also extending outside this already wide field. Usage of the proposed methods and algorithms are in many cases illustrated in the real-world applications that motivated the research. Topics covered include dynamics modeling and estimation, model-based reinforcement learning, spectral estimation, friction modeling and state estimation and calibration in robotic machining.

In the work on modeling and identification of dynamics, we develop regularization strategies that allow us to incorporate prior domain knowledge into flexible, overparameterized models. We make use of classical control theory to gain insight into training and regularization while using flexible tools from modern deep learning. A particular focus of the work is to allow use of modern methods in scenarios where gathering data is associated with a high cost.

In the robotics-inspired parts of the thesis, we develop methods that are practically motivated and ensure that they are implementable also outside the research setting. We demonstrate this by performing experiments in realistic settings and providing open-source implementations of all proposed methods and algorithms.

# Acknowledgements

I would like to acknowledge the influence of my PhD thesis supervisor Prof. Rolf Johansson and my Master's thesis advisor Dr. Vuong Ngoc Dung at SIMTech, who both encouraged me to pursue the PhD degree, for which I am very thankful. Prof. Johansson has continuously supported my ideas and let me define my work with great freedom, thank you.

My thesis co-supervisor, Prof. Anders Robertsson, thank you for your never-ending enthusiasm, source of good mood and encouragement. When working 100% overtime during hot July nights in the robot lab, it helps to know that one is never alone.

I would further like to direct my appreciation to friends and colleagues at the department. It has often fascinated me, how a passionate and excited speaker can make a boring topic appear interesting. No wonder a group of 50+ highly motivated and passionate individuals can make an already interesting subject fantastic. In particular Prof. Bo Bernhardsson, my office mates Gautham Nayak Seetanadi and Mattias Fält and my travel mates Martin Karlsson, Olof Troeng and Richard Pates, you have all made the last 5 years outside and at the department particularly enjoyable.

Credit also goes to Jacob Wikmark, Dr. Björn Olofsson and Dr. Martin Karlsson for incredibly generous and careful proof reading of the manuscript to this thesis, and to Leif Andersson for helping out with typesetting, you have all been very helpful!

Finally, I would like to thank my family in Vinslöv who have provided and continue to provide a solid foundation to build upon, to my family from Sparta who provided a second home and a source of both comfort and adventure, and to the welcoming new addition to my family in the Middle East.

**Financial support**

Parts of the presented research were supported by the European Commission under the 7th Framework Programme under grant agreement 606156 Flexifab. Parts of the presented research were supported by the European Commission under the Framework Programme Horizon 2020 under grant agreement 644938 SARAFun. The author is a member of the LCCC Linnaeus Center and the ELLIIT Excellence Center at Lund University.

# Contents

# 1

# Introduction

Technical computing, sensing and control are well-established fields, still making steady progress today. Rapid advancements in the ability to train flexible machine learning models, enabled by amassing data and breakthroughs in the understanding of the difficulties behind gradient-based training of deep architectures, have made the considerably younger field of machine learning explode with interest. Together, they have made automation feasible in situations we previously could not dream of.

The vast majority of applications within machine learning are, thus far, in domains where data is plentiful, such as image classification and text analysis. Flexible machine-learning models thrive on large datasets, and much of the advancements of deep learning is often attributed to growing datasets, rather than algorithmic advancements [Goodfellow et al., 2016]. In practice, it took a few breakthrough ideas to enable training of these deep and flexible architectures, but few argue with the fact that the size of the dataset is of great importance. In many domains, notably domains involving mechanical systems such as robots and cars, gathering the data required to make use of a modern machine-learning model often proves difficult. While a simple online search returns thousands of pictures of a particular object, and millions of Wikipedia articles are downloaded in seconds, collecting a single example of a robot task requires actually operating a robot, in real time. Not only is this associated with a tremendous overhead, but the data collected during this experiment using a particular policy or controller is also not always informative of the system and its behavior when it has gone through training. This has seemingly made the progress of machine learning in control of physical systems lag behind, and traditional methods are still dominating today. Design methods based on control theory have long served us well. Complex problems are broken down into subproblems which are easily solved. The complexity arising when connecting these subsystems together is handled by making the design of each subsystem robust to uncertainties in its inputs [Åström and Murray, 2010]. While this has been a very successful strategy, it leaves us with a number of questions. Are we leaving performance on the table by connecting individually designed systems together instead of optimizing the complete system? Are we wasting effort designing subsystems using time-consuming, traditional methods, when larger, more complex subsystems could be designed automatically using data-based methods?

In this thesis, we will draw inspiration from both classical system identification and machine learning. The combination is potentially powerful, where system identification's deep roots in physics and domain knowledge allow us to use flexible machine-learning methods in applications where the data alone is insufficient. The motivation for the presented research mainly comes from the projects Flexifab and SARAFun. The Flexifab project investigated the use of industrial robots for friction stir welding, whereas the SARAFun project considered robotic assembly. While these two projects may seem dissimilar, and indeed they are, they have both presented research problems within estimation in physical systems. The thesis is divided into three parts, not related to the project behind the research, but rather based on the types of problems considered. In Part I, we consider modeling, learning and identification problems. Many of these problems were encountered in a robotics context but result in generic methods that extend outside the field of robotics. We also illustrate the use of some of the developed methods in reinforcement learning and trajectory optimization. In Part II, we consider problems motivated by the friction-stir-welding (FSW) process. FSW is briefly introduced, whereafter we consider a number of calibration problems, arising in the FSW context, but finding application also outside of FSW [Bao et al., 2017; Chalus and Liska, 2018; Yongsheng et al., 2017]. We also discuss state estimation in the FSW context, a problem extending to general machining with industrial manipulators.

The outline of the thesis is given in Chap. 2 and visualized graphically in Fig. 1.1.

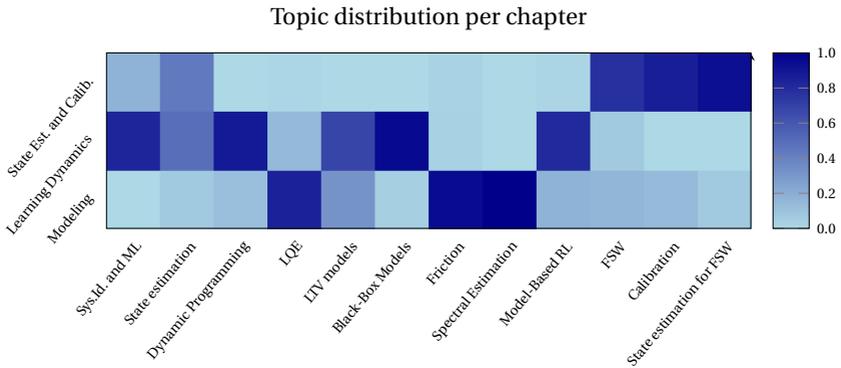

**Figure 1.1** This thesis can be divided into three main topics. This figure indicates the topic distribution for each chapter, where a dark blue color indicates a strong presence of a topic. The topic distribution was automatically found using latent Dirichlet allocation (LDA) [Murphy, 2012].

## 1.1  Notation

Notation frequently used in the thesis is summarized in Table 1.1. Many methods developed in the thesis are applied within robotics and we frequently reference different coordinate frames. The tool-flange frame $\mathcal{TF}$ is attached to the tool flange of a robot, the mechanical interface between the robot and the payload or tool. The robot base frame $\mathcal{RB}$ is the base of the forward-kinematics function of a manipulator, but could also be, e.g., the frame of an external optical tracking system that measures the location of the tool frame in the case of a flying robot etc. A sensor delivers measurements in the sensor frame $\mathcal{S}$. The joint coordinates, e.g., joint angles for a serial manipulator, are denoted $q$. The vector of robot joint torques is denoted $\tau$, and external forces and torques acting on the robot are gathered in the wrench $\mathfrak{f}$. The Jacobian of a function is denoted $J$ and the Jacobian of a manipulator is denoted $J(q)$. We use $k$ to denote a vector of parameters to be estimated except in the case of deep networks, which we parameterize by the weights $w$. The gradient of a function $f$ with respect to $x$ is denoted $\nabla_x f$. We use $x_t$ to denote the state vector at time $t$ in Markov systems, but frequently omit this time index and use $x^+$ to denote $x_{t+1}$ in equations where all other variables are given at time $t$. The matrix $\langle s \rangle \in so$ is formed by the elements of a vector $s$ and has the skew-symmetric property $\langle s \rangle + \langle s \rangle^\top = 0$ [Murray et al., 1994].

**Table 1.1**  Definition and description of coordinate frames, variables and notation.

| | | |
|---|---|---|
| $\mathcal{RB}$ | | Robot base frame. |
| $\mathcal{TF}$ | | Tool-flange frame, attached to the TCP. |
| $\mathcal{S}$ | | Sensor frame. |
| $q$ | $\in \mathbb{R}^n$ | Joint Coordinate |
| $\dot{q}$ | $\in \mathbb{R}^n$ | Joint velocity |
| $\tau$ | $\in \mathbb{R}^n$ | Torque vector |
| $\mathfrak{f}$ | $\in \mathbb{R}^6$ | External force/torque wrench |
| $R_A^B$ | $\in SO(3)$ | Rotation matrix from $\mathcal{B}$ to $\mathcal{A}$ |
| $T_A^B$ | $\in SE(3)$ | Transformation matrix from $\mathcal{B}$ to $\mathcal{A}$ |
| $F_k(q)$ | $\in SE(3)$ | Robot forward kinematics at pos. $q$ |
| $J(q)$ | $\in \mathbb{R}^{6 \times n}$ | Manipulator Jacobian at pos. $q$ |
| $\langle s \rangle$ | $\in so(3)$ | Skew-symmetric matrix with parameters $s \in \mathbb{R}^3$ |
| $x^+ = f(x, u)$ | $\mathbb{R}^n \times \mathbb{R}^m \mapsto \mathbb{R}^n$ | Dynamics model |
| $x$ | | State variable |
| $u$ | | Input/control signal |
| $x^+$ | | $x$ at the next sample instant |
| $k$ | | Parameter vector |
| $\nabla_x f$ | | Gradient of $f$ with respect to $x$ |
| $\hat{x}$ | | Estimate of variable $x$ |
| $x_{i:j}$ | | Elements $i, i+1, ..., j$ of $x$ |

# 2

# Publications and Contributions

The contributions of this thesis and its author, as well as a list of the papers this thesis is based on, are detailed below.

## Included publications

This thesis is based on the following publications:

Bagge Carlson, F., A. Robertsson, and R. Johansson (2015a). "Modeling and identification of position and temperature dependent friction phenomena without temperature sensing". In: *Int. Conf. Intelligent Robots and Systems (IROS), Hamburg*. IEEE.

Bagge Carlson, F., R. Johansson, and A. Robertsson (2015b). "Six DOF eye-to-hand calibration from 2D measurements using planar constraints". In: *Int. Conf. Intelligent Robots and Systems (IROS), Hamburg*. IEEE.

Bagge Carlson, F., A. Robertsson, and R. Johansson (2017). "Linear parameter-varying spectral decomposition". In: *2017 American Control Conf (ACC), Seattle*.

Bagge Carlson, F., A. Robertsson, and R. Johansson (2018a). "Identification of LTV dynamical models with smooth or discontinuous time evolution by means of convex optimization". In: *IEEE Int. Conf. Control and Automation (ICCA), Anchorage, AK*.

Bagge Carlson, F., R. Johansson, and A. Robertsson (2018b). "Tangent-space regularization for neural-network models of dynamical systems". *arXiv preprint arXiv:1806.09919*.

In the publications listed above, F. Bagge Carlson developed manuscripts, models, identification procedures, implementations and performed experiments. A. Robertsson and R. Johansson assisted in improving the manuscripts.

Bagge Carlson, F., M. Karlsson, A. Robertsson, and R. Johansson (2016). "Particle filter framework for 6D seam tracking under large external forces using 2D laser sensors". In: *Int. Conf. Intelligent Robots and Systems (IROS), Daejeong, South Korea*.

In this publication, F. Bagge Carlson contributed with a majority of the implementation and structure of the state estimator and manuscript. M. Karlsson assisted in parts of the implementation and contributed with ideas on the structure of the state estimator, as well as assistance in preparing the manuscript. A. Robertsson and R. Johansson assisted in improving the manuscript.

Parts of the work presented in this thesis have previously been published in the Licentiate Thesis by the author

Bagge Carlson, F. (2017). *Modeling and Estimation Topics in Robotics*. Licentiate Thesis TFRT-3272. Dept. Automatic Control, Lund University, Sweden.

## **Other publications**

The following papers, authored or co-authored by the author of this thesis, cover related topics in robotics but are not included in this thesis:

Bagge Carlson, F., N. D. Vuong, and R. Johansson (2014). "Polynomial reconstruction of 3D sampled curves using auxiliary surface data". In: *2014 IEEE Int. Conf. Robotics and Automation (ICRA) Hong-Kong*.

Stolt, A., F. Bagge Carlson, M. M. Ghazaei Ardakani, I. Lundberg, A. Robertsson, and R. Johansson (2015). "Sensorless friction-compensated passive lead-through programming for industrial robots". In: *Int. Conf. Intelligent Robots and Systems (IROS), Hamburg*.

Karlsson, M., F. Bagge Carlson, J. De Backer, M. Holmstrand, A. Robertsson, and R. Johansson (2016). "Robotic seam tracking for friction stir welding under large contact forces". In: *7th Swedish Production Symposium (SPS), Lund*.

Karlsson, M., F. Bagge Carlson, J. De Backer, M. Holmstrand, A. Robertsson, R. Johansson, L. Quintino, and E. Assuncao (2019). "Robotic friction stir welding, challenges and solutions". *Welding in the World, The Int. Journal of Materials Joining*. ISSN: 0043-2288. Submitted.

Karlsson, M., F. Bagge Carlson, A. Robertsson, and R. Johansson (2017). "Two-degree-of-freedom control for trajectory tracking and perturbation recovery during execution of dynamical movement primitives". In: *20th IFAC World Congress, Toulouse*.

Bagge Carlson, F. and M. Haage (2017). *YuMi low-level motion guidance using the Julia programming language and Externally Guided Motion Research Interface*. Technical report TFRT-7651. Department of Automatic Control, Lund University, Sweden.

## Outline and Contributions

The following is an outline of the contents and contributions of subsequent chapters.

Chapters 3 to 6 serve as an introduction and the only contribution is the organization of the material. An attempt at highlighting interesting connections between control theory, system identification and machine learning is made, illustrating similarities between the fields. Methods from the literature serving as background and inspiration for the contributions outlined in subsequent chapters are introduced here.

Chapter 7 is based on "Identification of LTV Dynamical Models with Smooth or Discontinuous Time Evolution by means of Convex Optimization" and presents a framework for identification of Linear Time-Varying models. The contributions made in the chapter include

- Organization of identification methods into a common framework.

- Development of efficient algorithms for solving a set of optimization problems based on dynamic programming.

- Proof of well-posedness for a set of optimization problems.

- Modification of a standard dynamic-programming algorithm to allow inclusion of prior information.

Usage of the proposed methods is demonstrated in numerical examples and an open-source framework implementing the methods is made available. Methods developed in this chapter are further used in Chap. 11.

Chapter 8 is based on "Tangent-Space Regularization for Neural-Network Models of Dynamical Systems" and treats identification of dynamics models using methods from deep learning. The chapter provides an analysis of how standard deep-learning regularization affects the learning of dynamical systems and a new regularization approach is proposed and shown to introduce less bias compared to traditional regularization. Structural choices in the deep-learning model are further viewed from a dynamical-systems perspective and the effects of these choices are analyzed from an optimization perspective. The discussion is supported by extensive numerical evaluation.

Chapter 9 is based on "Modeling and identification of position and temperature dependent friction phenomena without temperature sensing" and introduces two new friction models. It is shown how, for some industrial manipulators, the joint friction varies with the joint angle. A model and identification procedure for this angle-dependent friction is introduced and verified experimentally to reduce errors in friction modeling.

Chapter 9 further introduces a friction model that makes use of estimated power losses due to friction. Power losses are known to increase the joint temperature and in turn, influence friction. The main benefit of the model is the offline identification and open-loop application, eliminating the need for adaptation of friction parameters during operation. Also this model is verified experimentally as well as in simulations.

The work in Chap. 10 was motivated by observations gathered during the work presented in Chap. 9, where residuals from friction modeling indicated the presence of a highly periodic disturbance. Analysis of this disturbance, which turned out to be modulated by the velocity of the joint, led to the development of a new spectral estimation method, the main topic and contribution of this chapter. The method decomposes the spectrum of a signal along an auxiliary dimension and allows for the estimation of a functional dependence between the auxiliary variable and the Fourier coefficients of the signal under analysis. The method was demonstrated on a simulated signal as well as applied to the residual signal from Chap. 9. The chapter also includes a statistical proof of consistency of the proposed method.

In Chap. 11, usage of the methods developed in Chapters 7 and 8 is illustrated in an application of model-based reinforcement learning, parts of which were originally introduced in "Identification of LTV Dynamical Models with Smooth or Discontinuous Time Evolution by means of Convex Optimization". It is shown how the regularized methods presented in Chap. 7 allow solving a model-based trajectory-optimization problem without any prior model of the system. It is further shown how incorporating the deep-learning models of Chap. 8 using the modified dynamic-programming solver presented in Chap. 7 can accelerate the learning procedure by accumulating experience between experiments. The combination of dynamics model and learning algorithm was shown to result in a highly data-efficient reinforcement-learning algorithm.

Chapter 12 introduces the friction-stir-welding (FSW) process that served as motivation for the conducted research. Chapter 13 introduces algorithms to calibrate force sensors and laser sensors that make use of easily gathered data, important for practical application of the methods.

The algorithm developed for calibration of force/torque sensors solves a convex relaxation of an optimization problem, and it is shown how the optimal solution to the originally constrained problem is obtained by a projection onto the constraint set. The main benefit of the proposed algorithm is its numerical robustness and the lack of requirement for special calibration equipment.

The algorithm proposed for calibration of laser sensors, originally presented in "Six DOF eye-to-hand calibration from 2D measurements using planar constraints", was motivated by the FSW process and finds the transformation matrix between the coordinate systems of the sensor and the tool. This method eliminates the need for special-purpose equipment in the calibration procedure and was shown to be robust to errors in the required initial guess. Use of the algorithm was demonstrated in both simulations and using a real sensor.

Chapter 14 is based on "Particle Filter Framework for 6D Seam Tracking Under Large External Forces Using 2D Laser Sensors" and builds upon the work from Chap. 13. In this chapter, a state estimator capable of incorporating the sensing modalities described in Chap. 13 is introduced. The main contribution is an integrated framework for state estimation in the FSW context, with discussions about, and proposed solutions to, many unique problems arising in the FSW context. The chapter also outlines an open-source software framework for simulation of the state-estimation procedure, intended to guide the user in application of the method and assembly of the hardware sensing.

The thesis is concluded in Sec. 14.5 with a brief discussion around directions for future work.

**Software**

The research presented in this thesis is accompanied by open-source software implementing all proposed methods and allowing reproduction of simulation results. A summary of the released software is given below.

**[*Robotlib.jl,* B.C., 2015]**  Robot kinematics, dynamics and calibration. Implements [Bagge Carlson et al., 2015b; Bagge Carlson et al., 2015a].

**[*Robotlab.jl,* B.C. et al., 2017** ] Real-time robot controllers in Julia. Connections to ABB robots [Bagge Carlson and Haage, 2017].

**[*LPVSpectral.jl,* B.C., 2016]**  (Sparse and LPV) Spectral estimation methods, implements [Bagge Carlson et al., 2017].

**[*PFSeamTracking.jl,* B.C. et al., 2016]**  Seam tracking and simulation [Bagge Carlson et al., 2016].

**[*LowLevelParticleFilters.jl,* B.C., 2018]**  General state estimation and parameter estimation for dynamical systems.

**[*BasisFunctionExpansions.jl,* B.C., 2016]**  Tools for estimation and use of basis-function expansions.

**[*DifferentialDynamicProgramming.jl,* B.C., 2016]**  Optimal control and model-based reinforcement learning.

**[*DynamicMovementPrimitives.jl,* B.C. et al., 2016]**  DMPs in Julia, implements [Karlsson et al., 2017].

**[*LTVModels.jl,* B.C., 2017]**  Implements all methods in [Bagge Carlson et al., 2018b].

**[*JacProp.jl,* B.C., 2018]**  Implements all methods in [Bagge Carlson et al., 2018a].

**Part I**

# Model Estimation

# 3

# Introduction—System Identification and Machine Learning

Estimation, identification and learning are three words often used to describe similar notions. Different fields have traditionally preferred one or the other, but no matter what term has been used, the concepts involved have been similar, and the end goals have been the same. The machine learning community talks about model *learning*. The act of observing data generated by a system and building a model that can either predict the output given an unseen input, or generate new data from the same distribution as the observed data was generated from [Bishop, 2006; Murphy, 2012; Goodfellow et al., 2016]. The control community, on the other hand, talks about *system identification*, the act of perturbing a system using a controlled input, observing the response of the system and estimating/identifying a model that agrees with the observations [Ljung, 1987; Johansson, 1993]. Although terminology, application and sometimes also methods have differed, both fields are concerned with building models that capture structure observed in data.

This thesis will use the terms more or less interchangeably and they will always refer to solving an optimization problem. The function we optimize is specifically constructed to encode how well the model agrees with the observations, or rather, the degree of *mismatch* between the model predictions and the data. Optimization of a *cost function* is a very common and the perhaps dominating strategy in the field, but approaches such as Bayesian inference offer an alternative strategy, focusing on statistical models. Bayesian methods offer interesting and often valuable insight into the complete posterior distribution of the model parameters after having observed the data [Bishop, 2006; Murphy, 2012]. This comes at the cost of computational complexity. Bayesian methods often involve intractable high-dimensional integration, necessitating approximate solution methods such as Monte Carlo methods. Variational inference is another popular approximate solution method that transform the Bayesian inference problem to an optimization problem over a parameterized probability density [Bishop, 2006; Murphy, 2012].

No matter what learning paradigm one chooses to employ, a model structure must be chosen before any learning or identification can begin. The choice of model is not always trivial and must be guided by application-specific goals. Are we estimating a model to learn something about the system or to predict future output of the system? Do we want to use the model for simulation or control synthesis?

*Linear models* offer a strong baseline, they are easy to fit and provide excellent interpretability. While few systems are truly linear, many systems are described well *locally* by a linear model [Åström and Murray, 2010; Glad and Ljung, 2014]. A system actively regulated to stay around an operating point is, for instance, often well described by a linear model. Linear models further facilitate easy control design thanks to the very well-developed theory for linear control system analysis and synthesis.

When a linear model is inadequate, we might consider first principles and specify a *gray-box model*, a model with well motivated structure but unknown parameters [Johansson, 1993]. The parameters are then chosen so as to agree with observed data. Specification of a gray-box model requires insight into the physics of the system. Complicated systems might defy our efforts to write down simple governing equations, making gray-box modeling hard. However, when we are able to use them, we are often rewarded with further insight into the system provided by the identification of the model parameters.

A third modeling approach is what is often referred to as *black-box modeling*. We refer to the model as a black box since it offers little or no insight into how the system is actually working. It does, however, offer potentially unlimited modeling flexibility, the ability to fit any data-generating system [Sjöberg et al., 1995]. The structure of the black-box model is chosen so as to promote both flexibility, but also ease of learning. Alongside giving up interpretability[1] of the resulting model, the fitting of black-box models is associated with the risk of overfitting—a failure to capture the true governing mechanisms of the system [Murphy, 2012]. An overfit model agrees very well with the data used for training, but fails to generalize to novel data. A common explanation for the phenomenon is the flexible model being deceived by noise present in the data. Combatting overfitting has been a major research topic for a long time and remains so today. Oftentimes, *regularization*—a restriction of flexibility—is employed, a concept this thesis will explore in detail and make great use of.

## 3.1 Models of Dynamical Systems

For control design and analysis, Linear Time-Invariant (LTI) models have been hugely important, mainly motivated by their simplicity and the fact that both performance and robustness properties are well understood. The identification of linear models shares these properties in many regards, and has been a staple of system identification since the early beginning [Ljung, 1987]. Not only are the theory and properties of linear identification well understood, the computational complexity of many of the linear identification algorithms is also favorable.

---

[1] Interpretable machine learning is an emerging field trying to provide insight into the workings of black-box models.

Methods that have been made available by decades of progression of Moore's law are, however, often underappreciated among system identification practitioners. With the computational power available today, one can solve large optimization problems and high dimensional integrals, leading to the emergence of the fields of deep learning [Goodfellow et al., 2016], large-scale convex optimization [Boyd and Vandenberghe, 2004] and Bayesian nonparametrics [Hjort et al., 2010; Gershman and Blei, 2011]. In this thesis, we hope to contribute to bridging some of the gap between the system-identification literature and modern machine learning. We believe that the interchange will be bidirectional, because even though new powerful methods have been developed in the learning communities, classical system identification has both useful domain knowledge and a strong systems-theoretical background, with well developed concepts such as stability, identifiability and input design, that are seldom talked about in the learning community.

**Prediction error vs. simulation error**

Common for all models linear in the parameters, is that paired with a quadratic cost function, the solution to the prediction error problem is available on closed form [Johansson, 1993]. Linear time-invariant (LTI) dynamic models on the form (3.1) are no exceptions and they can indeed be estimated from data by solving the normal equations, provided that the full state is measured.

$$x_{t+1} = Ax_t + Bu_t + v_t$$
$$y_t = x_t + e_t \tag{3.1}$$

In (3.1), $x \in \mathbb{R}^n, \quad y \in \mathbb{R}^p$ and $u \in \mathbb{R}^m$ are the state, measurement and input respectively.

The name prediction error method (PEM) refers to the minimization of the prediction errors

$$x^+ - \hat{x}^+ = v$$
$$\hat{x}^+ = Ax + Bu \tag{3.2}$$

and PEM constitutes the optimal method if all errors are *equation errors* [Ljung, 1987], i.e., $e = 0$. If we instead adopt the model $v = 0$, we arrive at the *output-error* or *simulation-error* method [Ljung, 1987; Sjöberg et al., 1995], where we minimize

$$y^+ - \hat{x}^+ = e$$
$$\hat{x}^+ = A\hat{x} + Bu \tag{3.3}$$

The difference between (3.2) and (3.3) may seem subtle, but has big consequences. In (3.3), no measurements of $x$ are ever used to form the predictions $\hat{x}$. Instead, the model is applied recursively with previous predictions as inputs. In (3.2), however, a measurement of $x$ is used as input to the model to form the prediction $\hat{x}^+$. While the prediction error can be minimized with standard LS, output error minimization is a highly nonlinear problem that requires additional care. Sophisticated methods based on matrix factorizations exist for solving the OE problem for linear models [Verhaegen and Dewilde, 1992], but in general, the problem is hard. The difficulty stems from the recursive application of the model parameters, introducing the risk for exploding/vanishing gradients and nonconvex loss surfaces. The system-identification literature is full of methods to mitigate these issues, the more common of which include multiple shooting and collocation [Stoer and Bulirsch, 2013].

Optimization of the simulation-error metric leads to long-standing challenges that have resurfaced recently in the era of deep learning [Goodfellow et al., 2016]. The notion of backpropagation through time for training of modern recurrent neural networks and all its associated computational challenges are very much the same challenges as those related to solving the simulation-error problem. When simulating the system more than one step forward in time, the state sequence becomes a product of both parameters and previous states, which in turn are functions of the parameters. While an LTI model is linear in the parameters, the resulting optimization problem is not. Both classical and recent research have made strides towards mitigating some of these issues [Hochreiter and Schmidhuber, 1997; Stoer and Bulirsch, 2013; Xu et al., 2015], but the fundamental problem remains [Pascanu et al., 2013]. One of the classical approaches, multiple shooting [Stoer and Bulirsch, 2013], successfully mitigates the problem with a deep computational graph by breaking it up and introducing constraints on the boundary conditions between the breakpoints. While methods like multiple shooting work well also for training of recurrent neural networks, they are seldom used, and the deep-learning community has invented its own solutions [Pascanu et al., 2013].

## 3.2 Stability

An important theoretical aspect of dynamical systems is the notion of stability [Khalil, 1996; Åström and Murray, 2010]. Loosely speaking, a stable system is one where neither the output nor the internal state of the system goes to infinity unless we supply an infinite input. When estimating a model for a system known to be stable, one would ideally like to obtain a stable model. Some notions of stability imply the convergence of system trajectories to, e.g., an equilibrium point or a limit cycle. The effect of perturbations, noise or small model errors will for a stable model have an eventually vanishing effect. For unstable systems and models, small perturbations in initial conditions or perturbations to the trajectory can have an unbounded effect. For simulation, obtaining a stable model of a stable system is thus important. Many model sets, including the set of LTI models of a particular dimension, include unstable models. If the feasible set contains unstable models, the search for the model that best agrees with the data is not guaranteed to return a stable model. One can imagine many ways of dealing with this issue. A conceptually simple way is to search only among stable models. This strategy is in general hard, but successful approaches include [Manchester et al., 2012]. Model classes that include only stable models may unfortunately be restrictive and limit the use of intuition in choosing a model architecture. Another strategy is to project the found model onto a subset of stable models, provided that such a projection is available. There is, however, no guarantee that the projection is the optimal model in the set of stable models. A hybrid approach is to, in each iteration of an optimization problem, project the model onto the set of stable models, a technique that in general gradient-based optimization is referred to as projected gradient descent [Goldstein, 1964]. The hope with such a strategy is that the optimization procedure will stay close to the desired target set and thus seek out favorable points within this set, whereas projection of only the final solution might allow the optimization procedure to stray far away from good solutions within the desired target set. A closely related approach will be used in Chap. 13, where the optimization variable is a rotation matrix in *SO*(3), a space which is easy to project onto but harder to optimize over directly.

The set of stable discrete-time LTI models is easy to describe; as long as the *A* matrix in (3.1) has eigenvalues no greater than 1, the model is stable [Åström and Murray, 2010; Glad and Ljung, 2014]. If the eigenvalues are strictly less than one, the model is exponentially stable and all energy contained within the system will eventually decay to zero. For nonlinear models, characterizing the set of stable models is in general much harder. One way of proving that a nonlinear system is stable is to find a Lyapunov function. Systematic ways of finding such a function are unfortunately lacking.

## 3.3   Inductive Bias and Prior Knowledge

In the control literature, it is well known that a continuous-time linear system with long time constants correspond to small eigenvalues of the dynamics matrix, or eigenvalues close to 1 in the discrete-time case. The success of the LSTM (Long Short-Term Memory), a form of recurrent neural network [Hochreiter and Schmidhuber, 1997], in learning long time dependencies seem natural in this light. The LSTM is essentially introducing an inductive bias towards models with long time constants.

In fact, many success stories in the deep-learning field can be traced back to the invention of a model architecture with appropriate inductive bias for a specific task. The perhaps most prominent example of this is the success of convolutional neural networks (CNN) for computer vision tasks. Ulyanov et al. (2017) showed that a CNN can be used remarkably successfully for computer vision tasks such as de-noising and image in-painting *completely without pre-training*. The CNN architecture simply learns to fit the desirable structure in a *single* natural image much faster and better than it fits, say, random noise. Given enough training epochs, the complex neural network manages to fit also the noise, showing that the capacity is there. The inductive bias, however, is clearly more towards natural images.

Closely related to inductive bias are the concepts of statistical priors and regularization, both of which are explicit attempts at endowing the model with inductive bias [Murphy, 2012]. The concept of using regularization to encode prior knowledge will be used extensively in the thesis.

A different approach to encoding prior knowledge is intelligent initialization of overparameterized models. It is well known that the gradient descent algorithm converges to the minimum-norm solution for overparametereized convex problems if initialized near zero [Wilson et al., 2017]. This can be seen as an implicit bias or regularization, encoded by the initialization. Similarly, known time constants can be encoded by initialization of matrices in recurrent mappings with well chosen eigenvalues, or as differentiation chains etc. This topics will not be discussed much further in this thesis, but may be worthwhile considering during modeling and training.

Can the problem of estimating models for dynamical control system be reduced to that of finding an architecture with the appropriate inductive bias? We argue that it is at least beneficial to have the model architecture working with us rather than against us. The question then becomes: How can we construct our models such that they *want* to learn good dynamics models? Decades of research in classical control theory and system identification hopefully become useful in answering these questions. We hope that the classical control perspective and the modern machine learning perspective come together in this thesis, helping us finding good models for dynamical systems.

# 4

# State Estimation

The state of a system is a collection of information that summarizes everything one needs to know in addition to the model in order to predict the future state of the system. As an example, consider a point mass—a suitable state-representation for this system is its position and velocity. A dynamics model for the movement of the point mass might be a double integrator with acceleration as input. We refer to the function of the dynamics model that evolves the state in time as the *state-transition function*.

The notion of state is well developed in control. Recurrent neural networks introduce and learn their own state representation, similar to how subspace-based identification [Van Overschee and De Moor, 1995] can identify both a state representation and a model for linear systems. LSTMs [Hochreiter and Schmidhuber, 1997] were introduced to mitigate vanishing/exploding gradients and to allow the model to learn a state representation that remembers information on longer time-scales. Unfortunately, also LSTMs forget; to mitigate this, the *attention mechanism* was introduced by the deep learning community [Xu et al., 2015]. The attention vector is essentially containing the entire input history, but the use of it is gated by a nonlinear, learned, model. Attention as used in the literature is a sequence-to-sequence model, often in a smoothing fashion, where the input is encoded both forwards and backwards. Use of smoothing is feasible for reasoning about a system on an episode basis, but not for prediction.

While mechanisms such as attention [Xu et al., 2015] have been very successful in tasks such as natural language translation, the classical notion of state provides a terser representation of the information content that can give insight into the modeled system. Given a state-space model of the system, state estimation refers to the act of identifying the sequence of states that best agrees with both the specified model and with observations made of the system. Observations might come at equidistant or nonequidistant points in time and consist of parts of the state, the whole state or, in general, a function of the state. We refer to this function as an *observation model*.

## 4.1 General State Estimation

The state-estimation problem is conceptually simple; solve an optimization problem for the state-sequence that minimizes residuals between model predictions and observations. How the size of the residuals is measured is often determined by either practical considerations or statistical assumptions on the noise acting on the system and the observations. The complexity of this straightforward approach naturally grows with the length of the data collected.[1] Potential mitigations include moving-horizon estimation [Rawlings and Mayne, 2009], where the optimization problem is solved for a fixed-length data record which is updated at each time step.

It is often desirable to estimate not only the most likely state sequence, but also the uncertainty in the estimate. Given a generative model of the data, one can estimate the full posterior density of the state sequence after having seen the data. Full posterior density estimation is a powerful concept, but exact calculation is unfortunately only tractable in a very restricted setting, namely the linear-Gaussian case. In this case, the optimal estimate is given exactly by the Kalman filter [Åström, 2012], which we will touch upon in Sec. 4.3. Outside the linear and Gaussian world, one is left with approximate solution strategies, one particularly successful one being the particle filter.

## 4.2 The Particle Filter

The particle filter is a sequential Monte-Carlo method for approximate solution of high dimensional integrals with a sequential structure [Gustafsson, 2010]. We will not develop much of the theory of particle filters here, but will instead give a brief intuitive introduction.

We begin by associating a statistical model with the state-transition function $x^+ \sim p(x^+|x)$. One example is $x^+ = Ax + v$, where $v \sim \mathcal{N}(0, 1)$. At time $t = 0$, we may summarize our belief of the state in some distribution $p(x_0)$. At the next time instance $t = 1$, the distribution of the state will then be given by

$$p(x_1) = \int p(x_1, x_0) \, dx_0 = \int p(x_1|x_0) p_0(x_0) \, dx_0 \qquad (4.1)$$

Unfortunately, very few pairs of distributions $p(x^+|x)$ and $p_0$ will lead to a tractable integral in (4.1) and a distribution $p(x^+)$ that we can represent on closed form. The particle filter therefore approximates $p_0$ with a collection of samples or *particles* $\{\hat{x}^i\}_{i=1}^N$, where each particle can be seen as a distinct hypothesis of the correct state. Particles are easily propagated through $p(x^+|x)$ to obtain a new collection at time $t = 1$, forming a sampled representation of $p(x_1)$.

When a measurement $y$ becomes available, we associate each particle with a weight given by the likelihood of the measurement given the particle state and the observation model $p(y|x)$. Particles that represent state hypotheses that yield a high likelihood are determined more likely to be correct, and are given a higher weight.

---

[1] A notable exception to this is recursive least-squares estimation of a linear combination of parameters [Ljung and Söderström, 1983; Åström and Wittenmark, 2013b].

The collection of particles will spread out more and more with each application of the dynamics model $f$. This is a manifestation of the curse of dimensionality, since the dimension of the space that the density $p(x_{0:t})$ occupies grows with $t$. To mitigate this, a re-sampling step is performed. The re-sampling favors particles with higher weights and thus focuses the attention of the finite collection of particles to areas of the state-space with high posterior density. We can thus think of the particle filter as a continuous analogue to the approximate branch-and-bound method beam search [Zhang, 1999].

The recent popularity of particle filters, fueled by the increase in available computational power, has led to a corresponding increase in publications describing the subject. Interested readers may refer to one of such publications for a more formal description of the subject, e.g., [Gustafsson, 2010; Thrun et al., 2005; Rawlings and Mayne, 2009].

We summarize the particle filter algorithm in Algorithm 1

---
**Algorithm 1** A simple particle filter algorithm.

---
Initialize particles using a prior distribution
**repeat**
    Assign weights to particles using likelihood under observation model $p(y|x)$
    (Optional) Calculate a state estimate based on the weighted collection of
        particles
    Re-sample particles based on weights
    Propagate particles forward using $p(x^+|x)$
**until** End of time

---

## 4.3 The Kalman Filter

The Kalman filter is a well-known algorithm to estimate the sequence of state distributions in a linear Gaussian state-space system, given noisy measurements [Åström, 2012]. The Kalman filter operates in one of the very few settings where the posterior density is available in closed form. Since both the state-transition function and the observation model are affine transformations of the state, the Gaussian distribution of the initial state remains Gaussian, both after a time update with Gaussian noise and after incorporating measurements corrupted with Gaussian noise. Instead of representing densities with a collection of particles as we did in the particle filter, we can now represent them exactly by a mean vector and a covariance matrix.

We will now proceed to derive the Kalman filter to establish the foundation for extensions provided later in the thesis. To facilitate the derivation, we provide two well-known lemmas regarding normal distributions:

LEMMA 1

The affine transformation of a normally distributed random variable is normally distributed with the following mean and variance

$$x \sim \mathcal{N}(\mu, \Sigma) \tag{4.2}$$

$$y = c + Bx \tag{4.3}$$

$$y \sim \mathcal{N}(c + B\mu, B\Sigma B^{\mathsf{T}}) \tag{4.4}$$

$$\square$$

LEMMA 2

When both prior and likelihood are Gaussian, the posterior distribution is Gaussian with

$$\mathcal{N}(\bar{\mu}, \bar{\Sigma}) = \mathcal{N}(\mu_0, \Sigma_0) \cdot \mathcal{N}(\mu_1, \Sigma_1) \tag{4.5}$$

$$\bar{\Sigma} = (\Sigma_0^{-1} + \Sigma_1^{-1})^{-1} \tag{4.6}$$

$$\bar{\mu} = \bar{\Sigma}(\Sigma_0^{-1}\mu_0 + \Sigma_1^{-1}\mu_1) \tag{4.7}$$

**Proof.** By multiplying the two probability-denisity functions in (4.5), we obtain (constants omitted)

$$\exp\left(-\frac{1}{2}(x - \mu_0)^{\mathsf{T}}\Sigma_0^{-1}(x - \mu_0) - \frac{1}{2}(x - \mu_1)^{\mathsf{T}}\Sigma_1^{-1}(x - \mu_1)\right)$$

$$= \exp\left(-\frac{1}{2}(x - \bar{\mu})^{\mathsf{T}}\bar{\Sigma}^{-1}(x - \bar{\mu})\right) \tag{4.8}$$

$$\bar{\Sigma} = (\Sigma_0^{-1} + \Sigma_1^{-1})^{-1} \tag{4.9}$$

$$\bar{\mu} = \bar{\Sigma}(\Sigma_0^{-1}\mu_0 + \Sigma_1^{-1}\mu_1) \tag{4.10}$$

where the terms in the first equation were expanded, all terms including $x$ collected and the square completed. Terms not including $x$ become part of the normalization constant and do not determine the mean or covariance.  $\square$

COROLLARY 1

The equations for the posterior mean and covariance can be written in *update form* according to

$$\bar{\mu} = \mu_0 + K(\mu_1 - \mu_0) \tag{4.11}$$

$$\bar{\Sigma} = \Sigma_0 - K\Sigma_0 \tag{4.12}$$

$$K = \Sigma_0(\Sigma_0 + \Sigma_1)^{-1} \tag{4.13}$$

**Proof.** The expression for $\bar{\Sigma}$ is obtained from the matrix inversion lemma applied to (4.6) and $\bar{\mu}$ is obtained by expanding $\bar{\Sigma}$, first in front of $\mu_0$ using (4.12), and then in front of $\mu_1$ using (4.6) together with the identity $(\Sigma_0^{-1} + \Sigma_1^{-1})^{-1} = \Sigma_0(\Sigma_0 + \Sigma_1)^{-1}\Sigma_1$.  $\square$

We now consider a state-space model of the form

$$x_{t+1} = Ax_t + Bu_t + v_t \tag{4.14}$$

$$y_t = Cx_t + e_t \tag{4.15}$$

where the noise terms $v$ and $e$ are independent[2] and Gaussian with mean zero and covariance $R_1$ and $R_2$, respectively. The estimation begins with an initial estimate of the state, $x_0$, with covariance $P_0$. By iteratively applying (4.14) to $x_0$, we obtain

$$\hat{x}_{t|t-1} = A\hat{x}_{t-1|t-1} + Bu_{t-1} \tag{4.16}$$

$$P_{t|t-1} = AP_{t-1|t-1}A^\mathsf{T} + R_1 \tag{4.17}$$

where both equations follow from Lemma 1 and the notation $\hat{x}_{i|j}$ denotes the estimate of $x$ at time $i$, given information available at time $j$. Equation (4.17) clearly illustrates that the covariance after a time update is the sum of a term due to the covariance from the previous time step and the added term $R_1$, which is the uncertainty added by the state-transition noise $v$. We further note that the properties of $A$ determine whether or not these equations alone are stable. For stable $A$ and $u \equiv 0$, the mean estimate of $x$ converges to zero with a stationary covariance given by the solution to the discrete-time Lyapunov equation $P = APA^\mathsf{T} + R_1$.

Equations (4.16) and (4.17) constitute the *prediction step*, we will now proceed to incorporate also a measurement of the state in the *measurement update step*.

By Lemma 1, the mean and covariance of the expected measurement is given by

$$\hat{y}_{t|t-1} = C\hat{x}_{t|t-1} \tag{4.18}$$

$$P_{t|t-1}^y = CP_{t|t-1}C^\mathsf{T} \tag{4.19}$$

We can now, using Corollary 1, write the posterior measurement as

$$\hat{y}_{t|t} = C\hat{x}_{t|t-1} + K_t^y(y_t - C\hat{x}_{t|t-1}) \tag{4.20}$$

$$P_{t|t}^y = P_{t|t-1}^y - K_t^y P_{t|t-1}^y \tag{4.21}$$

$$K_t^y = CP_{t|t-1}C^\mathsf{T}(CP_{t|t-1}C^\mathsf{T} + R_2)^{-1} \tag{4.22}$$

which, if we drop $C$ in front of both $\hat{y}$ and $P^y$, and $C^\mathsf{T}$ at the end of $P^y$, turns into

$$\hat{x}_{t|t} = \hat{x}_{t|t-1} + K_t(y_t - C\hat{x}_{t|t-1}) \tag{4.23}$$

$$P_{t|t} = P_{t|t-1} - K_t CP_{t|t-1} \tag{4.24}$$

$$K_t = P_{t|t-1}C^\mathsf{T}(CP_{t|t-1}C^\mathsf{T} + R_2)^{-1} \tag{4.25}$$

where $K$ is the *Kalman gain*.

---

[2] The case of correlated state-transition and measurement noise requires only a minor modification, but is left out for simplicity.

# 5

# Dynamic Programming

Dynamic programming (DP) is a general strategy due to Bellman (1953) for solving problems that enjoy a particular structure, often referred to as *optimal substructure*. In DP, the problem is broken down recursively into overlapping sub-problems, the simplest of which is easy to solve. While DP is used to solve problems in a diverse set of applications, such as sequence alignment, matrix-chain multiplication and scheduling, we will focus our introduction on the application to optimization problems where the sequential structure arises due to time, such as state-estimation, optimal control and reinforcement learning.

## 5.1 Optimal Control

A particularly common application of DP is optimal control [Åström, 2012; Bertsekas et al., 2005]. Given a cost function $c(x, u)$, a dynamics model $x^+ = f(x, u)$, and a fixed controller $\mu$ generating $u$, the sum of future costs at time $t$ can be written as a sum of the cost in the current time step $c_t = c(x_t, u_t)$, and the sum of future costs $c_{t+1} + c_{t+2} + ... + c_T$. We call this quantity the *value function $V^\mu(x_t)$* of the current policy $\mu$, and note that it can be defined recursively as

$$V^\mu(x_t) = c_t + V^\mu(x_{t+1}) = c_t + V^\mu\big(f(x_t, u_t)\big) \tag{5.1}$$

Of particular interest is the *optimal value function $V^*$*, i.e., the value function of the optimal controller $\mu^*$:

$$V^*(x_t) = \min_u \big(c(x_t, u) + V^*\big(f(x_t, u)\big)\big) \tag{5.2}$$

which defines the optimal controller $\mu^* = \operatorname{argmin}_u c(x_t, u) + V^*\big(f(x_t, u)\big)$. Thus, if we could somehow determine $V^*$, we would be one step closer to having found the optimal controller (solving for $\operatorname{argmin}_u$ could still be a difficult problem). Determining $V^*$ is in general hard and the literature is full of methods for both the general and special cases. We will refrain from discussing the general case here and only comment on some special cases.

**Linear Quadratic Regulation**

Just as the state-estimation problem enjoyed a particularly simple solution when the dynamics were linear and the noise was Gaussian, the optimal control problem has a particularly simple solution when the same conditions apply [Åström, 2012]. The value function in the last time step is simply $V_T^* = \min_u c(x_T, u)$ and is thus a quadratic function in $x_T$. The real magic happens when we note that the set of convex quadratic functions is closed under summation and minimization, meaning that $V_{T-1}^* = \min_u (c_{T-1} + V_T^*)$ is also a quadratic function, this time in $x_{t-1}$.[1] We can thus both solve for $V_{T-1}^*$ and represent it efficiently using a single positive definite matrix. The algorithm for calculating the optimal $V^*$ and the optimal controller $\mu^*$ is in this case called the Linear-Quadratic Regulator (LQR) [Åström, 2012].

The similarity with the Kalman filter is no coincidence. The Kalman filter essentially solves the maximum-likelihood problem, which when the noise is Gaussian is equivalent to solving a quadratic optimization problem. The LQR algorithm and the Kalman filter are thus dual to each other. This duality between linear control and estimation problems is well known and most classical control texts discuss it. In Chap. 7, we will explore the similarities further and let them guide us to efficient algorithms for identification problems.

**Iterative LQR**

The LQR algorithm is incredibly powerful in the restricted setting where it applies. In $\mathcal{O}(T)$ time it calculates both the optimal policy and the optimal value function. Its applicability is unfortunately limited to linear systems, but these systems may be time varying. An algorithm that makes use of LQR for nonlinear systems is Iterative LQR (iLQR) [Todorov and Li, 2005]. By linearizing the nonlinear system along the trajectory, the LQR algorithm can be employed to estimate an optimal control signal sequence. This sequence can be applied to the nonlinear system in simulation to obtain a new trajectory along which we can linearize the system and repeat the procedure. This algorithm is a special case of a more general algorithm, Differential Dynamic Programming (DDP) [Mayne, 1966], where a quadratic approximation to both a general cost function and a nonlinear dynamics model is formed along a trajectory, and the dynamic-programming problem is solved.

Since both DDP, and the special case iLQR, make use of linear approximations of the dynamics, a line search or trust region must be employed in order to ensure convergence. We will revisit this topic in Chap. 11, where we employ iLQR to solve an reinforcement-learning problem using estimation techniques developed in Chap. 7.

---

[1] Showing this involves algebra remarkably similar to the derivations in Sec. 4.3.

## **5.2 Reinforcement Learning**

The field of Reinforcement Learning (RL) has grown tremendously in recent years as the first RL methods making use of deep learning made significant strides to solving problems that were previously thought to be decades away from a solution. Noteworthy examples include the victory of the RL system AlphaGO against the human world champion in the game of GO [Silver et al., 2016].

When both cost function and dynamics are known, solving for $V^*$ is referred to as optimal control [Bertsekas et al., 2005]. If either of the two functions is unknown, the situation is made considerably more difficult. If the cost is known but the dynamics are unknown, one common strategy is to perform system identification and use the estimated model for optimal control. The same can be done with a cost function that is only available through sampling. Oftentimes, however, the state space is too large, and one can not hope to obtain globally accurate models of $c$ and $f$ from identification. In this setting, we may instead resort to reinforcement learning.

Reinforcement learning is, in all essence, a trial-and-error approach in which a controller interacts with the environment and uses the observed outcome to guide future interaction. The goal is still the same, to minimize a cumulative cost. The way we make use of the observed data to guide future interaction to reach this goal is what distinguishes different RL methods from each other. RL is very closely related to the field of adaptive control [Åström and Wittenmark, 2013b], although the terminology and motivating problems often differ. The RL community often considers a wider range of problems, such as online advertising and complex games with a discrete action set, while the adaptive control community long has had an emphasis on control using continuous action sets and low-complexity controllers, one of the main areas in which RL techniques have yet to prove effective.

RL algorithms can be broadly classified using a number of dichotomies; some methods try to estimate the value function, whereas some methods estimate the policy directly. Some methods estimate a dynamics model, we call these model-based methods, whereas some are model free. We indicate how some examples from the literature fit into this framework in Table 5.1.

Algorithms that try to estimate the value function can further be subdivided into two major camps; some use the Bellman equation and hence a form of dynamic programming, whereas some estimate the value function based on observed samples of the cost function alone in a Monte-Carlo fashion.

The failure of RL methods in continuous domains can often be traced back to their inefficient use of data. Many state-of-the-art methods require on the order of millions of episodic interactions with the environment in order to learn a successful controller [Mnih et al., 2015]. A fundamental problem with data efficiency in many modern RL methods stems from what they choose to model and learn. Methods that learn the value function are essentially trying to use the incoming data to hit a moving target. In the early stages of learning, the estimate of the value function and the controller are sub-optimal. In this early stage, the incoming data does not always hold any information regarding the optimal value function, which is the ultimate goal of learning. Model-based methods, on the other hand, use the incoming data to learn about the dynamics of the agent and the environment. While it is possible to imagine an environment with evolving dynamics, the dynamics are often laws of nature and do not change, or at least change much slower than the value function and the policy, quantities we are explicitly modifying continuously. This is one of the main reasons model-based methods tend to be more data efficient than model-free methods.

**Table 5.1**   The RL landscape. Methods marked with * or (*) estimate (may estimate) a value function and methods marked with a † or (†) estimate (may estimate) an explicit policy [1][Levine and Koltun, 2013], [2][Sutton et al., 2000], [3][Watkins and Dayan, 1992], [4][Sutton, 1991], [5][Silver et al., 2014], [6][Rummery and Niranjan, 1994], [7][Williams, 1988], [8][Schulman et al., 2015].

|  | Model based | Model free |
|---|---|---|
| Dynamics known | Optimal control (*,†) Policy/Value iteration | If simulation/experiments are very fast |
| Dynamics unknown | Guided Policy Search[2]*† Model free methods with simulation (DYNA[5]) (*,†) | Policy gradient[3]† Q-learning[4]* DPG[6]*† SARSA[7]* REINFORCE[8]† TRPO[9]† |

Model-based methods are not without problems though. Optimization under an inaccurate model might cause the RL algorithm to diverge. In Chap. 11, we will make use of models and identification methods developed in Part I for reinforcement-learning purposes. The strategy will be based on trajectory optimization under estimated models and an estimate of the uncertainty in the estimated model will be taken into account during the optimization.

# 6

# Linear Quadratic Estimation and Regularization

This chapter introduces a number of well-known topics and serves as an introduction to the reader unfamiliar with concepts such as singular value decomposition, linear least-squares, regularization and basis-function expansions. These methods will be used extensively in this work, where they are only briefly introduced as needed. Readers familiar with these topics can skip this chapter.

## 6.1 Singular Value Decomposition

The singular value decomposition (SVD) [Golub and Van Loan, 2012] was first developed in the late 1800s for bilinear forms, and later extended to rectangular matrices by [Eckart and Young, 1936]. The SVD is a factorization of a matrix $A \in \mathbb{R}^{N \times M}$ on the form

$$A = USV^\mathsf{T}$$

where the matrices $U \in \mathbb{R}^{N \times N}$ and $V \in \mathbb{R}^{M \times M}$ are orthonormal, such that $U^\mathsf{T}U = UU^\mathsf{T} = I_N$ and $V^\mathsf{T}V = VV^\mathsf{T} = I_M$, and $S = \text{diag}(\sigma_1, ..., \sigma_m) \in \mathbb{R}^{N \times M}$ is a rectangular, diagonal matrix with the singular values on the diagonal. The singular values are the square roots of the eigenvalues of the matrices $AA^\mathsf{T}$ and $A^\mathsf{T}A$ and are always nonnegative and real. The orthonormal matrices $U$ and $V$ can be shown to have columns consisting of a set of orthonormal eigenvectors of $AA^\mathsf{T}$ and $A^\mathsf{T}A$ respectively.

One of many applications of the SVD that will be exploited in this thesis is to find the equation for a plane that minimizes the sum of squared distances between the plane and a set of points. The normal to this plane is simply the singular vector corresponding to the smallest singular value of a matrix composed of all point coordinates. The smallest singular value will in this case correspond to the mean squared distance between the points and the plane, i.e., the variance of the residuals.

**Finding the closest orthonormal matrix**

A matrix $R$ is said to be orthonormal if $R^\mathsf{T}R = RR^\mathsf{T} = I$. If the additional fact $\det(R) = 1$ holds, the matrix is said to be a rotation matrix, an element of the $n$-dimensional special orthonormal group $SO(n)$ [Murray et al., 1994; Mooring et al., 1991].

Given an arbitrary matrix $\tilde{R} \in \mathbb{R}^{3 \times 3}$, the closest rotation matrix in $SO(3)$, in the sense $||R - \tilde{R}||_F$, can be found by Singular Value Decomposition according to [Eggert et al., 1997]

$$\tilde{R} = USV^\mathsf{T} \tag{6.1}$$

$$R = U \begin{bmatrix} 1 & & \\ & 1 & \\ & & \det(UV^\mathsf{T}) \end{bmatrix} V^\mathsf{T} \tag{6.2}$$

## 6.2 Least-Squares Estimation

This thesis will frequently deal with the estimation of models which are linear in the parameters, and thus can be written on the form

$$y = \mathbf{A}k \tag{6.3}$$

where $\mathbf{A}$ denotes the regressor matrix and $k$ denotes a vector of coefficients to be identified. Models on the form (6.3) are commonly identified with the well-known least-squares procedure [Johansson, 1993]. As an example, we consider the model $y_n = k_1 u_n + k_2 v_n$, where a measured signal $y$ is a *linear combination* of two input signals $u$ and $v$. The identification task is to identify the parameters $k_1$ and $k_2$. In this case, the procedure amounts to arranging the data according to

$$y = \begin{bmatrix} y_1 \\ \vdots \\ y_N \end{bmatrix}, \quad \mathbf{A} = \begin{bmatrix} u_1 & v_1 \\ \vdots & \vdots \\ u_N & v_N \end{bmatrix} \in \mathbb{R}^{N \times 2}, \quad k = \begin{bmatrix} k_1 \\ k_2 \end{bmatrix}$$

and solving the optimization problem of Eq. (6.4) with solution (6.5).

THEOREM 1
The vector $k^*$ of parameters that solves the optimization problem

$$k^* = \underset{k}{\arg\min} \left\| y - \mathbf{A}k \right\|_2^2 \tag{6.4}$$

is given by the closed-form expression

$$k^* = \left(\mathbf{A}^\mathsf{T}\mathbf{A}\right)^{-1}\mathbf{A}^\mathsf{T}y \tag{6.5}$$

***Proof.*** Completion of squares in the least-squares cost function $J$ yields

$$
\begin{aligned}
J &= \left\| y - \mathbf{A}k \right\|_2^2 = (y - \mathbf{A}k)^{\mathsf{T}}(y - \mathbf{A}k) \\
&= y^{\mathsf{T}}y - y^{\mathsf{T}}\mathbf{A}k - k^{\mathsf{T}}\mathbf{A}^{\mathsf{T}}y + k^{\mathsf{T}}\mathbf{A}^{\mathsf{T}}\mathbf{A}k \\
&= \left( k - \left(\mathbf{A}^{\mathsf{T}}\mathbf{A}\right)^{-1}\mathbf{A}^{\mathsf{T}}y \right)^{\mathsf{T}}\mathbf{A}^{\mathsf{T}}\mathbf{A}\left( k - \left(\mathbf{A}^{\mathsf{T}}\mathbf{A}\right)^{-1}\mathbf{A}^{\mathsf{T}}y \right) + y^{\mathsf{T}}(I - \mathbf{A}\left(\mathbf{A}^{\mathsf{T}}\mathbf{A}\right)^{-1}\mathbf{A}^{\mathsf{T}})y
\end{aligned}
$$

where we identify the last expression as a sum of two terms, one that does not depend on $k$, and a term which is a positive definite quadratic form ($\mathbf{A}^{\mathsf{T}}\mathbf{A}$ is always positive (semi)definite). The estimate $k^*$ that minimizes $J$ is thus the value that makes the quadratic form equal to zero. □

The expression (6.5) is known as the least-squares solution and the full-rank matrix $(\mathbf{A}^{\mathsf{T}}\mathbf{A})^{-1}\mathbf{A}^{\mathsf{T}}$ is commonly referred to as the pseudo inverse of $\mathbf{A}$. If $\mathbf{A}$ is a square matrix, the pseudo inverse reduces to the standard matrix inverse. If $\mathbf{A}$, however, is a tall matrix, the equation $y = \mathbf{A}k$ is over determined and (6.5) produces the solution $k^*$ that minimizes (6.4). We emphasize that the important property of the model $y_n = k_1 u_n + k_2 v_n$ that allows us to find the solution to the optimization problem on closed-form is that the parameters to be estimated enter linearly. The signals $u$ and $v$ may be arbitrarily complex functions of some other variable, as long as these are known.

### Consistency

A consistent estimate is one which is asymptotically unbiased and has vanishing variance as the number of data points grows. The consistency of the least-squares estimate can be analyzed by calculating the bias and variance properties. Consider the standard model, with an added noise term $v$, for which consistency is given by the following theorem:

THEOREM 2
The closed-form expression $\hat{k} = \left(\mathbf{A}^{\mathsf{T}}\mathbf{A}\right)^{-1}\mathbf{A}^{\mathsf{T}}y$ is an unbiased and consistent estimate of $k$ in the model

$$
\begin{aligned}
y &= \mathbf{A}k + v \\
v &\sim \mathcal{N}(0, \sigma^2) \\
\mathbb{E}\left\{\mathbf{A}^{\mathsf{T}}v\right\} &= 0
\end{aligned}
$$

***Proof.*** The bias and variance of the resulting least-squares based estimate are:

**Bias**  We begin be rewriting the expression for the estimate $\hat{k}$ as

$$
\begin{aligned}
\hat{k} &= \left(\mathbf{A}^{\mathsf{T}}\mathbf{A}\right)^{-1}\mathbf{A}^{\mathsf{T}}y \\
&= \left(\mathbf{A}^{\mathsf{T}}\mathbf{A}\right)^{-1}\mathbf{A}^{\mathsf{T}}(\mathbf{A}k + v) \\
&= k + \left(\mathbf{A}^{\mathsf{T}}\mathbf{A}\right)^{-1}\mathbf{A}^{\mathsf{T}}v
\end{aligned}
$$

If the regressors are uncorrelated with the noise, $\mathbb{E}\left\{\left(\mathbf{A}^\mathsf{T}\mathbf{A}\right)^{-1}\mathbf{A}^\mathsf{T}\nu\right\} = 0$, we can conclude that $\mathbb{E}\left\{\hat{k}\right\} = k$ and the estimate is unbiased.

**Variance**   The variance is given by

$$\mathbb{E}\left\{(\hat{k} - k)(\hat{k} - k)^\mathsf{T}\right\} = \mathbb{E}\left\{(\mathbf{A}^\mathsf{T}\mathbf{A})^{-1}\mathbf{A}^\mathsf{T}\nu\,\nu^\mathsf{T}\mathbf{A}(\mathbf{A}^\mathsf{T}\mathbf{A})^{-\mathsf{T}}\right\}$$
$$= \sigma^2\mathbb{E}\left\{(\mathbf{A}^\mathsf{T}\mathbf{A})^{-1}\right\}$$
$$= \sigma^2(\mathbf{A}^\mathsf{T}\mathbf{A})^{-1}$$

where the second equality holds if $\nu$ and $\mathbf{A}$ are uncorrelated. As $N \to \infty$, we have $\sigma^2(\mathbf{A}^\mathsf{T}\mathbf{A})^{-1} \to 0$, provided that the Euclidean length of all columns in $\mathbf{A}$ increases as $N$ increases. $\qquad\square$

### Other loss functions

The least-squares loss function

$$k^* = \operatorname*{arg\,min}_{k} \left\| y - \mathbf{A}k \right\|_2^2$$

is convex and admits a particularly simple, closed-form expression for the minimum. If another norm is used instead of the $L_2$ norm, the estimate will have different properties. The choice of other norms will, in general, not admit a solution on closed form, but for many norms of interest, the optimization problem remains *convex*. This fact will in practice guarantee that a global minimum can be found easily using iterative methods. Many of the methods described in this thesis could equally well be solved with another convex loss function, such as the $L_1$ norm for increased robustness, or the $L_\infty$ norm for a minimum worst-case scenario. For an introduction to convex optimization and a description of the properties of different convex loss functions, see [Boyd and Vandenberghe, 2004].

### Computation

Although the solutions to the least-squares problems are available in closed form, it is ill-advised to actually perform the calculation $k = \left(\mathbf{A}^\mathsf{T}\mathbf{A}\right)^{-1}\mathbf{A}^\mathsf{T}y$ [Golub and Van Loan, 2012]. Numerically more robust strategies include

- performing a Cholesky factorization of the symmetric matrix $\mathbf{A}^\mathsf{T}\mathbf{A}$.

- performing a QR-decomposition of $\mathbf{A}$.

- performing a singular value decomposition (SVD) of $\mathbf{A}$.

where the latter two methods avoid the calculation of $\mathbf{A}^\mathsf{T}\mathbf{A}$ altogether, which can be subject to numerical difficulties if $\mathbf{A}$ has a high condition number [Golub and Van Loan, 2012]. In fact, the method of performing a Cholesky decomposition of $\mathbf{A}^{\mathsf{T}}\mathbf{A}$ can be implemented using the QR-decomposition since triangular matrix $R$ obtained by a QR-decomposition of $\mathbf{A}$ is a Cholesky factor of $\mathbf{A}^{\mathsf{T}}\mathbf{A}$:

$$\mathbf{A}^{\mathsf{T}}\mathbf{A} = (QR)^{\mathsf{T}}(QR) = R^{\mathsf{T}}R$$

Many numerical computation tools, including *Julia*, *Matlab* and *numpy*, provide numerically robust methods to calculate the solution to the least-squares problem, indicated in Algorithm 2. These methods typically analyze the matrix $\mathbf{A}$ and choose a suitable numerical algorithm to execute based on its properties [Julialang, 2017].

---

**Algorithm 2** Syntax for solving the least-squares problem $k = (\mathbf{A}^{\mathsf{T}}\mathbf{A})^{-1}\mathbf{A}^{\mathsf{T}}y$ in different programming languages.

---

```
k = A\y                        # Julia
k = A\y                        % Matlab
k = numpy.linalg.solve(A, y)   # Python with numpy
k <- solve(A, y)               # R
```

---

If the number of features, and thus the matrix $\mathbf{A}^{\mathsf{T}}\mathbf{A}$, is too large for the problem to be solved by factorizing $\mathbf{A}$, an iterative method such as conjugate gradients or GMRES [Saad and Schultz, 1986] can solve the problem by performing matrix-vector products only.

## 6.3 Basis-Function Expansions

When estimating a functional relationship between two or more variables, i.e., $y = f(v)$, a standard initial approach is *linear regression* using the least-squares procedure. A strong motivation for this is the fact that the optimal linear combination of the chosen basis functions, or *regressors*, is available in closed form. A typical choice of basis functions is low-order monomials, e.g., a decomposition of a signal $y$ according to

$$y = \phi(v)k = k_0 + k_1 v^1 + k_2 v^2 + ... + k_J v^J \tag{6.6}$$

where $\phi(v) = [v^0 \ v^1 \ ... \ v^J]$ is the set of basis function activations. The function $f(v) = \phi(v)k$ can be highly nonlinear and even discontinuous in $v$, but is *linear in the parameters*, making it easy to fit to data.

While the low order monomials $v^i$ are easy to work with and provide reasonable fit when the relationship between $y$ and $v$ is simple, they tend to perform worse when the relationship is complex.

Intuitively, a basis-function expansion (BFE) decomposes an intricate function or signal as a linear combination of simple basis functions. The Fourier transform can be given this interpretation, where an arbitrary signal is decomposed as a sum

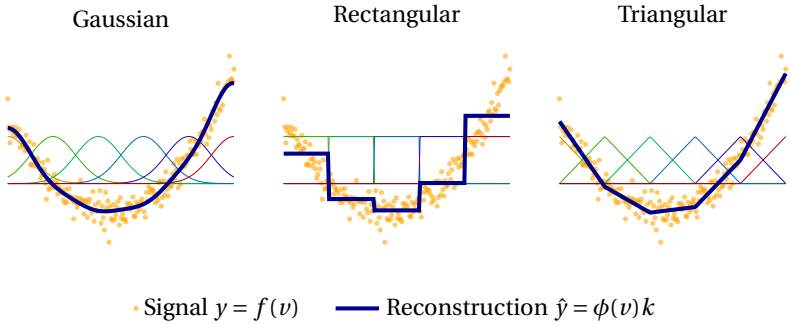

Gaussian   Rectangular   Triangular

• Signal $y = f(v)$ ━━ Reconstruction $\hat{y} = \phi(v)k$

**Figure 6.1** Reconstructions of a sampled signal $y = f(v)$ using different sets of basis functions. The basis functions used for the decomposition of $y$ is shown in the background.

of complex-valued sinusoids. Similarly, a stair function can be decomposed as a sum of step functions.

In many situations, there is no a priori information regarding the relationship between the free variable $v$ and the dependent variable $y$, and it might be hard to choose a suitable set of basis functions to use for a decomposition of the signal $y$. In such situations, an alternative is to choose a set of functions with *local support*, spread out to cover the domain of $v$. Some examples of basis functions with local support are: Gaussian radial basis functions $\kappa(v) = \exp\left(-\gamma(v-\mu)^2\right)$ which carry the implicit assumption of a smooth function $f$, triangular functions $\kappa(v) = \max(0, 1-\gamma|v-\mu|)$ resulting in a piecewise affine $\hat{y}$ and rectangular functions $\kappa(v) = |v-\mu| < \gamma$ where $\gamma = \Delta\mu$, resulting in piecewise constant $\hat{y}$. In the last example, we interpret the Boolean values true/false as 1/0. In all cases, $\mu$ determines the center of the basis function and $\gamma$ determines the width. Examples of decompositions using these basis functions are shown in Fig. 6.1. In many situations, there is no a priori information regarding the relationship between the free variable $v$ and the dependent variable $y$, and it might be hard to choose a suitable set of basis functions to use for a decomposition of the signal $y$. In such situations, an alternative is to choose a set of functions with *local support*, spread out to cover the domain of $v$. Some examples of basis functions with local support are: radial basis functions $\kappa(v) = \exp\left(-\gamma(v-\mu)^2\right)$ which carry the implicit assumption of a smooth function $f$, triangular functions $\kappa(v) = \max(0, 1-\gamma|v-\mu|)$ resulting in a piecewise affine $\hat{y}$ and rectangular functions $\kappa(v) = |v-\mu| < \gamma$ where $\gamma = \Delta\mu$, resulting in piecewise constant $\hat{y}$. In the last example, we interpret the Boolean values true/false as 1/0. In all cases, $\mu$ determines the center of the basis function and $\gamma$ determines the width. Examples of decompositions using these basis functions are shown in Fig. 6.1. Other motivations for considering basis functions with local support include making the modeling more intuitive and the result easier to interpret. This in contrast to basis functions with global support,

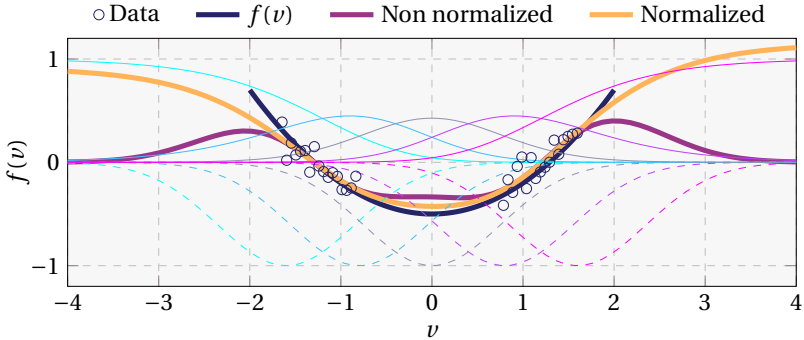

**Figure 6.2** Basis-function expansions fit to noisy data from the function $f(v) = 0.3v^2 - 0.5$ using normalized (-) and nonnormalized (- -) basis functions. Non-normalized basis functions are shown mirrored in the vertical axis.

such as sigmoid-type functions or monomials.

Basis-function expansions are related to Gaussian processes [Rasmussen, 2004]. We will not make this connection explicit in this work, but the interested reader might find the Bayesian interpretation provided by a Gaussian process worthwhile.

The concept of basis-function expansions will be used extensively in the thesis. The open-source software accompanying many of the papers in this thesis makes use of basis-function expansions. This functionality has been externalized into the software package [*BasisFunctionExpansions.jl*, B.C., 2016], which provides many convenient methods for working with basis-function expansions.

### Normalization

For some applications, it may be beneficial to normalize the kernel vector for each input point [Bugmann, 1998] such that

$$\bar{\phi}(v) = \left( \sum_{i=1}^{K} \kappa(v, \mu_i, \gamma_i) \right)^{-1} \phi(v)$$

One major difference between a standard BFE and a normalized BFE (NBFE) is the behavior far (in terms of the width of the basis functions) from the training data. The prediction of a BFE will tend towards zero, whereas the prediction from an NBFE tends to keep its value close to the boundary of the data. Figure 6.2 shows the fit of two BFEs of the function $f(v) = 0.3v^2 - 0.5$ together with the basis functions used. The BFE tends towards zero both outside the data points and in the interval of missing data in the center. The NBFE on the other hand generalizes better and keeps its current prediction trend outside the data. The performance of NBFEs is studied in detail in [Bugmann, 1998].

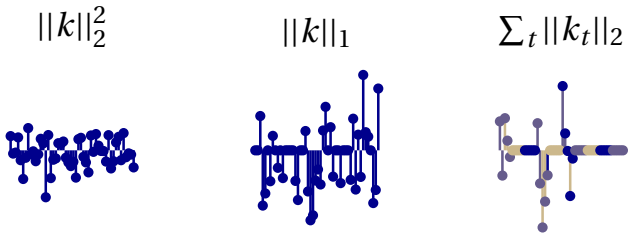

$||k||_2^2$ $||k||_1$ $\sum_t ||k_t||_2$

**Figure 6.3** Illustration of the effect of different regularization terms on the structure of the solution vector. The squared 2-norm promotes small components, the 1-norm promotes sparse components (few nonzero elements) and the nonsquared, group-wise 2-norm promotes sparse groups (clusters of elements).

## 6.4 Regularization

Regularization is a general concept which can, as a result of several overloaded meanings of the term, be hard to define. The purposes of regularization include preventing *overfitting* by penalizing complexity, improving numerical robustness at the cost of bias and imposing a special *structure* on the solution such as sparsity [Murphy, 2012].

Regularization will in our case amount to adding a term to the cost function of the optimization problem we are solving according to

$$\underset{k}{\text{minimize}} \left\| y - \hat{y}(k) \right\|^2 + \lambda^2 f(k)$$

Examples of $f$ include

- $\left\| k \right\|_2^2$ to promote *small k*

- $\left\| k \right\|_1$ to promote *sparse k*

- $\left\| k_t \right\|_2$ to promote *group-sparse $k_t$*

with effects illustrated in Fig. 6.3. We will detail the effects of the mentioned example regularization terms in the following sections.

### Squared $L_2$-regularized regression

For certain problems, it might be desirable to add a term to the cost function (6.4) that penalizes the size of the estimated parameter vector, for some notion of size. This might be the case if the problem is *ill-posed*, or if we have the *a priori knowledge* that the parameter vector is small. Classically, methods such as Akaike Information Criterion (AIC) and Bayesian Information Criterion (BIC) have penalized the dimensionality of the parameter vector [Ljung, 1987; Johansson, 1993]. These methods are, however, impractical in modern machine learning where the number of model parameters is very large and therefore seldom used. In machine learning it is more common to penalize a convex function of the parameter vector, such as its norm. Depending on the norm in which we measure the size of the parameter vector, this procedure has many names. For the common $L_2$ norm, the resulting method is commonly referred to as Tikhonov regularized regression, ridge regression or weight decay if one adopts an optimization perspective, or maximum a posteriori (MAP) estimation with a Gaussian prior, if one adopts a Bayesian view on the estimation problem [Murphy, 2012]. If the problem is linear in the parameters, the solution to the resulting optimization problem remains on a closed form, as indicated by the following theorem. Here, we demonstrate an alternative way of proving the least-squares solution, based on differentiation instead of completion of squares.

THEOREM 3

The vector $k^*$ of parameters that solves the optimization problem

$$k^* = \underset{k}{\arg\min} \frac{1}{2} \left\| y - \mathbf{A}k \right\|_2^2 + \frac{\lambda}{2} \left\| k \right\|_2^2 \tag{6.7}$$

is given by the closed-form expression

$$k^* = (\mathbf{A}^\mathsf{T}\mathbf{A} + \lambda I)^{-1}\mathbf{A}^\mathsf{T} y \tag{6.8}$$

***Proof.*** Differentiation of the cost function yields

$$J = \frac{1}{2} \left\| y - \mathbf{A}k \right\|_2^2 + \frac{\lambda}{2} \left\| k \right\|_2^2 = \frac{1}{2}(y - \mathbf{A}k)^\mathsf{T}(y - \mathbf{A}k) + \frac{\lambda}{2} k^\mathsf{T} k$$

$$\frac{dJ}{dk} = -\mathbf{A}^\mathsf{T}(y - \mathbf{A}k) + \lambda k$$

If we equate this last expression to zero we get

$$\frac{dJ}{dk} = -\mathbf{A}^\mathsf{T}(y - \mathbf{A}k) + \lambda k = 0$$

$$(\mathbf{A}^\mathsf{T}\mathbf{A} + \lambda I)k = \mathbf{A}^\mathsf{T} y$$

$$k = (\mathbf{A}^\mathsf{T}\mathbf{A} + \lambda I)^{-1}\mathbf{A}^\mathsf{T} y$$

Since $\mathbf{A}^\mathsf{T}\mathbf{A}$ is positive semi-definite, both first- and second-order conditions for a minimum are satisfied by $k^* = (\mathbf{A}^\mathsf{T}\mathbf{A} + \lambda I)^{-1}\mathbf{A}^\mathsf{T} y$. $\qquad\Box$

REMARK 1

When deriving the expression for $k^*$ by differentiation, the terms $1/2$ appearing in (6.7) are commonly inserted for aesthetical purposes, they do not affect $k^*$. $\quad\Box$

We immediately notice that the solution to the regularized problem (6.7) reduces to the solution of the ordinary least-squares problem (6.4) in the case $\lambda = 0$. The regularization adds the positive term $\lambda$ to all diagonal elements of $\mathbf{A}^\mathsf{T}\mathbf{A}$, which reduces the condition number of the matrix to be inverted and ensures that the problem is well posed [Golub and Van Loan, 2012]. The regularization reduces the variance in the estimate at the expense of the introduction of a bias.

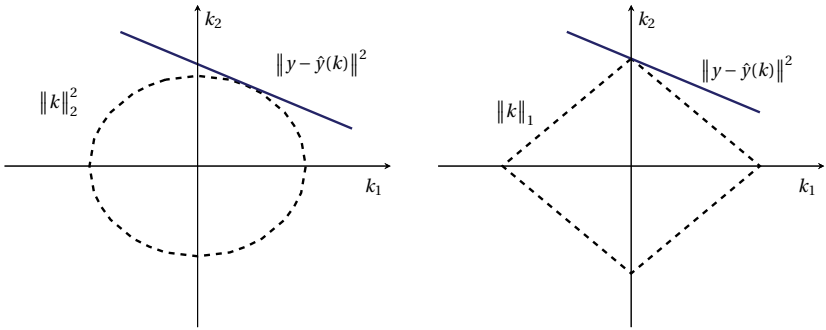

**Figure 6.4** The sparsity-promoting effect of the $L_1$ norm can be understood by considering the level surfaces of the penalty function (dashed). The level surface to the objective function (solid) is likely to hit a particular level surface of $||k||_1$ on a corner, whereas this does not hold for the $||k||_2^2$ penalty function.

For numerically robust methods of solving the ridge-regression problem, see, e.g., the excellent manual by Hansen (1994).

### $L_1$-regularized regression

$L_1$ regularization is commonly used to promote sparsity in the solution [Boyd and Vandenberghe, 2004; Murphy, 2012]. The sparsity promoting effect can be understood by considering the gradient of the penalty function, which remains large even for small values of the argument. In contrast, the squared $L_2$ norm has a vanishing gradient for small arguments. Further intuition for the sparsity promoting quality of the $L_1$ norm is gained by considering the level curves of the function, see Fig. 6.4. A third way of understanding the properties of the $L_1$ norm penalty is as the convex relaxation of the $L_0$ penalty $\sum_i \mathbb{I}(k_i)$, i.e., the number of nonzero entries in $k$.

The $L_1$ norm is a convex function, but $L_1$-regularized problems do not admit a solution on closed form and worse yet, the $L_1$ is nonsmooth. When this kind of problems arises in this thesis, the ADMM algorithm [Parikh and Boyd, 2014] will be employed to efficiently find a solution.[1]

### Group $L_2$-regularized regression

In some applications, the parameter vector $k$ of an optimization problem is naturally divided into groups $\{k_t\}_{t=1}^T$. One such situation is the estimation of a linear time-varying model $x^+ = A_t x_t + B_t u_t$, where $k_t = \text{vec}(\begin{bmatrix} A_t^\mathsf{T} & B_t^\mathsf{T} \end{bmatrix})$ forms a natural grouping of variables.

---

[1] Our implementation of ADMM makes use of the software package [*ProximalOperators.jl*, S. et al., 2016] for efficient and convenient proximal operations.

If the parameter vector is known to be *group sparse*, i.e., some of the groups are exactly zero, one way to encourage a solution with this property is to add the group-lasso penalty [Yuan and Lin, 2006]

$$\sum_t \| k_t \|_2 \tag{6.9}$$

The addition of (6.9) to the cost function will promote a solution where the length of some $k_t$ is exactly zero. To understand why this is the case, one can interpret (6.9) as the $L_1$-norm of lengths of vectors $k_t$. Of key importance in (6.9) is that the norm is nonsquared, as the sum of squared $L_2$ norms over groups coincides with the standard squared $L_2$ norm penalty without groups

$$\sum_t \| k_t \|_2^2 = \| k \|_2^2$$

The group $L_2$-regularization term is also convex but nonsmooth. We employ the *linearized* ADMM algorithm [Parikh and Boyd, 2014] to find a solution.

**Trend filtering**

An important class of signal-reconstruction methods that has been popularized lately is *trend filtering* methods [Kim et al., 2009; Tibshirani et al., 2014]. Trend filtering methods work by specifying a *fitness criterion* that determines the goodness of fit, as well as a regularization term, often chosen with sparsity promoting qualities. As a simple example, consider the reconstruction $\hat{y}$ of a noisy signal $y = \{y_t \in \mathbb{R}\}_{t=1}^T$ with piecewise constant segments. To this end, we may formulate and solve the convex optimization problem

$$\underset{\hat{y}}{\text{minimize}} \, \| y - \hat{y} \|_2^2 + \lambda \sum_t |\hat{y}_{t+1} - \hat{y}_t| \tag{6.10}$$

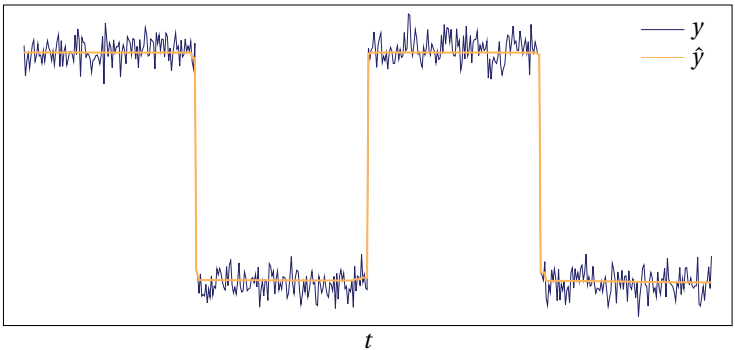

**Figure 6.5** Example of signal reconstruction by means of trend filtering with a sparsity promoting regularization term (6.11).

We note that (6.10) can be written on the form

$$\underset{\hat{y}}{\text{minimize}} \, \|y - \hat{y}\|_2^2 + \lambda \, \|D_1 \hat{y}\|_1 \tag{6.11}$$

where $D_1$ is the first-order difference operator, and we thus realize that the solution will have a sparse first order time difference, see Fig. 6.5 for an example application.

We remark that trend filtering is a noncasual operation and would with the terminology employed in this thesis technically be referred to as a smoothing operation.

## 6.5 Estimation of LTI Models

Linear time-invariant models are fundamental within the field of control, and decades of research have been devoted to their identification. We do not intend to cover much of this research here, but instead limit ourselves to establish notation and show how an LTI model lends itself to estimation by means of LS if the full state-sequence is known.

A general LTI model takes the form

$$x_{t+1} = Ax_t + Bu_t + v_t, \quad t \in [1, T] \tag{6.12}$$

$$y_t = Cx_t + Du_t + e_t \tag{6.13}$$

where $x \in \mathbb{R}^n$, $y \in \mathbb{R}^p$ and $u \in \mathbb{R}^m$ are the state, measurement and input respectively. A discussion around the noise terms $v_t$ and $e_t$ is deferred until Sec. 7.5, where we indicate how statistical assumptions on $v_t$ influence the cost function and the properties of the estimate. We further limit ourselves to the case where the state and input sequences are measured, i.e., $C = I$. This makes a plethora of methods for estimating the parameters available. A common method for identification of systems that are linear in the parameters is the least-squares (LS), prediction-error method (PEM), which in case of Gaussian noise, $v$, coincides with the maximum likelihood (ML) estimate. To facilitate estimation using the LS method, we write the model on the form $y = \mathbf{A}k$, and arrange the data according to

$$y = \begin{bmatrix} x_1 \\ \vdots \\ x_T \end{bmatrix} \qquad\qquad \in \mathbb{R}^{Tn}$$

$$k = \text{vec}\left(\begin{bmatrix} A^{\mathsf{T}} & B^{\mathsf{T}} \end{bmatrix}\right) \qquad\qquad \in \mathbb{R}^{K}$$

$$\mathbf{A} = \begin{bmatrix} I_n \otimes x_0^{\mathsf{T}} & I_n \otimes u_0^{\mathsf{T}} \\ \vdots & \vdots \\ I_n \otimes x_{T-1}^{\mathsf{T}} & I_n \otimes u_{T-1}^{\mathsf{T}} \end{bmatrix} \qquad\qquad \in \mathbb{R}^{Tn \times K}$$

where $\otimes$ denotes the Kronecker product and $K = n^2 + nm$ is the number of model parameters. We then solve the optimization problem (6.4) with closed-form solution (6.5).

Any linear system of order $n$ can be represented by a minimal representation where the matrix $A$ has $n$ free parameters [Ljung, 1987]. The measured state sequence might, however, not correspond to this minimal representation. Known sparsity structure in $A$ given a measured state representation can be enforced using constrained optimization or regularization.

In the next chapter, we will extend our view to linear time-varying models, where the parameters of the matrices $A$ and $B$ evolve as functions of time.

# 7

# Estimation of LTV Models

## 7.1 Introduction

Time-varying systems and models arise in many situations. A common case is the linearization of a nonlinear system along a trajectory [Ljung, 1987]. A linear time-varying (LTV) model obtained from such a procedure facilitates control design and trajectory optimization using linear methods, which are in many respects better understood than nonlinear control synthesis.

The difficulty of the task of identifying time-varying dynamical models varies greatly with the model considered and the availability of measurements of the state sequence. For smoothly changing dynamics, linear in the parameters, the recursive least-squares algorithm with exponential forgetting (RLS$\lambda$) is a common option. If a Gaussian random-walk model for the parameters is assumed, a Kalman filtering/smoothing algorithm [Rauch et al., 1965] gives the filtering/smoothing densities of the parameters in closed form. However, the assumption of Brownian-walk dynamics is often restrictive. Discontinuous dynamics changes occur, for instance, when an external controller changes operation mode, when a sudden contact between a robot and its environment is established, when an unmodeled disturbance enters the system or when a component in the system suddenly fails.

Identification of systems with nonsmooth dynamics evolution has been studied extensively. The book by Costa et al. (2006) treats the case where the dynamics are known, but the state sequence unknown, i.e., state estimation. Nagarajaiah and Li (2004) examine the residuals from an initial constant dynamics fit to determine regions in time where improved fit is needed, addressing this need by the introduction of additional constant dynamics models. Results on identifiability and observability in jump-linear systems in the noncontrolled (autonomous) setting are available due to Vidal et al. (2002). The main result on identifiability in [Vidal et al., 2002] was a rank condition on a Hankel matrix constructed from the collected output data, similar to classical results on the least-squares identification of ARX models which appears as rank constraints on the, typically Toeplitz or block-Toeplitz, regressor matrix. Identifiability of the problems proposed in this chapter is discussed in Sec. 7.3.

In this work, we draw inspiration from the trend-filtering literature to develop new system-identification methods for LTV models. In trend filtering, a curve is decomposed into a set of polynomial segments. In the identification methods proposed in this work, we instead consider a multivariable state sequence as the output of an LTV model, where the model coefficients evolve as polynomial functions of time. We start by defining a set of optimization problems of trend-filtering character, with a least-squares loss function and carefully chosen regularization terms. Similar inspiration was seen in [Ohlsson, 2010] for SISO systems, where generic solvers were employed to find a solution. We proceed to establish a connection between the proposed optimization problems and a statistical model of the evolution of the parameters constituting the LTV model, and use this model to guide us to efficient algorithms for solving the problems. We then discuss how prior information can be utilized to increase the accuracy of the identification in situations with poor excitation provided by the input signal, and end the chapter with two examples. The identification methods developed in this chapter are later used for model-based reinforcement learning in Chap. 11, where we note that an LTV model can be seen as a first-order approximation of the dynamics of a nonlinear system around a trajectory. We emphasize that such an approximation will, in general, fail to generalize far from this trajectory, but many methods in reinforcement learning and control make efficient use of the linearized dynamics for optimization, while ensuring validity of the approximation by constraints or penalty terms. A significant part of this chapter will be devoted to developing efficient solvers to the proposed methods and in some cases, provide a statistical interpretation that allows us to quantify the uncertainty in the model coefficients. This uncertainty estimate will be of importance in Chap. 11 where it allows us to adaptively constrain the step size of a trajectory optimization algorithm.

## 7.2  Model and Identification Problems

Linear time-varying models can be formulated in a number of different ways. We limit the scope of this work to models on the form

$$
\begin{aligned}
x_{t+1} &= A_t x_t + B_t u_t + v_t \\
k_t &= \text{vec}\left(\begin{bmatrix} A_t & B_t \end{bmatrix}^\mathsf{T}\right)
\end{aligned}
\tag{7.1}
$$

where the state-sequence $x_t$ is measured, possibly corrupted by measurement noise, and where the parameters $k$ are assumed to evolve according to the dynamical system

$$
\begin{aligned}
k_{t+1} &= H_t k_t + w_t \\
y_t &= \left(I_n \otimes \begin{bmatrix} x_t^\mathsf{T} & u_t^\mathsf{T} \end{bmatrix}\right) k_t + e_t
\end{aligned}
\tag{7.2}
$$

The model (7.1)-(7.2) is limited by its lack of noise models. However, this simple model will allow us to develop very efficient algorithms for identification. We defer the discussion on measurement noise to Sec. 7.9.

Upon inspection of (7.2), the connection between the present model identification problem and the state-estimation problem of Chap. 4 should be apparent.

The model (7.2) implies that the coefficients of the LTV model themselves evolve according to a linear dynamical system, and are thus amenable to estimation using state-estimation techniques. If no prior knowledge is available, the dynamics matrix $H_t$ can be taken as the identity matrix, $H = I$, implying that the model coefficients follow a random walk dictated by the properties of $w_t$, i.e., the state transition density function $p_w(k_{t+1}|k_t)$. This particular choice of $H$ corresponds to the optimization problem we will consider in the following section. The emission density function is given by $p_e(y_t|x_t, u_t, k_t)$. Particular choices of $p_e$ and $p_w$ emit data likelihoods concave in the parameters and hence amenable to convex optimization, a point that will be elaborated upon further in this chapter. We emphasize here that the state in the parameter evolution model refers to the current parameters $k_t$ and not the system state $x_t$ of (7.1).

The following sections will introduce a number of optimization problems with different regularization functions, corresponding to different choices of $p_w$, and different regularization arguments, corresponding to different choices of $H$. We also discuss the properties of the identification resulting from the different modeling choices. We divide our exposition into a number of cases characterized by the qualitative properties of the evolution of the parameter state $k_t$.

### Low-frequency time evolution

Many systems of practical interest exhibit slowly varying dynamics. Examples include friction varying due to temperature change or wear and tear, and electricity demand varying with season.

A slowly varying signal is characterized by *small first-order time differences*. To identify slowly varying dynamics parameters, we thus formulate an optimization problem where we in addition to penalizing the prediction errors of the model also penalize the squared 2-norm of the first-order time difference of the model parameters:

$$\underset{k}{\text{minimize}} \left\| y - \hat{y} \right\|_2^2 + \lambda^2 \sum_t \left\| k_{t+1} - k_t \right\|_2^2 \tag{7.3}$$

where $\sum_t$ denotes the sum over relevant indices $t$, in this case $t \in [1, T-1]$. This optimization problem has a closed-form solution given by

$$\tilde{k}^* = (\tilde{\mathbf{A}}^\mathsf{T} \tilde{\mathbf{A}} + \lambda^2 D_1^\mathsf{T} D_1)^{-1} \tilde{\mathbf{A}}^\mathsf{T} \tilde{Y} \tag{7.4}$$
$$\tilde{k} = \text{vec}(k_1, \dots, k_T)$$

where $\tilde{\mathbf{A}}$ and $\tilde{Y}$ are appropriately constructed matrices and the first-order differentiation operator matrix $D_1$ is constructed such that $\lambda^2 \left\| D_1 \tilde{k} \right\|_2^2$ equals the second term in (7.3). The computational complexity of computing $\tilde{k}^*$ using the closed-form solution (7.4), $\mathcal{O}\big((TK)^3\big)$ where $K = n^2 + nm$, becomes prohibitive for all but toy problems. An important observation to make to allow for an efficient method for solving (7.3) is that the cost function is the negative data log-likelihood of the Brownian random-walk parameter model (7.2) with $H = I$, which motivates us to develop a dynamic programming algorithm based on a Kalman smoother. Details on the estimation algorithms are deferred until Sec. 7.5.

A system with low-pass character is often said to have a long time constant [Åström and Murray, 2010]. For a discrete-time linear system, long time constants correspond to the dynamics matrix having eigenvalues close to the point 1 in the complex plane. The choice $H = I$ have all eigenvalues at the point 1, reinforcing the intuition of (7.3) promoting a low-frequency evolution of the dynamics. The connection between eigenvalues, small time-differences and long time constants will be explored further in Chap. 8, where inspiration from (7.3) is drawn to enhance dynamics models in the deep-learning setting.

An example of identification by solving Eq. (7.3) is provided in Sec. 7.7, where the influence of $\lambda$ is illustrated.

**Smooth time evolution**

Additional assumptions can be put on the qualitative evolution of $k_t$, such as its evolution being smooth and differentiable. A smoothly varying signal is characterized by *small second-order time differences*. To identify smoothly time-varying dynamics parameters, we thus penalize the squared 2-norm of the second-order time difference of the model parameters, and solve the optimization problem

$$\underset{k}{\text{minimize}} \left\| y - \hat{y} \right\|_2^2 + \lambda^2 \sum_t \left\| k_{t+2} - 2k_{t+1} + k_t \right\|_2^2 \tag{7.5}$$

Also this optimization problem has a closed-form solution on the form (7.4) with the corresponding second-order differentiation operator $D_2$. Equation (7.5) is the negative data log-likelihood of a Brownian random-walk parameter model with added momentum. The matrix $H$ corresponding to this model is derived in Sec. 7.4, where a Kalman smoother with augmented state is developed to find the optimal solution. We also extend problem (7.5) to more general regularization terms in Sec. 7.5.

**Piecewise constant time evolution**

Thus far, we have considered situations in which the parameter state $k_t$ evolves in a continuous fashion. In the presence of discontinuous or abrupt changes in the dynamics, estimation by solving (7.3) might perform poorly. A signal which is mostly flat, with a small number of distinct level changes, is characterized by a *sparse first-order time difference*. To encourage a solution where $k_t$ remains approximately constant most of the time, but exhibits sudden changes in dynamics at a few but unspecified number of time steps, we formulate and solve the problem

$$\underset{k}{\text{minimize}} \left\| y - \hat{y} \right\|_2^2 + \lambda \sum_t \left\| k_{t+1} - k_t \right\|_2 \tag{7.6}$$

We can give (7.6) an interpretation as a *grouped-lasso* cost function, where instead of groups being formed out of variables, our groups are defined by differences between variables. The group-lasso is a *sparsity-promoting* penalty, hence a solution in which only a small number of nonzero first-order time differences in the model parameters is favored, i.e., a piecewise constant dynamics evolution. At a first glance, one might consider the formulation

$$\underset{k}{\text{minimize}} \, \|y - \hat{y}\|_2^2 + \lambda \sum_t \|k_{t+1} - k_t\|_1 \tag{7.7}$$

which results in a dynamics evolution with sparse changes in the coefficients, but changes to different entries of $k_t$ are not necessarily occurring at the same time instants. The formulation (7.6), however, promotes a solution in which the change occurs at the same time instants for all coefficients in $A$ and $B$, i.e., $k_{t+1} = k_t$ for most $t$.

Equations (7.3) and (7.5) admitted simple interpretations as the likelihood of a dynamical model on the form (7.2). Unfortunately, Eq. (7.6) does not admit an as simple interpretation. The solution hence requires an iterative solver and is discussed in Sec. 7.A. Example usage of this optimization problem for identification is illustrated in Sec. 7.6.

### Piecewise constant time evolution with known number of steps

If the maximum number of switches in dynamics parameters, $M$, is known in advance, the optimal problem to solve is

$$\begin{aligned} \underset{k}{\text{minimize}} \qquad & \|y - \hat{y}\|_2^2 \\ \text{subject to} \qquad & \sum_t \mathbf{1}\{k_{t+1} \neq k_t\} \leq M \end{aligned} \tag{7.8}$$

where $\mathbf{1}\{\cdot\}$ is the indicator function. This problem is nonconvex and we propose solving it using dynamic programming (DP). The proposed algorithm is outlined in Sec. 7.B.

### Piecewise linear time evolution

The group-sparsity promoting effects of the group-lasso can be explored further. A piecewise linear signal is characterized by a *sparse second-order time difference*, i.e., it has a small number of changes in the slope. A piecewise linear time-evolution of the dynamics parameters is hence obtained if we solve the optimization problem

$$\underset{k}{\text{minimize}} \, \|y - \hat{y}\|_2^2 + \lambda \sum_t \|k_{t+2} - 2k_{t+1} + k_t\|_2 \tag{7.9}$$

The solution to this problem is discussed in Sec. 7.A.

### Summary

The qualitative results of solving the proposed optimization problems are summarized in Table 7.1. The table illustrates how the choice of regularizer and order of time-differentiation of the parameter vector affect the resulting solution.

**Table 7.1** Summary of optimization problem formulations. $D_n$ refers to parameter vector time-differentiation of order $n$.

| Norm | $D_n$ | Result |
|------|-------|--------|
| 1 | 1 | Small number of steps (piecewise constant) |
| 1 | 2 | Small number of bends (piecewise affine) |
| 2 | 1 | Small steps (slowly varying) |
| 2 | 2 | Small bends (smooth) |

**Two-step refinement**

Since many of the proposed formulations of the optimization problem penalize the size of the changes to the parameters in order to promote sparsity, a bias is introduced and solutions in which the changes are slightly underestimated are favored. To mitigate this issue, a two-step procedure can be implemented wherein the first step, time instances where the parameter state vector $k_t$ changes significantly are identified, we call these time instances *knots*. To identify the knots, we observe the argument inside the sum of the regularization term, i.e., $a_{t1} = \left\| k_{t+1} - k_t \right\|_2$ or $a_{t2} = \left\| k_{t+2} - 2k_{t+1} + k_t \right\|_2$. Time instances where $a_t$ is taking large values indicate suitable time indices for knots.

In the second step, the sparsity-promoting penalty is removed and equality constraints are introduced between the knots. The second step can be computed very efficiently by noticing that the problem can be split into several identical sub-problems, which each has a closed-form solution on the form

$$k^* = (\mathbf{A}^\mathsf{T}\mathbf{A} + \lambda^2 D_n^\mathsf{T} D_n)^{-1}\mathbf{A}^\mathsf{T}\tilde{Y} \tag{7.10}$$

See Sec. 6.5 for additional details on LTI-model identification.

## 7.3 Well-Posedness and Identifiability

To assess the well-posedness of the proposed identification methods, we start by noting that the problem of finding $A$ in $x_{t+1} = Ax_t$ given a pair $(x_{t+1}, x_t)$ is an ill-posed problem in the sense that the solution is non unique. If we are given several pairs $(x_{t+1}, x_t)$, for different $t$, while $A$ remains constant, the problem becomes over-determined and well-posed in the least-squares sense, provided that the vectors of state components $\{x_t^{(i)}\}_{t=1}^T$ span $\mathbb{R}^n$. The LTI-case in Sec. 6.5 is well posed according to classical results when $\mathbf{A}$ has full column rank.

When we extend our view to LTV models, the number of free parameters is increased significantly, and the corresponding regressor matrix $\tilde{\mathbf{A}}$ will never have full column rank, necessitating the introduction of a regularization term. Informally, for every $n$ measurements, the problem has $K = n^2 + nm$ free parameters. If we consider the identification problem of Eq. (7.6) and let $\lambda \to \infty$, the regularizer terms essentially become equality constraints. This will enforce a solution in which all parameters in $k$ are constant over time, and the problem reduces to the LTI-problem. As $\lambda$ decreases, the effective number of free parameters increases until the problem gets ill-posed for $\lambda = 0$. We formalize the above arguments as

PROPOSITION 1
Optimization problems (7.3) and (7.6) have unique global minima for $\lambda > 0$ if and only if the corresponding LTI optimization problem has a unique solution.     $\square$

***Proof.*** The cost function is a sum of two convex terms. For a global minimum to be nonunique, the Hessians of the two terms must have intersecting nullspaces. In the limit $\lambda \to \infty$ the problem reduces to the LTI problem. The nullspace of the regularization Hessian, which is invariant to $\lambda$, does thus not share any directions with the nullspace of $\bar{\mathbf{A}}^\mathsf{T}\bar{\mathbf{A}}$ which establishes the equivalence of identifiability between the LTI problem and the LTV problems.     $\square$

PROPOSITION 2
Optimization problems (7.5) and (7.9) with higher order differentiation in the regularization term have unique global minima for $\lambda > 0$ if and only if there does not exist any vector $v \neq 0 \in \mathbb{R}^{n+m}$ such that

$$C_t^{xu} v = \begin{bmatrix} x_t x_t^\mathsf{T} & x_t u_t^\mathsf{T} \\ u_t x_t^\mathsf{T} & u_t u_t^\mathsf{T} \end{bmatrix} v = 0 \ \forall \, t \qquad\qquad \square$$

***Proof.*** Again, the cost function is a sum of two convex terms and for a global minimum to be nonunique, the Hessians of the two terms must have intersecting nullspaces. In the limit $\lambda \to \infty$ the regularization term reduces to a linear constraint set, allowing only parameter vectors that lie along a line through time. Let $\tilde{v} \neq 0$ be such a vector, parameterized by $t$ as $\tilde{v} = \begin{bmatrix} \bar{v}^\mathsf{T} & 2\bar{v}^\mathsf{T} & \cdots & t\bar{v}^\mathsf{T} & \cdots & T\bar{v}^\mathsf{T} \end{bmatrix}^\mathsf{T} \in \mathbb{R}^{TK}$ where $\bar{v} = \mathrm{vec}(\{v\}_1^n) \in \mathbb{R}^K$ and $v$ is an arbitrary vector $\in \mathbb{R}^{n+m}$; $\tilde{v} \in \mathrm{null}\,(\bar{\mathbf{A}}^\mathsf{T}\bar{\mathbf{A}})$ implies that the loss is invariant to the perturbation $\alpha \tilde{v}$ to $\tilde{k}$ for an arbitrary $\alpha \in \mathbb{R}$. $(\bar{\mathbf{A}}^\mathsf{T}\bar{\mathbf{A}})$ is given by $\mathrm{blkdiag}(\{I_n \otimes C_t^{xu}\}_1^T)$ which means that $\tilde{v} \in \mathrm{null}\,(\bar{\mathbf{A}}^\mathsf{T}\bar{\mathbf{A}}) \iff \alpha\, t (I_n \otimes C_t^{xu})\,\bar{v} = 0 \ \forall(\alpha, t) \iff \bar{v} \in \mathrm{null}\,(I_n \otimes C_t^{xu}) \ \forall\, t$, which implies $v \in \mathrm{null}\, C_t^{xu}$ due to the block-diagonal nature of $I_n \otimes C_t^{xu}$.     $\square$

REMARK 2
If we restrict our view to constant systems with stationary Gaussian inputs with covariance $\Sigma_u$, we have as $T \to \infty$, $(1/T)\sum xx^\mathsf{T}$ approaching the stationary controllability Gramian given by the solution to $\Sigma = A\Sigma A^\mathsf{T} + B\Sigma_u B^\mathsf{T}$. Not surprisingly, the well-posedness of the optimization problem is thus linked to the excitation of the system modes through the controllability of the system.     $\square$

For the LTI problem to be well-posed, the system must be identifiable and the input $u$ must be persistently exciting of sufficient order [Johansson, 1993].

## 7.4 Kalman Smoother for Identification

We now elaborate on the connection between the proposed optimization problems and the dynamical system governing the evolution of the parameter state $k_t$ (7.2).

We note that (7.3) is the negative log-likelihood of the dynamics model (7.2) with state vector $k_t$. The identification problem is thus reduced to a standard state-estimation problem.

To develop a Kalman smoother-based algorithm for solving (7.5), we augment the model (7.2) with the state variable $k'_t = k_t - k_{t-1}$ and note that $k'_{t+1} - k'_t = k_{t+1} - 2k_t + k_{t-1}$. We thus introduce the augmented-state model

$$\begin{bmatrix} k_{t+1} \\ k'_{t+1} \end{bmatrix} = \begin{bmatrix} I_K & I_K \\ 0_K & I_K \end{bmatrix} \begin{bmatrix} k_t \\ k'_t \end{bmatrix} + \begin{bmatrix} 0 \\ w_t \end{bmatrix} \tag{7.11}$$

$$y_t = \begin{bmatrix} \left( I_n \otimes \begin{bmatrix} x_t^\mathsf{T} & u_t^\mathsf{T} \end{bmatrix} \right) & 0 \end{bmatrix} \begin{bmatrix} k_t \\ k'_t \end{bmatrix} \tag{7.12}$$

which is on a form suitable for filtering/smoothing with the machinery developed in Sec. 4.3.

*General case*   The Kalman smoother-based identification method can be generalized to solving optimization problems where the argument in the regularizer appearing in (7.5) is replaced by a general linear operation on the parameter vector, $P(z)k$

$$\underset{k}{\text{minimize}} \, \|y - \hat{y}\|_2^2 + \lambda^2 \sum_t \|P(z)k_t\|_2^2 \tag{7.13}$$

where $P(z)$ is a polynomial of degree $n > 0$ in the time-difference operator $z$. For (7.5), $P(z)$ equals $z^2 - 2z + 1$ and $P^{-1}(z)$ has a realization on the form (7.11).

## 7.5 Dynamics Priors

As discussed in Sec. 7.3, the identifiability of the parameters in a dynamics model hinges on the observability of the dynamical system (7.2), or more explicitly, only modes excited by the input $u$ will be satisfactorily identified. In many practical scenarios, such as if the identification is part of an iterative learning and control scheme, e.g., Iterative Learning Control (ILC) or reinforcement learning, it might be undesirable to introduce excessive noise in the input to improve excitation for identification. This section will introduce additional information about the dynamics in the form of priors, which mitigate the issue of poor excitation of the system modes. The prior information might come from, e.g., a nominal model known to be inaccurate, or an estimated global model such as a Gaussian mixture model (GMM). A statistical model of the joint density $p(x_{t+1}, x_t, u_t)$ constructed from previously collected tuples $(x_{t+1}, x_t, u_t)$, for instance, provides a dynamical model of the system through the conditional probability-density function $p(x_{t+1}|x_t, u_t)$.

We will see that for priors from certain families, the resulting optimization problem remains convex. For the special case of a Gaussian prior over the dynamics parameters or the output, the posterior mean of the parameter vector is once again conveniently obtained from a Kalman-smoothing algorithm, modified to include the prior.

## General case

A general prior over the parameter-state variables $k_t$ can be specified as $p(k_t|z_t)$, where the variable $z_t$ is a placeholder for whatever signal might be of relevance, for instance, the time index $t$ or state $x_t$. The data log-likelihood of (7.2) with the prior $p(k_t|z_t)$ added takes the form

$$\log p(k, y|x, z)_{1:T} = \sum_{t=1}^{T} \log p(y_t|k_t, x_t) + \sum_{t=1}^{T-1} \log p(k_{t+1}|k_t) + \sum_{t=1}^{T} \log p(k_t|z_t) \text{ (7.14)}$$

which factors conveniently due to the Markov property of a state-space model. For particular choices of density functions in (7.14), notably Gaussian and Laplacian, the negative log-likelihood function becomes convex. The next section will elaborate on the Gaussian case and introduce a recursive algorithm that efficiently solves for the full posterior. The Laplacian case, while convex, does not admit an equally efficient algorithm. The Laplacian likelihood is, however, more robust to outliers in the data, making the trade-off worth consideration.

## Gaussian case

If all densities in (7.14) are Gaussian and $k$ is modeled with the Brownian random walk model (7.2) (Gaussian $v_t$), (7.14) can be written on the form (scaling constants omitted)

$$-\log p(k, y|x, z)_{1:T} = \sum_{t=1}^{T} \left\| y_t - \hat{y}(k_t, x_t) \right\|_{\Sigma_e^{-1}}^2$$
$$+ \sum_{t=1}^{T-1} \left\| k_{t+1} - k_t \right\|_{\Sigma_w^{-1}}^2 + \sum_{t=1}^{T} \left\| \mu_0(z_t) - k_t \right\|_{\Sigma_0^{-1}(z_t)}^2 \qquad (7.15)$$

for some function $\mu_0(z_t)$ which produces the prior mean of $k_t$ given $z_t$. $\Sigma_e, \Sigma_w, \Sigma_0(z_t)$ are the covariance matrices of the measurement noise, parameter drift and prior respectively and $\left\| x \right\|_{\Sigma^{-1}}^2 = x^\mathsf{T} \Sigma^{-1} x$.

For this special case, we devise a modified Kalman-smoothing algorithm, where the conditional mean of the state is updated with the prior, to solve the estimation problem. The standard Kalman-filtering equations for a general linear system are derived in Sec. 4.3—the modification required to incorporate a Gaussian prior on the state variable $p(x_t|z_t) = \mathcal{N}(\mu_0(z_t), \Sigma_0(z_t))$ involves a repeated correction step and takes the form

$$\bar{K}_t = P_{t|t}\big(P_{t|t} + \Sigma_0(z_t)\big)^{-1} \tag{7.16}$$

$$\bar{x}_{t|t} = \hat{x}_{t|t} + \bar{K}_t\big(\mu_0(z_t) - \hat{x}_{t|t}\big) \tag{7.17}$$

$$\bar{P}_{t|t} = P_{t|t} - \bar{K}_t P_{t|t} \tag{7.18}$$

where $\bar{\cdot}$ denotes the posterior value. This additional correction can be interpreted as receiving a second measurement $\mu_0(z_t)$ with covariance $\Sigma_0(z_t)$. For the Kalman-smoothing algorithm, $\hat{x}_{t|t}$ and $P_{t|t}$ in (7.17) and (7.18) are replaced with $\hat{x}_{t|T}$ and $P_{t|T}$.

A prior over the output of the system, or a subset thereof, is straightforward to include in the estimation by means of an extra update step, with $C, R_2$ and $y$ being replaced with their appropriate values according to the prior.

We remark that although all optimization problems proposed in this chapter could be solved by generic solvers, the statistical interpretation and solution provided by the Kalman-smoothing algorithm provide us not only with the optimal solution, but also an uncertainty estimate. The covariance matrix of the state, Eq. (4.24), is an estimate of the covariance of the parameter vector for each time step. This uncertainty estimate may be of importance in downstream tasks using the estimated model. One such example is trajectory optimization, where it is useful to limit the step length based on the uncertainty in the model. We will make use of this when performing trajectory optimization in a reinforcement-learning setting in Chap. 11, where model uncertainties are taken into account by enforcing a constraint on the Kullback-Leibler divergence between two consecutive trajectory distributions.

## 7.6   Example—Jump-Linear System

To demonstrate how the proposed optimization problems may be used for identification, we now consider a simulated example. We generate a state sequence from the following LTV system, where the dynamics change from

$$A_t = \left[\begin{array}{cc} 0.95 & 0.1 \\ 0.0 & 0.95 \end{array}\right], \quad B_t = \left[\begin{array}{c} 0.2 \\ 1.0 \end{array}\right]$$

to

$$A_t = \left[\begin{array}{cc} 0.5 & 0.05 \\ 0.0 & 0.5 \end{array}\right], \quad B_t = \left[\begin{array}{c} 0.2 \\ 1.0 \end{array}\right]$$

at $t = 200$. The input was Gaussian noise of zero mean and unit variance, state transition noise and measurement noise ($y_t = x_{t+1} + e_t$) of zero mean and $\sigma_e = 0.2$ were added. In this problem the parameters change abruptly, a suitable choice of identification algorithm is thus (7.6). Figure 7.1 depicts the estimated coefficients in the dynamics matrices after solving (7.6), for a value of $\lambda$ chosen using the L-curve method [Hansen, 1994]. The figure indicates that the algorithm correctly identifies the abrupt change in the system parameters at $t = 200$ and maintains an otherwise near constant parameter vector. This example highlights how the sparsity-promoting penalty can be used to indicate whether or not something has abruptly changed the dynamics, without specifying the number of such changes a priori. The methods briefly discussed in Sec. 7.2 can be utilized, should it be desirable to have two separate LTI-models describing the system.

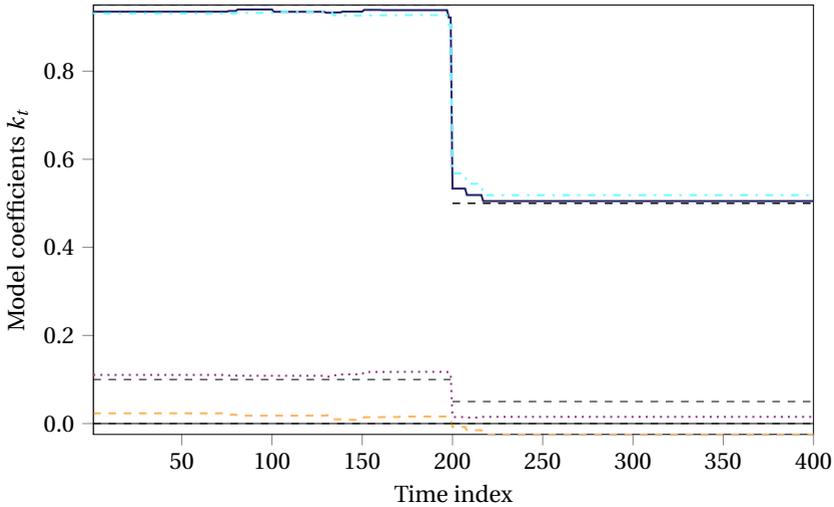

**Figure 7.1** Piecewise constant state-space dynamics. True values are shown with dashed, black lines. Gaussian state-transition and measurement noise with $\sigma = 0.2$ were added. At $t = 200$ the dynamics of the system change abruptly. A suitable choice of regularization term $\left( \lambda \left\| k^+ - k \right\|_2 \right)$ allows us to estimate a dynamics model that exhibit an abrupt change in the coefficients, without specifying the number of such changes a priori. Please note that this figure shows the coefficients of $k$ corresponding the $A$-matrix of Eq. (7.1) only.

## 7.7 Example—Low-Frequency Evolution

In this example, we simulate the system (7.1)-(7.2) with $H = I$ to generate an LTV system with low-frequency time evolution of the dynamics. We let the state $x \in \mathbb{R}^3$ drift with isotropic covariance of $\sigma_v^2 = 0.01^2$, and let the parameters $k$ drift with isotropic covariance of $\sigma_w^2 = 0.001^2$. The input $u \in \mathbb{R}^2$ was Gaussian with isotropic covariance 1. The evolution of the true parameters in the $A$-matrix is shown in black in Fig. 7.2.

We estimate LTV models by solving Eq. (7.3) (regularization term $\lambda \left\| k^+ - k \right\|_2^2$)

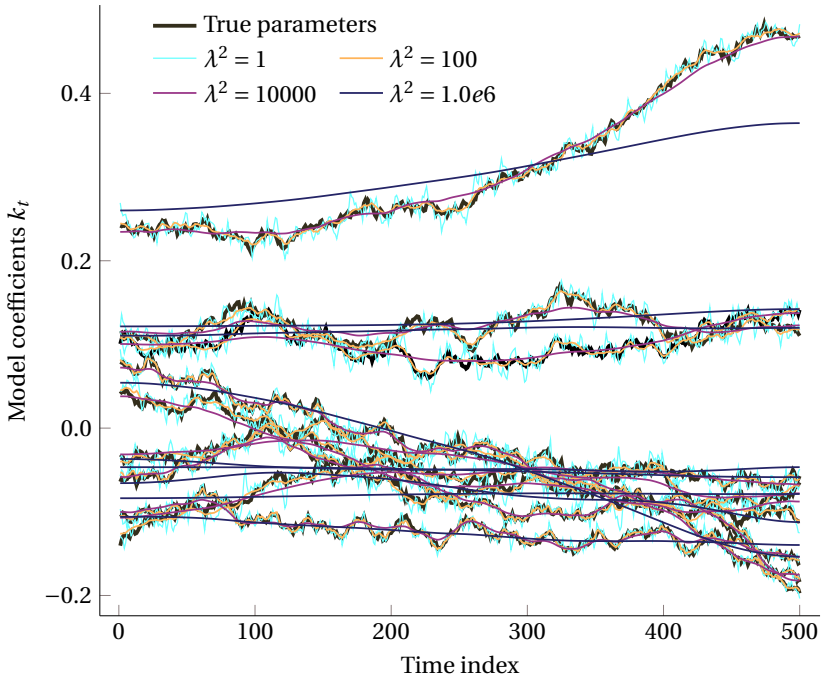

**Figure 7.2** Trajectories of the system (7.2) (black) together with models estimated by solving (7.3) with varying values of $\lambda$. The smoothing effect of a high $\lambda$ is illustrated, and as $\lambda \to \infty$, the estimated model converges to an LTI model. The optimal value is given by $\lambda \approx 10$. Please note that this figure shows the coefficients of $k$ corresponding the $A$-matrix of (7.1) only. The coefficients of the $B$-matrix evolve similarly but are omitted for clarity.

using the Kalman-smoothing algorithm and 4 different values of $\lambda$, shown in color in Fig. 7.2. The optimal value of $\lambda$ is approximately given by $\lambda = \sigma_v / \sigma_w = 10$. In Fig. 7.2 we see how a too small value of the regularization parameter leads to noisy estimates of the time-varying parameters, whereas a too high value leads to overly conservative changes. Arguably, the optimal value of $\lambda = 10$ performs best. In Fig. 7.3, we illustrate the log-distributions of prediction errors in the top-left pane, and model errors in the top-right pane, in both cases on the training data. It is clear that the prediction error is an increasing function of the regularization parameter, as expected. However, the model error, calculated as the sum of squared differences between the true system parameters and the estimated parameters, is minimized by the optimal value of $\lambda$. In reality, this optimal value is unknown and determining it is not always easy. In the bottom four panes, we show quantile-quantile plots [Wilk and Gnanadesikan, 1968] of the prediction errors. The optimal value for $\lambda$ produces normal residuals, whereas other choices for $\lambda$ produce heavy-tailed distributions. While this analysis is available even when the true system is not known, it might produce less clear outcomes when the data is not generated by a system included in the model set being searched over. Alternative ways of setting values for $\lambda$ include cross validation and maximum-likelihood estimation under the statistical model (7.1)-(7.2), illustrated in Fig. 7.4. The likelihood of the data given a model on the form (7.2) is easily calculated during the forward-pass of the Kalman algorithm.

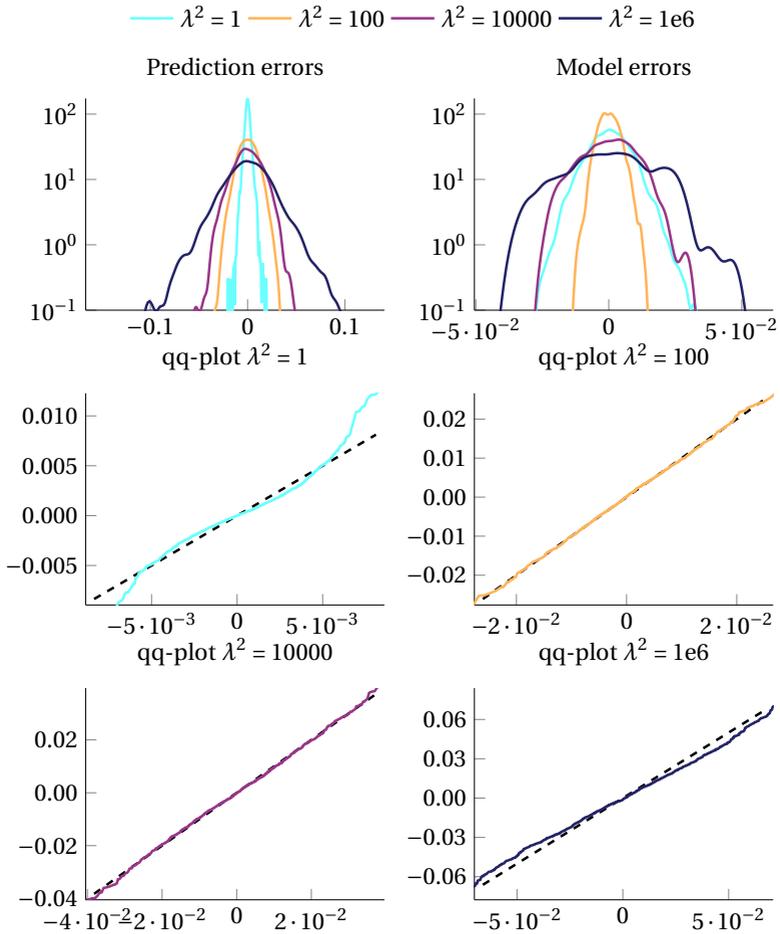

**Figure 7.3**  The top panes show (log) error distributions on training data when solving Eq. (7.3) on data generated by (7.1)-(7.2). The prediction error is a strictly increasing function of $\lambda$, whereas the model error is minimized by the optimal value for $\lambda$. The bottom four panes show quantile-quantile plots of prediction errors on the training data, for the different choices of $\lambda$.

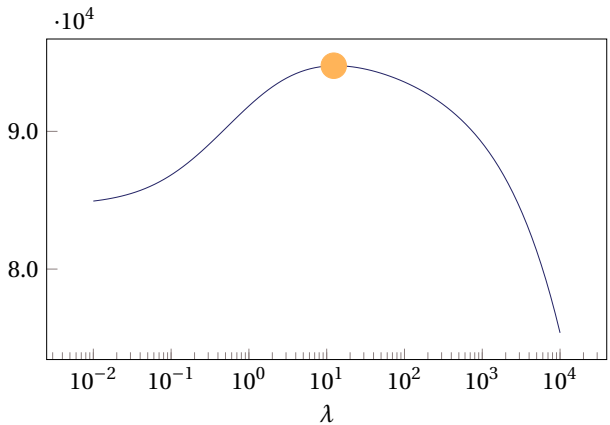

**Figure 7.4** Maximum-likelihood estimation to determine a suitable value of $\lambda$. The curve illustrates the likelihood under the model (7.1)-(7.2) for different choices of $\lambda$, the maximum is marked with a dot.

Yet another option for determining the value of $\lambda$ is to consider it as a relative time-constant between the evolution of the state and the evolution of the model parameters. A figure like Fig. 7.2, together with prior knowledge of the system, is often useful in determining $\lambda$.

## 7.8 Example—Nonsmooth Robot Arm with Stiff Contact

To illustrate the ability of the proposed models to represent the nonsmooth dynamics along a trajectory of a robot arm, we simulate a two-link robot with discontinuous Coulomb friction. We also let the robot establish a stiff contact with the environment to illustrate both strengths and weaknesses of the modeling approach.

The state of the robot arm consists of two joint coordinates, $q$, and their time derivatives, $\dot{q}$. The control signal trajectory was computed using an inverse dynamics model of the robot, and Gaussian noise was superimposed the computed torque trajectory. Figure 7.5 illustrates the state trajectories, control torques and simulations of a model estimated by solving (7.6). The figure clearly illustrates that the model is able to capture the dynamics both during the nonsmooth sign change of the velocity, and also during the establishment of the stiff contact. The learned dynamics of the contact is, however, time-dependent. This time-dependence is, in some situations, a drawback of LTV-models. This drawback is illustrated in Fig. 7.6, where the model is used on a validation trajectory where a different noise sequence was added to the control torque. Due to the novel input signal, the contact is established at a different time-instance and as a consequence, there is an error transient in the simulated data.

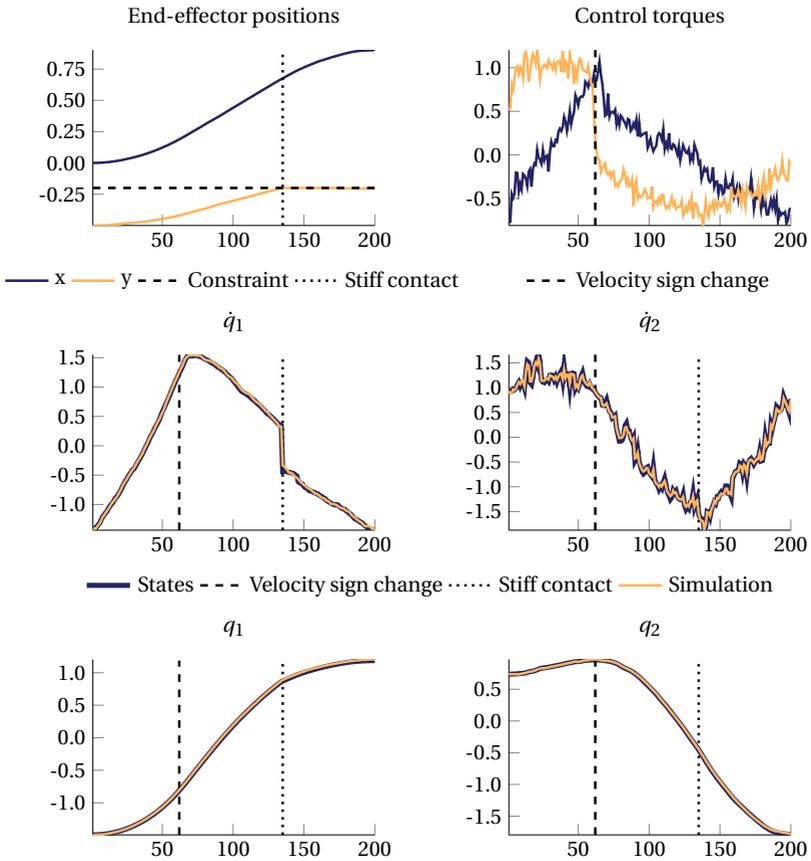

**Figure 7.5** Simulation of nonsmooth robot dynamics with stiff contact—training data vs. sample time index. A sign change in velocity, and hence a discontinuous change in friction torque, occurs in the time interval 50-100 and the contact is established in the time interval 100-150. For numerical stability, all time-series are normalized to zero mean and unit variance, hence, the original velocity zero crossing is explicitly marked with a dashed line. The control signal plot clearly indicates the discontinuity in torque around the unnormalized zero crossing of $\dot{q}_2$.

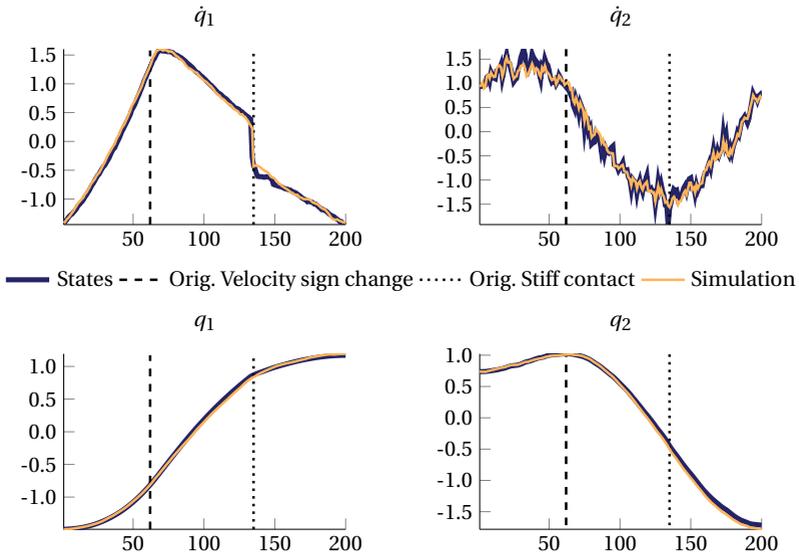

**Figure 7.6** Simulation of nonsmooth robot dynamics with stiff contact—validation data vs. sample time index. The dashed lines indicate the event times as occurring for the original *training data*, highlighting that the model is able to deal effortlessly with the nonsmooth friction, but inaccurately predicts the time evolution around the contact event, which now occurs at a slightly different time instance.

## 7.9   Discussion

This chapter presents methods for estimation of linear, time-varying models. The methods presented extend directly to nonlinear models that remain *linear in the parameters*.

   When estimating an LTV model from a trajectory obtained from a nonlinear system, one is effectively estimating the linearization of the system around that trajectory. A first-order approximation to a nonlinear system is not guaranteed to generalize well as deviations from the trajectory become large. Many nonlinear systems are, however, approximately *locally* linear, such that they are well described by a linear model in a small neighborhood around the linearization/operating point. For certain methods, such as iterative learning control and trajectory centric reinforcement learning, a first-order approximation to the dynamics is used for efficient optimization, while the validity of the approximation is ensured by incorporating penalties or constraints on the deviation between two consecutive trajectories [Levine and Koltun, 2013]. We explore this concept further using the methods proposed in this chapter, in Chap. 11.

The methods presented allow very efficient learning of this first-order approximation due to the postulated prior belief over the nature of the change in dynamics parameters, encoded by the regularization terms. Prior knowledge encoded this way puts less demand on the data required for successful identification. The identification process will thus not be as invasive as when excessive noise is added to the input for identification purposes, allowing learning of flexible, overparameterized models that fit available data well. This makes the proposed identification methods attractive in applications such as guided policy search (GPS) [Levine and Koltun, 2013; Levine et al., 2015] and nonlinear iterative learning control (ILC) [Bristow et al., 2006], where they can lead to dramatically decreased sample complexity.

The proposed methods that lend themselves to estimation through the Kalman smoother-based algorithm could find use as a layer in a neural network. Amos and Kolter (2017) showed that the solution to a quadratic program is differentiable and can be incorporated as a layer in a deep-learning model. The forward pass through such a network involves solving the optimization problem, making the proposed methods attractive due to the $\mathcal{O}(T)$ solution.

When faced with a system where time-varying dynamics is suspected and no particular knowledge regarding the dynamics evolution is available, or when the dynamics are known to vary slowly, a reasonable first choice of algorithm is (7.5). This algorithm is also, by far, the fastest of the proposed methods due to the Kalman-smoother implementation of Sec. 7.5.[1] As a consequence of the ease of solution, finding a good value for the regularization parameter $\lambda$ is also significantly easier. Example use cases include when dynamics are changing with a continuous auxiliary variable, such as temperature, altitude or velocity. If a smooth parameter drift is found to correlate with an auxiliary variable, LPV-methodology can be employed to model the dependency explicitly, something that will not be elaborated upon further in this thesis but was implemented and tested in [*BasisFunctionExpansions.jl*, B.C., 2016].

Dynamics may change abruptly as a result of, e.g., system failure, change of operating mode, or when a sudden disturbance enters the system, such as a policy change affecting a market or a window opening, affecting the indoor temperature. The identification method (7.6) can be employed to identify when such changes

---

[1] The Kalman-smoother implementation is often *several* orders of magnitude faster than solving the optimization problems with an iterative solver.

occur, without specifying a priori how many changes are expected.

For simplicity, the regularization weights were kept as simple scalars in this chapter. However, all terms $\lambda \left\| \Delta k \right\|_2^2 = (\Delta k)^\mathsf{T}(\lambda I)(\Delta k)$ can be generalized to $(\Delta k)^\mathsf{T} \Lambda (\Delta k)$, where $\Lambda$ is an arbitrary positive definite matrix. This allows incorporation of different scales for different variables with little added implementation complexity. Known structure, such as sparsity patterns in the matrices $A$ and $B$, is easily incorporated into $\Lambda$ or the covariance matrix of $w$ for the Kalman-smoother solver.

**Measurement-noise model**

The identification algorithms developed in this chapter were made available by the simple nature of the dynamical model Eq. (7.1). In practice, measurements are often corrupted by noise, in particular if parts of the state-sequence is derived from time differentiation of measured quantities. We will not cover the topic of noise-model estimation in depth here, but will provide a few suggested treatments from the literature that could be considered in a scenario with poor signal-to-noise ratio.

A very general approach to estimation of noise models is pseudo-linear regression [Ljung and Söderström, 1983; Ljung, 1987]. The general idea is to estimate the noise components and include them in the model. In the present context, this could amount to estimating a model using any of the methods described above, calculate the model residuals $e_t = y_t - \hat{y}_t$, and build a model $\hat{e}_t = \rho(e_{t-1}, e_{t-2}, ...)$.

The combined problem of estimating both states $x$ and parameters $k$ can be cast as a nonlinear filtering problem [Ljung and Söderström, 1983]. The nonlinear nature of the resulting problem necessitates a nonlinear filtering approach, such as the extended Kalman filter [Ljung and Söderström, 1983] or the particle filter (see Sec. 4.2). The literature on *iterated filtering* [Ionides et al., 2006; Lindström et al., 2012] considers this nonlinear filtering problem in the context of constant dynamics.

## 7.10 Conclusions

We have proposed a framework for identification of linear, time-varying models using convex optimization. We showed how a Kalman smoother can be used to estimate the dynamics efficiently in a few special cases, and demonstrated the use of the proposed LTV models on two examples, highlighting their efficiency for jump-linear system identification and learning the linearization of a nonlinear system along a trajectory. We further demonstrated the ability of the models to handle nonsmooth friction dynamics as well as analyzed the identifiability of the models.

Implementations of all discussed methods are made available in [*LTVModels.jl*, B.C., 2017].

## Appendix  A.  Solving (7.6)

Due to the nonsquared norm penalty $\sum_t \left\| k_{t+1} - k_t \right\|_2$, the problems (7.6) and (7.9) are significantly harder to solve than (7.3). An efficient implementation using the linearized ADMM algorithm [Parikh and Boyd, 2014] is made available in the accompanying repository.

The linearized ADMM algorithm [Parikh and Boyd, 2014] solves the problem

$$\underset{k}{\text{minimize}} f(k) + g(Ak) \tag{7.19}$$

where $f$ and $g$ are convex functions and $A$ is a matrix. The optimization problems with the group-lasso penalty (7.6) and (7.9) can be written on the form (7.19) by constructing $A$ such that it performs the computations $k_{t+1} - k_t$ or $k_{t+2} - 2k_{t+1} + k_t$ and letting $g$ be $\left\| \cdot \right\|_2$.

## Appendix  B.  Solving (7.8)

The optimization problem of Eq. (7.8) is nonconvex and harder to solve than the other problems proposed in this chapter. To solve small-scale instances of the problem, we modify the algorithm developed in [Bellman, 1961], an algorithm frequently referred to as segmented least-squares [Bellman and Roth, 1969]. Bellman (1961) approximates a curve by piecewise linear segments. We instead associate each segment (set of consecutive time indices during which the parameters are constant) with a dynamics model, as opposed to a simple straight line.[2]

The algorithm relies on the key fact that the value function for a sequential optimization problem with quadratic cost and parameters entering linearly, is quadratic. This allows us to find the optimal solution in $\mathcal{O}(T^2)$ time instead of the

$$\mathcal{O}\left(\binom{T}{M}\right)$$

complexity of the naive solution. A simplified implementation is provided in Algorithm 3.[3]

Unfortunately, the computational complexity of the dynamic-programming solution, $\mathcal{O}(T^2 K^3 M)$, becomes prohibitive for large $T$, in which case an approximate solution can be obtained by, e.g., solving (7.6) and then projecting the solution onto the constraint set by only keeping the $M$ largest parameter changes.

---

[2] Indeed, if a simple integrator is chosen as dynamics model and a constant input is assumed, the result of our extended algorithm reduces to the segmented least-squares solution.

[3] The help of Pontus Giselsson in developing this algorithm is gratefully acknowledged.

**Algorithm 3** Simple dynamic-programming solver without memoization. Efficient solver provided in [*LTVModels.jl*, B.C., 2017]. The algorithm can be used for the original purpose of [Bellman, 1961] by letting `costfun(y,a,b)` and `argmin(y,a,b)` calculate the cost and optimal estimate of an affine approximation to *y* between *a* and *b*. By instead letting these functions calculate the optimal cost and solution to the LTI identification problem of Sec. 6.5, we obtain the optimal solution to (7.8). Return values *V*, *t*, *a* constitute the value function, breakpoints and optimal parameters, respectively.

```
function seg_bellman(y,M)
    T  = length(y)
    B  = zeros(Int, M-1, T) # back-pointer matrix
    fi = [costfun(y,j,T) for j = 1:T] # initialize Bellman iteration
    # Bellman iteration
    fnext = Vector{Float64}(T)
    for j = M-1:-1:1
        for k = j:T-(M-j)
            opt, optl = Inf, 0
            for l = k+1:T-(M-j-1)
                cost = costfun(y,k,l-1) + fi[l]
                if cost < opt; opt, optl = cost, l; end
            end
            fnext[k] = opt
            B[j,k]   = optl-1
        end
        fi .= fnext
    end
    V = [costfun(y,1,j)+fi[j+1] for j = 1:T-M] # last Bellman iterate
    # Backward pass
    t = Vector{Int}(M)
    a = Vector{typeof(argmin(y,1,2))}(M+1)
    _,t[1] = findmin(V[1:end-M]) # t = index of minimum
    a[1]   = argmin(y,1,t[1])
    for j = 2:M
        t[j] = B[j-1,t[j-1]]
        a[j] = argmin(y,t[j-1]+1,t[j])
    end
    a[M+1] = argmin(y,t[M]+1,T)
    return V,t,a
end
```

# 8

# Identification and Regularization of Nonlinear Black-Box Models

## 8.1 Introduction

Dynamical control systems are often described in continuous time by differential state equations on the form

$$\dot{x}(t) = f_c\big(x(t), u(t)\big) \tag{8.1}$$

where $x$ is a Markovian state vector, $u$ is the input and $f_c$ is a function that maps the current state and input to the state time derivative. An example of such a model is a rigid-body dynamical model of a robot

$$\ddot{q} = -M^{-1}(q)\big(C(q, \dot{q})\dot{q} + G(q) + F(\dot{q}) - u\big), \quad x = \begin{bmatrix} q \\ \dot{q} \end{bmatrix} \tag{8.2}$$

where $M, C, G$ and $F$ model phenomena such as inertia, Coriolis, gravity and friction [Spong et al., 2006] and $q$ are the joint coordinates.

In the discrete time domain, we often consider models on the form

$$x_{t+1} = f\big(x_t, u_t\big) \tag{8.3}$$

where $f$ is a function that maps the current state and input to the state at the next time-instance [Åström and Wittenmark, 2013a]. We have previously discussed the case where $f$ is a linear function of the state and the control input, and we now extend our view to nonlinear functions $f$.

Learning a globally valid dynamics model $\hat{f}$ of an arbitrary nonlinear system $f$ with little or no prior information is a challenging problem. Although in principle, any sufficiently complex function approximator, such as a deep neural network (DNN), could be employed, high demands are put on the amount of data required to prevent overfitting and to obtain a faithful representation of the dynamics over the entire state space. If prior knowledge is available, it can often be used to reduce the demands on the amount of data required to learn an accurate model [Sjöberg et al., 1995].

Early efforts in nonlinear modeling include Volterra-Wiener models that make use of basis-function expansions to model nonlinearities [Johansson, 1993]. This type of models exhibit several drawbacks and are seldom used in practice, one of which is the difficulty of incorporating prior knowledge into the model. Oftentimes, this can also be hard to incorporate into a flexible black-box model such as a deep neural network. This chapter will highlight a few general attempts at doing so, compatible with a wide class of function approximators.

In many applications, the linearization of $f$ is important, a typical example being linear control design [Glad and Ljung, 2014]. In applications such as iterative learning control (ILC) [Bristow et al., 2006] and trajectory centric, episode-based reinforcement learning (TCRL) [Levine and Koltun, 2013], the linearization of the nonlinear dynamics along a trajectory is often needed for optimization. Identification of $f$ must thus not only yield a good model for prediction/simulation, but also the Jacobian $J_{\hat{f}}$ of $\hat{f}$ must be close to the true system Jacobian $J$. The linearization of a model around a trajectory returns a Linear Time-Varying (LTV) model on the form

$$x_{t+1} = A_t x_t + B_t u_t$$

where the matrices $A$ and $B$ constitute the output Jacobian. This kind of model was learned efficiently using dynamic programming in Chap. 7. However, not all situations allow for accurate learning of an LTV model around a trajectory. A potential problem that can arise is insufficient excitation provided by the control input [Johansson, 1993]. Prior knowledge regarding the evolution of the dynamics, encoded in form of carefully designed regularization, was utilized in Chap. 7 in order to obtain a well-posed optimization problem and a meaningful result. While this proved to work well in many circumstances, it might fail if excitation is small relative to how fast the dynamics changes along a trajectory. When model identification is a subtask in an outer algorithm that optimizes a control-signal trajectory or a feedback policy, adding excessive noise for identification purposes may be undesirable, making regularization solely over time as in Chap. 7 insufficient. A step up in sophistication from LTV models is to learn a nonlinear model that is valid globally. A nonlinear model can be learned from several consecutive trajectories and is thus able to incorporate more data than an LTV model. Unfortunately, a black-box nonlinear model also *requires* more data to learn a high fidelity model and not suffer from overfitting.

The discussion so far indicates two issues; 1) Complex nonlinear models have the potential to be valid globally, but may suffer from overfitting and thus not learn a function that generalizes and learns the correct linearization. 2) LTV models can be learned efficiently and can represent the linearized dynamics well, but require sufficient excitation, are time-based and valid only locally.

In this chapter, we draw inspiration from the regularization methods detailed in Chap. 7 for learning of a general, nonlinear black-box model, $\hat{f}$. Since an LTV model lives in the tangent-space of a nonlinear model, we call the procedure tangent-space regularization. In this work, we will model $\hat{f}$ using a deep neural network (DNN), a class of models which is well suited to learn arbitrarily complex static mappings. We will explore how modern techniques for training of DNNs interact with the particular structure chosen to represent $\hat{f}$ and compare this regularization to a common choice in the literature, $L_2$ regularization or *weight decay*. We investigate how weight decay affects the fidelity and eigenvalue spectrum of the Jacobian of the learned model and reason about this from a control-theoretical viewpoint.

We proceed to introduce the problem of learning a dynamics model $\hat{f}$ in Sec. 8.3. We then discuss the influence of weight decay on different formulations of the learning problem and introduce tangent-space regularization formally in Sec. 8.5. Finally, we conduct numerical evaluations.

## 8.2 Computational Aspects

A problem very much related to that of simulation error minimization discussed in Chap. 3, long considered insurmountable, is that of training deep neural networks. It was believed that the propagation of gradients through deep models was either too sensitive and unstable, or would inevitably lead to either exploding or vanishing gradients. Many incremental steps have been taken recently to mitigate this problem, allowing training of ever deeper models. A major contribution was the notion of residual connections [He et al., 2015]. A residual connection is a simple idea: to a complicated function, add a parallel identity $y = f(x) = g(x) + x$. The function $g$ is called a *residual* function, acting around the *nominal* identity function $x$, where $x$ is sometimes called a skip connection. Looking at the Jacobian of these two functions, the impact of the modeling difference is obvious:

$$\nabla_x y = \nabla_x f(x) = \nabla_x g(x) + I \tag{8.4}$$

no matter the complicated nature of $g$, the Jacobian of the output with respect to the input will contain the identity component. This allows gradients to flow effortlessly through deep architectures composed of stacked residual units, making training of models as deep as 1000 layers possible [He et al., 2015]. In many cases, learning a function $g$ around the identity, is vastly easier than learning the full function $f$. Nguyen et al. (2018) even proved that under certain circumstances and with enough skip connections to the output layer, a DNN has no local minima and a continuous path of decreasing loss exists from any starting point to the global optimum.

Consider what it takes for a network with one hidden layer to learn the identity mapping. If we use tanh as the hidden layer activation function, the incoming weights must be small to make sure the tanh is operating in its linear region, while the outgoing weights have to be the reciprocal of the incoming. The number of neurons required is the same as the input dimension. For an activation function that does not have a near-linear region around zero, such as the relu or the sigmoid functions [Goodfellow et al., 2016], help from the bias term is further required to center the activation in the linear region. For a deep network, this has to happen for every layer.

While a deep model is perfectly capable of learning the identity mapping, it is needless to say that incorporating this mapping explicitly can sometimes be beneficial. In particular, if the input and output of the function live in the same domain, e.g., image in—image out, state in—state out, it is often easier to learn the residual around the identity. To quote Sjöberg et al. (1995), "Even if nonlinear structures are to be applied there is no reason to waste parameters to estimate facts that are already known". The identity can in this case be considered a nominal model. Whenever available, simple nominal models provide excellent starting points, and modeling and learning the residuals of such a model may be easier than learning the combined effect of the nominal model and residual effects.

While we argue for the use of prior knowledge where available, in particular when modeling physical systems where identification data can be hard to acquire, we would also like to offer a counter example highlighting the need to do so wisely. Frederick Jelinek, a researcher in natural language processing, famously said[1] "Every time I fire a linguist, the performance of the speech recognizer goes up". The quote indicates that the data—natural language as used by people—did not follow the grammatical and syntactical rules laid down by the linguists. Prior knowledge might in this case have introduced severe bias that held the performance of the model back. Natural language processing is a domain in which data is often easy to obtain and plentiful enough to allow fitting of very flexible models without negative consequences.

A different notion of residual connection is that found in a form of recurrent neural network called a Long Short-Term Memory (LSTM) network [Hochreiter and Schmidhuber, 1997]. While the *resnet* [He et al., 2015] was deep in the sense of multiple consecutive layers, an RNN is deep in the sense that a function is applied recursively in time. LSTMs were invented to mitigate the issue of vanishing/exploding gradients while backpropagating through time to train recurrent neural networks. When a model, or more generally a function $f$ with Jacobian $\nabla_x f(x) = J(x)$, is applied to a state recursively, the Jacobian of the recursive mapping grows as $\nabla_x f^{(n)}(x) \sim J^n(x)$, where $f^{(n)}(x)$ denotes the $n$ times recursive application of $f$, $f(...f(f(x)))$. Any eigenvalues of $J$ greater than 1 will grow exponentially—exploding gradients—and eigenvalues smaller than 1 will decay exponentially—vanishing gradients. If we model $f$ as

$$f(x) = g(x) + x \tag{8.5}$$

the Jacobian will be the identity plus some small deviation, effectively helping the eigenvalues of $J$ stay close to 1. Slightly simplified, LSTMs effectively propagate the state with an identity function according to (8.5).

---

[1] The exact wording and circumstances around this quote constitute the topic of a very lengthy footnote on the Wikipedia page of Jelinek `https://en.wikipedia.org/wiki/Frederick_Jelinek`.

Other efforts at mitigating issues with the training of deep models include careful weight initialization. If, for instance, all weights are initialized to be unitary matrices, the norm of vectors propagated through the network stays constant. The initialization of the network can further be used to our advantage. Prior knowledge of, e.g., resonances etc. can be encoded into the initial model weights. In this chapter, we will explore this concept further, along with the effects of identity connections, and motivate them from a control-theoretic perspective.

## 8.3 Estimating a Nonlinear Black-Box Model

To frame the learning problem, we let the dynamics of a system be described by a neural network $\hat{f}$ to be fitted to input-output data $\tau = \{x_t, u_t\}_{t=1}^T$ according to

LEARNING OBJECTIVE 1

$$x_{t+1} = \hat{f}(x_t, u_t), \quad f : \mathbb{R}^n \times \mathbb{R}^m \mapsto \mathbb{R}^n \qquad \square$$

which we will frequently write on the form $x^+ = \hat{f}(x, u)$ by omitting the time index $t$ and letting $\cdot^+$ indicate $\cdot_{t+1}$. We further consider the linearization of $\hat{f}$ around a trajectory $\tau$

$$
\begin{aligned}
x_{t+1} &= A_t x_t + B_t u_t \\
k_t &= \text{vec}(\begin{bmatrix} A_t^\mathsf{T} & B_t^\mathsf{T} \end{bmatrix})
\end{aligned}
\tag{8.6}
$$

where the matrices $A$ and $B$ constitute the input-output Jacobian $J_f$ of $f$

$$J_f = \begin{bmatrix} \nabla_x f_1^\mathsf{T} & \nabla_u f_1^\mathsf{T} \\ \vdots & \vdots \\ \nabla_x f_n^\mathsf{T} & \nabla_u f_n^\mathsf{T} \end{bmatrix} \in \mathbb{R}^{n \times (n+m)} = \begin{bmatrix} A & B \end{bmatrix}$$

and $\nabla_x f_i$ denotes the gradient of the $i$:th output of $f$ with respect to $x$. Our estimate $\hat{f}(x, u, w)$ of $f(x, u)$ will be parameterized by a vector $w$.[2] The distinction between $f$ and $\hat{f}$ will, however, be omitted unless required for clarity.

We frame the learning problem as an optimization problem with the goal of adjusting the parameters $w$ of $\hat{f}$ to minimize a cost function $V(w)$ by means of gradient descent. The cost function $V(w)$ can take many forms, but we will limit the scope of this work to quadratic loss functions of the one-step prediction error, i.e.,

$$V(w) = \frac{1}{2} \sum_t \left( x^+ - \hat{f}(x, u, w) \right)^\mathsf{T} \left( x^+ - \hat{f}(x, u, w) \right)$$

---

[2] We use $w$ to denote all the parameters of the neural network, i.e., weight matrices and bias vectors.

**Sampling of continuous-time models**

The functions $f$ and $f_c$ in (8.1) and (8.3) are quite different from each other [Åström and Wittenmark, 2013a]. It is well known that the Jacobian of the discrete-time model $f$ has eigenvalues different from that of the continuous-time counterpart $f_c$. The sampling procedure warps the space of Jacobian eigenvalues such that the origin is moved to the point 1 in the complex plane, and the imaginary axis is wrapped around the unit circle, with multiples of the Nyquist frequency occurring at the point -1. If the system is sampled with a high sample rate, or viewed differently, if the dynamics of the system is slow in relation to the time scale at which we are observing it, most eigenvalues[3] of $f_c$ are close to 0. The eigenvalues for the discrete-time $f$, however, tend to cluster around 1 when the sample rate is high. This disparity between continuous-time and discrete-time models has been of interest historically. It was found that controllers and models with slow dynamics relative to the sample rate suffered from implementations with low precision arithmetic [Åström and Wittenmark, 2011]. Middleton and Goodwin (1986) reformulated the discrete-time model using the so-called $\delta$-operator so that in the limit of infinite sample rate, the poles of the discrete-time model would coincide with the continuous-time counterpart. Related improvements in accuracy was found by Lennartson et al. (2012) when using the $\delta$-operator for solving linear-matrix inequalities related to systems with fast sampling. The reformulation is based on the definition of the derivative as the limit of a finite difference

$$\lim_{\Delta t \to 0} \frac{x^+ - x}{\Delta t} = f_c(x, u) \tag{8.7}$$

where a discrete-time model, with poles similar to the continuous-time poles, is obtained for a sufficiently small $\Delta t$. The increase in numerical accuracy from the reformulation can be understood from the Taylor expansion of the dynamics. If $\Delta t$ is small, the term $\Delta t f_c(x, u)$ will be insignificant next to the $x$ in the expression $x^+ = x + \Delta t \cdot f_c(x, u)$.

We will, alongside the naive discrete-time formulation $f$, also explore this reformulation of the discrete-time model, and introduce a new learning objective:

LEARNING OBJECTIVE 2

$$x^+ - x = \Delta x = g(x, u), \quad g : \mathbb{R}^n \times \mathbb{R}^m \mapsto \mathbb{R}^n$$
$$f(x, u) = g(x, u) + x \qquad \qquad \Box$$

where the second equation is equivalent to the first, but highlights a convenient implementation form that does not require transformation of the data.

To gain insight into how this seemingly trivial change in representation may affect learning, we note that this transformation will alter the Jacobian according to

$$J_g = \begin{bmatrix} A - I_n & B \end{bmatrix} \tag{8.8}$$

---

[3] We take the eigenvalues of a function to refer to the eigenvalues of the function Jacobian.

with a corresponding unit reduction of the eigenvalues of $A$. For systems with integrators, or slow dynamics in general, this transformation leads to a better conditioned estimation problem, something we will discuss in the next section. In Sec. 8.6 we investigate whether or not this transformation leads to a better prediction result and whether modern neural-network training techniques such as the ADAM optimizer [Kingma and Ba, 2014] and batch normalization [Ioffe and Szegedy, 2015] render this transformation superfluous. We further investigate how weight decay and choice of activation function affect the eigenvalue spectrum of the Jacobian of $f$ and $g$, and hence system stability.

From the expression $x^+ = g(x, u) + x$, the similarity with skip-connections as introduced in *resnet* [He et al., 2015] and LSTMs [Hochreiter and Schmidhuber, 1997] should be apparent, and we now have a control-theoretical, dynamical systems view of this transformation.

### Optimization landscape

To gain insight into the training of $f$ and $g$, we analyze the expressions for the gradient and Hessian of the respective cost functions. For a linear model $x^+ = Ax + Bu$, rewritten on regressor form $y = \mathbf{A}k$ with all parameters of $A$ and $B$ concatenated into the vector $k$, and a least-squares cost function $V(k) = \frac{1}{2}(y - \mathbf{A}k)^\mathsf{T}(y - \mathbf{A}k)$, the gradient and Hessian are given by

$$\nabla_k V = -\mathbf{A}^\mathsf{T}(y - \mathbf{A}k)$$
$$\nabla_k^2 V = \mathbf{A}^\mathsf{T}\mathbf{A}$$

The Hessian is clearly independent of both the output $y$ and the parameters $k$ and differentiating the output, i.e., learning a map to $\Delta x$ instead of to $x^+$, does not have any major impact on gradient-based learning. For a nonlinear model, this is not necessarily the case:

$$V(w) = \frac{1}{2}\sum_t \left(x^+ - f(x, u, w)\right)^\mathsf{T}\left(x^+ - f(x, u, w)\right)$$
$$\nabla_w V = \sum_{t=1}^{T}\sum_{i=1}^{n} -\left(x_i^+ - f_i(x, u, w)\right)\nabla_w f_i$$
$$\nabla_w^2 V = \sum_{t=1}^{T}\sum_{i=1}^{n} \nabla_w f_i \nabla_w f_i^\mathsf{T} - \left(x_i^+ - f_i(x, u, w)\right)\nabla_w^2 f_i$$

where $x_i^+ - f_i(x, u, w)$ constitute the prediction error. In this case, the Hessian depends on both the parameters and the target $x^+$. The transformation from $f$ to $g$ changes the gradients and Hessians according to

$$\nabla_w V = \sum_{t=1}^{T}\sum_{i=1}^{n} -\left(\Delta x_i - g_i(x, u, w)\right)\nabla_w g_i$$
$$\nabla_w^2 V = \sum_{t=1}^{T}\sum_{i=1}^{n} \nabla_w g_i \nabla_w g_i^\mathsf{T} - \left(\Delta x_i - g_i(x, u, w)\right)\nabla_w^2 g_i$$

When training begins, both $f$ and $g$ are initialized with small random weights, and $\|f\|$ and $\|g\|$ will typically be small. If the system we are modeling is of low-pass character, i.e., $\|\Delta x\|$ is small, the prediction error of $g$ will be closer to zero compared to $f$. The transformation from $f$ to $g$ can thus be seen as *preconditioning* the problem by decreasing the influence of the term $\nabla^2_w g = \nabla^2_w f$ in the Hessian. With only the positive semi-definite term $\nabla_w(g)\nabla_w(g)^\mathsf{T} = \nabla_w(f)\nabla_w(f)^\mathsf{T}$, corresponding to $\mathbf{A}^\mathsf{T}\mathbf{A}$ in the linear case, remaining, the optimization problem might become easier. Similarly, $g$ starts out closer to a critical point $\nabla_w V = 0$, which might make convergence faster. The last two claims will be investigated in Sec. 8.6.

Motivation for the statement that the optimization of $g$ is better conditioned is obtained by considering the derivation of the Gauss-Newton (GN) optimization algorithm [Nocedal and Wright, 1999]. GN resembles the Newton algorithm, but ignores the second-order term $(x^+ - f(w))\nabla^2_w f$ by considering only a linear approximation to the prediction error. This algorithm is likely to converge if either the Hessian $\nabla^2_w f$ is small, or, more importantly, the prediction errors $(x^+ - f(w))$ are small [Nocedal and Wright, 1999, p. 259].

An approximation to the GN algorithm, suitable for training of neural networks, was developed by Botev et al. (2017), but although they showed improved convergence over ADAM [Kingma and Ba, 2014] in terms of iteration count, ADAM outperformed the approximate GN algorithm in terms of wall-clock time on a GPU.

## 8.4 Weight Decay

Weight decay is commonly an integral part of training procedure in the deep learning setting, used to combat overfitting [Murphy, 2012; Goodfellow et al., 2016]. We can think of weight decay as either penalizing complexity of the model, or as encoding prior knowledge about the size of the model coefficients. $L_2$ weight decay is, however, a blunt weapon. While often effective at mitigating overfitting, the added penalty term might introduce a severe bias in the estimate. Since the bias always is directed towards smaller weights, it can have different consequences depending on what small weights imply for a particular model architecture. For a discrete-time model $f$, small weights intuitively imply small eigenvalues and a small output. For $x^+ = g(x, u) + x$, on the other hand, small weights imply eigenvalues closer to 1. Weight decay might thus have different effects on learning $f$ and $g + x$. With a high sample rate and consequently eigenvalues close to 1, weight decay is likely to bias the result of learning $g + x$ in a milder way as compared to $f$.

The intuition of a small $w$ implying small eigenvalues of the network Jacobian is, unfortunately, somewhat deceiving. The largest eigenvalue $\lambda_{\max}(\nabla_x f(x, u, w))$ is not constant on a level surface of $\|w\|$. As we move to a level surface with smaller value of $\|w\|$, the biggest $\lambda_{\max}(\nabla_x f)$ we can find is reduced, so an upper bound on $\|w\|$ effectively puts an upper bound on $\lambda_{\max}(\nabla_x f)$, but the connection between them is complicated and depends on the model of $f$.

A natural question to ask is if weight decay can bias the eigenvalues of the learned function to arbitrarily chosen locations. A generalized form of model formulation is

$$x^+ = h(x, u) + Ax + Bu$$

where $A$ and $B$ can be seen as a nominal linear model around which we learn the nonlinear behavior. Weight decay will for this formulation bias the Jacobian towards $A$ and $B$ which can be chosen arbitrarily. Obviously, choosing a nominal linear model is not always easy, and may in some cases not make sense. One can however limit the scope to damped formulations like $x^+ = h(x, u) + \gamma x$, where $\gamma$ is a scalar or a diagonal matrix that shifts the nominal eigenvalues along the real axis to, e.g., encourage stability.

The use of regularization to promote stability of the learned dynamical system is not the goal of this work, and weight decay is not well suited for the task. Oberman and Calder (2018) discuss Lipschitz regularization and show how the Lipschitz constant of a DNN can easily be upper bounded. Although Oberman and Calder (2018) do not consider learning of dynamical systems, we conjecture that their Lipschitz regularization would be better suited for the task. Oberman and Calder (2018) that most activation functions have a Lipschitz constant of 1, rendering the product of all weight matrices, $\|W_L W_{L-1} \cdots W_2 W_1\|$, an upper bound on the Lipschitz constant of the entire network. In the next section, we show how the input-output Jacobian of a standard neural network is similarly straightforward to compute and also this matrix could prove useful in regularizing the Lipschitz constant of the model.

In Sec. 8.6, we investigate the influence of weight decay on the learning of DNNs representing $f$ and $g$ and in the next section, we highlight a way of encoding prior knowledge of the system to be modeled and introduce a new form of regularization that, if the system enjoys certain properties, introduces less bias as compared to weight decay.

## 8.5 Tangent-Space Regularization

The topic of Chap. 7 was learning heavily overparameterized models by restricting the flexibility of the model by means of appropriate regularization. For systems where the function $f$ is known to be smooth, the Jacobian $J_f(t)$ will vary slowly. In the rigid-body dynamical model in (8.2), for instance, the inertial and gravitational forces are changing smoothly with the joint configuration. Similar to the penalty function in (7.3), a natural addition to the cost function of the optimization problem would thus be a tangent-space regularization term on the form

$$\sum_t \|\hat{J}_{t+1} - \hat{J}_t\| \tag{8.9}$$

which penalizes changes in the input-output Jacobian of the model over time, a strategy we refer to as Jacobian propagation or *Jacprop*.

Taking the gradient of terms depending on the model Jacobian requires calculation of higher order derivatives. Depending on the framework used for optimiza-tion, this can limit the applicability of the method. We thus proceed to describe how we implemented the penalty term of (8.9).

### Implementation details

The inclusion of (8.9) in the cost function implies the presence of nested differentiation in the gradient of the cost function with respect to the parameters, $\nabla_w V$. The complications arise in the calculation of the term

$$\nabla_w \sum_t \left\| \hat{J}_{t+1}(w) - \hat{J}_t(w) \right\| \tag{8.10}$$

where $J$ is composed of $\nabla_x f(x, u, w)$ and $\nabla_u f(x, u, w)$. Many, but not all, deep-learning frameworks unfortunately lack support for nested differentiation. Further, most modern deep-learning frameworks employ reverse-mode automatic differentiation (AD) for automatic calculation of gradients of cost functions with respect to network weight matrices [Merriënboer et al., 2018]. Reverse-mode AD is very well suited for scalar functions of many parameters. For vector-valued functions, however, reverse-mode AD essentially requires separate differentiation of each output of the function. This scales poorly and is a seemingly unusual use case in deep learning; most AD frameworks have no explicit support for calculating Jacobians. These two obstacles together might make implementing the suggested regularization hard in practice. Indeed, an attempt was made at implementing the regularization with nested automatic differentiation, where $\nabla_w V$ was calculated using reverse-mode AD and $J(w)$ using forward-mode AD. This required finding AD software capable of nested differentiation and was made very difficult by subtleties regarding closing over the correct variables for the inner differentiation. The resulting code was also very slow to execute.

The examples detailed later in this chapter instead make use of handwritten inner differentiation, where $\nabla_w V$ once again is calculated using reverse-mode AD, but the calculation of $J(w)$ is done manually. For a DNN composed of affine transformations ($Wx + b$) followed by elementwise nonlinearities ($\sigma$), this calculation can be expressed recursively as

$$
\begin{aligned}
a_1 &= W_1 x + b_1 \\
a_i &= W_i l_{i-1} + b_i, \quad i = 2 \dots L \\
l_i &= \sigma(a_i) \\
\nabla_x l_i &= \nabla_x \{\sigma(a_i)\} \nabla_x l_{i-1} \\
&= (\nabla_x \sigma)\big|_{a_i} \cdot \nabla_x a_i \cdot \nabla_x l_{i-1} \\
&= (\nabla_x \sigma)\big|_{a_i} \cdot W_i \cdot \nabla_x l_{i-1}
\end{aligned}
$$

where $\sigma$ denotes the activation function.

In a practical implementation, the Jacobian can be calculated at the same time as the forward pass of the network. An implementation is given in Algorithm 4.

The function defined in Algorithm 4 is suitable for use in an outer cost function which is differentiated with respect to $w$ using reverse mode AD.

**Algorithm 4** Julia code for calculation of both forward pass and input-output Jacobian of neural network $f$. The code uses the tanh activation function and assumes that the weight matrices are stored according to $w = \{W_1 \, b_1 \dots W_L \, b_L\}$

```julia
function forward_jac(w,x)
    l = x
    J = Matrix{eltype(w[1])}(I,length(x),length(x)) # Initial J = I_n
    for i = 1:2:length(w)-2
        W,b = w[i], w[i+1]
        a   = W*l .+ b
        l   = σ.(a)
        ∇a  = W
        ∇σ  = ∇σ(a)
        J   = ∇σ * ∇a * J
    end
    J = w[end-1] * J # Linear output layer
    return w[end-1]*l .+ w[end] , J
end
∇σ(a) = Matrix(Diagonal((sech.(a).^2)[:])) # ∇_a tanh(a)
```

## 8.6 Evaluation

The previously described methods were evaluated on two benchmark problems. We compared performance on one-step prediction and further compared the fidelity of the Jacobian of the estimated models, an important property for optimization algorithms making use of the models. The benchmarks consist of a pendulum on a cart, and randomized, stable linear systems.

   We initially describe a baseline neural-network model used in the experimental evaluation, which we use to draw conclusions regarding the different learning objectives. We describe how deviations from this baseline model alter the conclusions drawn in Sec. 8.B.

### Nominal model

Both functions $f$ and $g$ are modeled as neural networks. For the linear-system task, the networks had 1 hidden layer with 20 neurons; in the pendulum task, the networks had 3 hidden layers with 30 neurons each.

   A comparative study of 6 different activation functions, presented in Sec. 8.A, indicated that some unbounded activation functions, such as the relu and leaky relu functions [Ramachandran et al., 2017], are less suited for the task at hand, and generally, the tanh activation function performed best and was chosen for the evaluation.

   We train the models using the ADAM optimizer with a fixed step-size and fixed number of epochs. The framework for training, including all simulation experiments reported in this chapter, is published at [*JacProp.jl*, B.C., 2018] and is implemented in the Julia programming language [Bezanson et al., 2017] and the

---

**Algorithm 5** Generation of random, stable linear systems.

---

$A_0 = 10 \times 10$ matrix of random coefficients
$A = A_0 - A_0^\mathsf{T}$ skew-symmetric = pure imaginary eigenvalues
$A = A - \Delta t\,I$ make 'slightly' stable
$A = \exp(\Delta t\,A)$ discrete time, sample time $\Delta t$
$B = 10 \times 10$ matrix of random coefficients

---

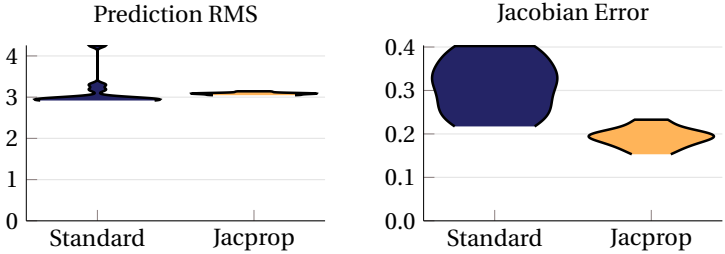

**Figure 8.1**  Left: Distribution of prediction errors on the validation data for $g + x$ trained on the linear-system task. Each violin represents 12 Monte-Carlo runs. Right: Distribution of errors in estimated Jacobians. The figure indicates that tangent-space regularization through Jacobian propagation is effective and reduces the error in the estimated Jacobian without affecting the prediction error performance.

Flux machine learning library [Innes, 2018].

**Randomized linear system**

To assess the effectiveness of Jacobian propagation we create random, stable linear systems according to Algorithm 5 and evaluate the Jacobian of the learned model for points sampled randomly in the state space. Learning a linear system allows us to easily visualize the learned Jacobian eigenvalues and relate them to the eigenvalues of the true system, which remain constant along a trajectory. While the proposed regularization penalty is theoretically optimal for a linear system, this simple setting allows us to verify that the additional penalty term does not introduce any undesired difficulty in the optimization that results in convergence to local optima, etc.

We train two models, the first model is trained using weight decay and the second using Jacprop. The regularization parameters are in both cases tuned such that prediction error on the test data is minimized. The models are trained for 300 epochs on two trajectories of 200 time steps each. The input was low-pass filtered Gaussian noise where the cutoff frequency was high in relation to the time constant of the system so as to excite all modes of the system.

The results are illustrated in Fig. 8.1 and Fig. 8.2. During training, the model trained without tangent-space regularization reaches a far lower training error, but validation data indicates that overfitting has occurred. The number of parameters in the models was 6.2 times larger than the number of parameters in the true linear system and overfitting is thus a concern in this scenario.

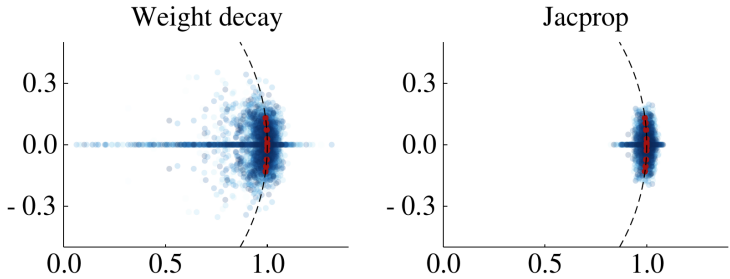

**Figure 8.2**    Learned Jacobian eigenvalues of $g + x$ for points sampled randomly in the state space (blue) together with the eigenvalues of the true model (red). Tangent-space regularization (right) leads to better estimation of the Jacobian with eigenvalues in a tighter cluster around the true eigenvalues close to the unit circle.

To calculate the Jacobian error, we calculate the shortest distance between each eigenvalue in the true Jacobian to any of the eigenvalues of the estimated Jacobians, and vice versa. We then sum these distances and take the mean over all the data points in the validation set. This allows us to penalize both failure to place an eigenvalue close to a true eigenvalue, and placing an eigenvalue without a true eigenvalue nearby. The model trained with tangent-space regularization learns better Jacobians while producing the same prediction error, indicated in Fig. 8.1 and Fig. 8.2.

The effect of weight decay on the learned Jacobian is illustrated in Fig. 8.3. Due to overparameterization, heavy overfitting is expected without adequate regularization. Not only is it clear that learning of $g + x$ has been more successful than learning of $f$ in the absence of weight decay, but we also see that weight decay has had a deteriorating effect on learning $f$, whereas it has been beneficial in learning $g + x$. This indicates that the choice of architecture interacts with the use of standard regularization techniques and must be considered while modeling.

### Pendulum-on-cart task

A pendulum attached to a moving cart is simulated to assess the effectiveness of Jacprop and weight decay on a system with nonlinear dynamics and thus a changing Jacobian along a trajectory. An example trajectory of the system described by (8.11)-(8.12), which has 4 states $(\theta, \dot{\theta}, p, v)$ and one control input $u$, is shown in Fig. 8.4. This task demonstrates the utility of tangent-space regularization for systems where the regularization term is not the theoretically perfect choice, as was the case with the linear system. Having access to the true system model and state representation also allows us to compare the learned Jacobian to the true system Jacobian. We simulate the system with a superposition of sine waves of different frequencies and random noise as input and compare prediction error as well as the error in the estimated Jacobian. The dynamical equations of the system are given by

$$\ddot{\theta} = -\frac{g}{l}\sin(\theta) + \frac{u}{l}\cos(\theta) - d\dot{\theta} \tag{8.11}$$

$$\dot{v} = \ddot{p} = u \tag{8.12}$$

where $g, l, d$ denote the acceleration of gravity, the length of the pendulum and the damping, respectively.

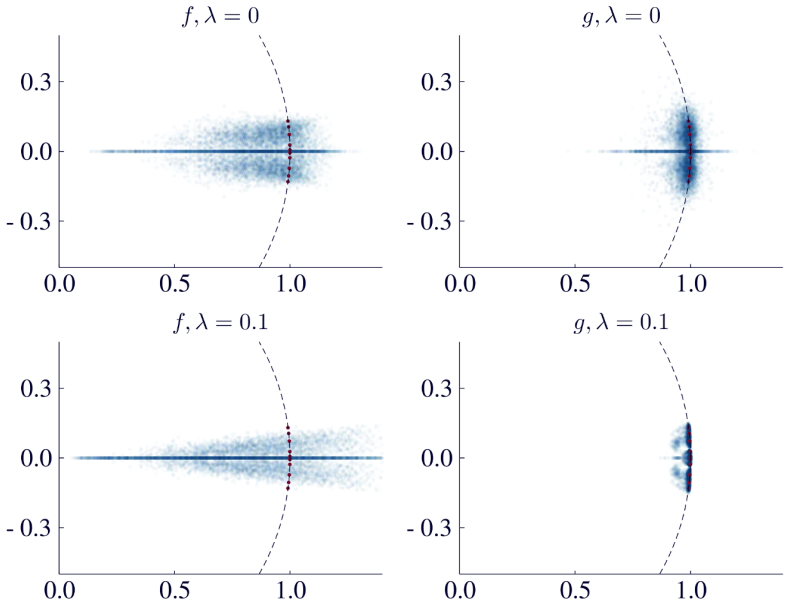

**Figure 8.3** Eigenvalues of learned Jacobians for the linear system task. True eigenvalues are shown in red, and eigenvalues of the learned model for points sampled randomly in the state space are shown in blue. The top/bottom rows show models trained without/with weight decay, left/right columns show $f/g$. Weight decay has a deteriorating effect on learning $f$, pulling some eigenvalues towards 0 while causing others to become much larger than 1, resulting in a very unstable system. Weight decay is beneficial for learning $g + x$, keeping the eigenvalues close to 1.

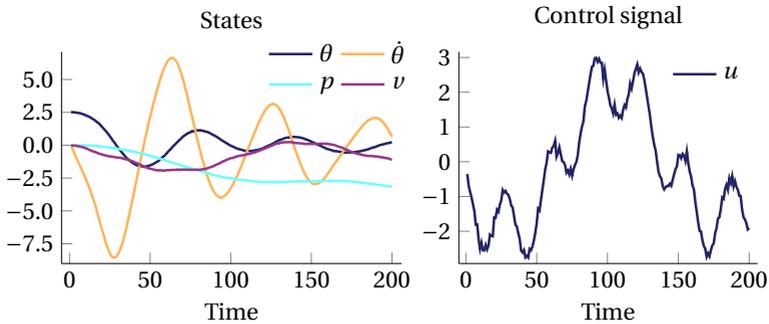

**Figure 8.4** Example trajectory of pendulum on a cart.

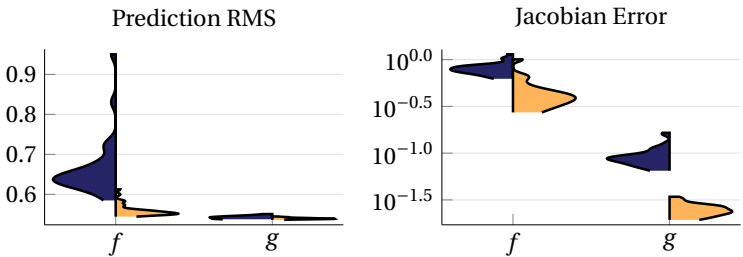

**Figure 8.5**   Left: Distribution of prediction errors on the validation data for the pendulum on a cart task using tanh activation functions. Each violin represents 30 Monte-Carlo runs. The figure indicates that tangent-space regularization through Jacobian propagation is effective and reduces prediction error for $f$, but not $g$, where weight decay performs equally well. Right: Distribution of errors in estimated Jacobians. Jacprop is effective at improving the fidelity of the model Jacobians for both $f$ and $g$.

Once again we train two models, one with weight decay and one with Jacprop. The regularization parameters were in both cases chosen such that prediction error on test data was minimized. The models were trained for 2500 epochs on two trajectories of 200 time steps, approximately an order of magnitude fewer data points than the number of parameters in the models.

The prediction and Jacobian errors for validation data, i.e., trajectories not seen during training, are shown in Fig. 8.5. The results indicate that while learning $f$, tangent-space regularization leads to reduced prediction errors compared to weight decay, with lower mean error and smaller spread, indicating more robust learning. Learning of $g + x$ did not benefit much from Jacobian propagation in terms of prediction performance compared to weight decay, and both training methods perform on par and reach a much lower prediction error than the $f$ models.

To assess the fidelity of the learned Jacobian, we compare it to the ground-truth Jacobian of the simulator. We display the distribution of errors in the estimated Jacobians in Fig. 8.5. The results show a significant benefit of tangent-space regularization over weight decay for learning both $f$ and $g + x$, with a reduction of the mean error as well as a smaller spread of errors (please note that the figure has logarithmic *y*-axis).

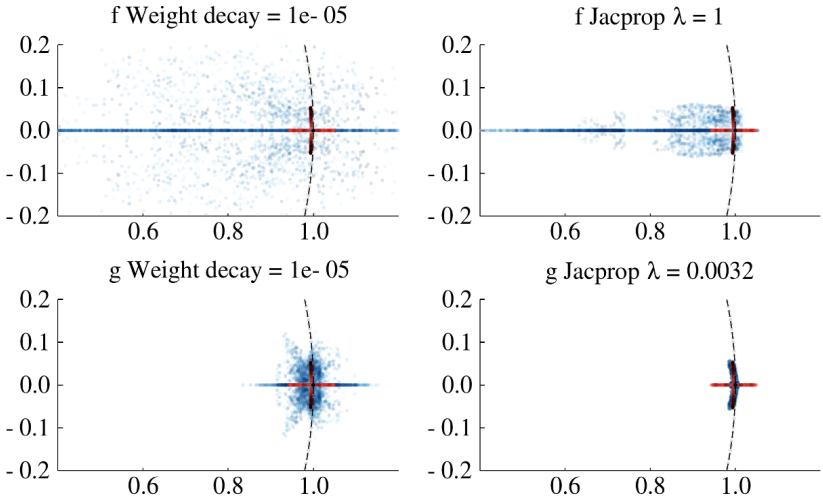

**Figure 8.6** Eigenvalues of the pendulum system using the tanh activation function on validation data.

The individual entries of the model Jacobian along a trajectory for one instance of the trained models are visualized as functions of time in Fig. 8.7. The figure illustrates the smoothing effect of the tangent-space regularization and verifies the smoothness assumption on the pendulum system. We also note that the regularization employed does not restrict the ability of the learned model to change its Jacobian along the trajectory, tracking the true system Jacobian. This is particularly indicated in the $(1, 2)$ and $(4, 3)$ entries in Fig. 8.7. The figure also shows how weight decay tuned for optimal prediction performance allows a rapidly changing Jacobian, indicating overfitting. If a higher weight-decay penalty is used, this overfitting is reduced, at the expense of prediction performance, hinting at the heavily biasing properties of excessive weight decay.

The eigenvalues of the true system and learned models are visualized in Fig. 8.6.

## 8.7 Discussion

Throughout experiments, we note that $g + x$ generally trains faster, reaches a lower value of the objective function compared to $f$ and learns a Jacobian that is closer to the Jacobian of the true system model. In a reinforcement-learning setting where available data is limited and Jacobians of the learned model are used for optimization, this property is of great importance.

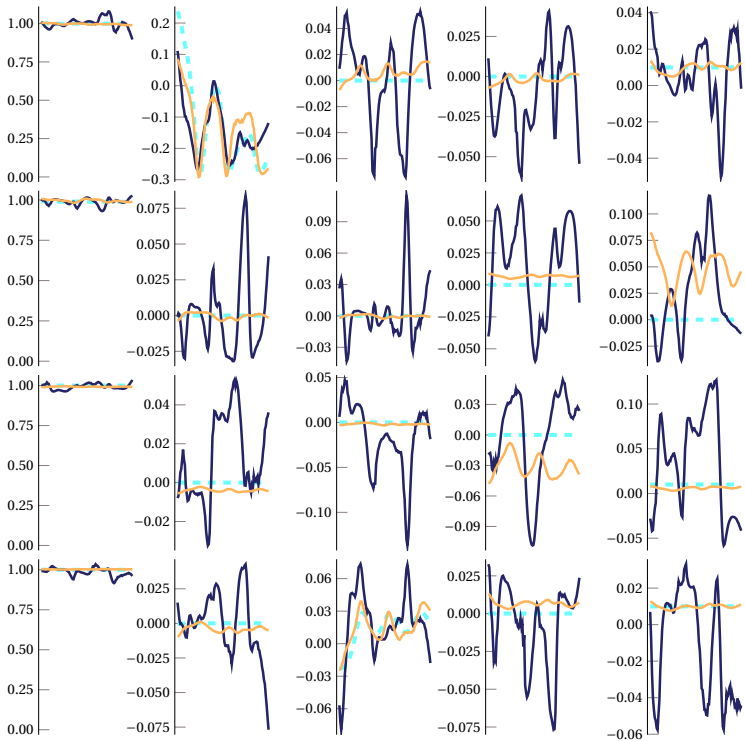

**Figure 8.7** Individual entries in the Jacobian as functions of time step along a trajectory of the pendulum system (Ground truth ▬ ▬, Weight decay ▬▬, Jacprop ▬▬). The figure verifies the smoothness assumption on the system and indicates that Jacprop is successful in promoting smoothness of the Jacobian of the estimated model. The entries $(1,2)$ and $(4,3)$ change the most along the trajectory, and the Jacprop-regularized model tracks these changes well without excessive smoothing.

The structure of $g + x$ resembles that of a residual network [He et al., 2016], where a *skip connection* is added between the input and a layer beyond the first adjacent layer, in our case, directly to the output. While skip connections have helped to enable successful training of very deep architectures for tasks such as image recognition, we motivated the benefit of the skip connection with classical theory for sampling of continuous-time systems [Middleton and Goodwin, 1986] and an analysis of the model Hessian, where we compared learning of $g + x$ to the Gauss-Newton algorithm. Exploring the similarities with residual networks remains an interesting avenue for future work. The notion of skip connections is seen also in LSTMs [Hochreiter and Schmidhuber, 1997], once again motivated by gradient flow. LSTMs are often used to model and generate natural language.

In this domain, the notion of time constants is less well defined. The state of a recurrent neural network for natural language modeling can change very abruptly depending on which word is input. LSTMs thus incorporate also a gating mechanism to allow components of the state to be "forgotten". For mechanical systems, an example of an analogous situation is a state constraint. A dynamical model of a bouncing ball, for instance, must learn to forget the velocity component of the state when the position reaches the constraint.

The scope of this chapter was limited to settings where a state sequence is known. This allowed us to reason about eigenvalues of Jacobians and compare the learned Jacobians to those of the ground-truth model of a simulator. In a more general setting, learning the transformation of past measurements and inputs to a state representation is required, e.g., using a network with recurrence or an auto-encoder [Karl et al., 2016]. Initial results indicate that the conclusions drawn regarding the formulation ($f$ vs. $g + x$) of the model and the effect of weight decay remain valid in the RNN setting, but a more detailed analysis is the target of future work. The concept of tangent-space regularization applies equally well to the Jacobian from input to hidden state in an RNN, and potential benefits of this kind of regularization in the general RNN setting remain to be investigated.

We also restricted our exposition to the simplest possible noise model, corresponding to the equation-error problem discussed in Sec. 3.1. In a practical scenario, estimating a more sophisticated noise model may be desirable. Noise models in the deep-learning setting add complexity to the estimation in the same way as for linear models. A practical approach, inspired by pseudo-linear regression [Ljung, 1987], is to train a model without noise model and use this model to estimate the noise sequence through the prediction errors. This noise sequence can then be used to train a noise model. If this noise model is trained together with the dynamics model, back-propagation through time is required and the computational complexity is increased. Deep-learning examples including noise models are found in [Karl et al., 2016].

## 8.8 Conclusions

We investigated different architectures of a neural-network model for modeling of dynamical systems and found that the relationship between sample time and system bandwidth affects the preferred choice of architecture, where an approximator architecture incorporating an identity element similar to that of LSTMs and *resnet*, train faster and generally generalize better in terms of all metrics if the sample rate is high. An analysis of gradient and Hessian expressions motivated the difference and conclusions were reinforced by experiments.

The effect of including $L_2$ weight decay was investigated and shown to vary greatly with the model architecture. Implications on the stability and tangent-space eigenvalues of the learned model highlight the need to consider the architecture choice carefully.

We further demonstrated how tangent-space regularization by means of Ja-cobian propagation can be used to incorporate prior knowledge of the modeled system and regularize the learning of a neural network model of a smooth dynamical system with an increase in prediction performance as well as increasing the fidelity of the learned Jacobians as result.

## Appendix A. Comparison of Activation Functions

Figure 8.8 displays the distribution of prediction errors over 200 Monte-Carlo runs with different random seeds and different activation functions. The results indicate that the relu and leaky relu functions are worse suited for the considered task and are thus left out from the set of selected bootstrap ensemble activation functions. For a definition of the activation functions, see, e.g., [Ramachandran et al., 2017].

## Appendix B. Deviations from the Nominal Model

***Number of neurons*** Doubling or halving the number of neurons generally led to worse performance.

***Number of layers*** Adding a fully connected layer with the same number of neurons did not change any conclusions.

***Dropout*** Inclusion of dropout (20%) increased prediction and simulation performance for $g$ while performance was decreased for $f$. Performance on Jacobian estimation was in general worse.

***Layer normalization*** Prediction and simulation performance remained similar or slightly worse, whereas Jacobian estimation broke down completely.

***Measurement noise*** Although not a hyper parameter, we investigated the influence of measurement noise on the identification results. Increasing degree of measurement noise degraded all performance metrics in predictable ways and did not change any qualitative conclusions.

***Optimizer, Batch normalization, Batch size*** Neither of these modifications altered the performance significantly.

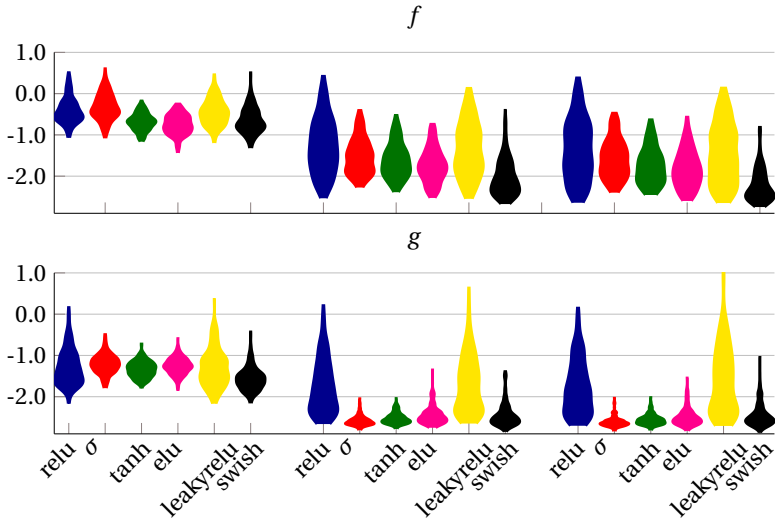

**Figure 8.8**   Distributions (cropped at extreme values) of log-prediction errors on the validation data after 20, 500 and 1500 (left, middle, right) epochs of training for different activation functions. Every violin is representing 200 Monte-Carlo runs and is independently normalized such that the width is proportional to the density, with a fixed maximum width.

# 9

# Friction Modeling and Estimation

## 9.1 Introduction

All mechanical systems with moving parts are subject to friction. The friction force is a product of interaction forces on an atomic level and is always resisting relative motion between two elements in contact. Because of the complex nature of the interaction forces, friction is usually modeled based on empirical observations. The simplest model of friction is the Coulomb model, (9.1), which assumes a constant friction force acting in the reverse direction of motion

$$F_f = k_c \operatorname{sign}(v) \tag{9.1}$$

where $k_c$ is the Coulomb friction constant and $v$ is the relative velocity between the interacting surfaces.

A slight extension to the Coulomb model includes also velocity dependent terms

$$F_f = k_v v + k_c \operatorname{sign}(v) \tag{9.2}$$

where $k_v$ is the viscous friction coefficient. The Coulomb model and the viscous model are illustrated in Fig. 9.1. If the friction is observed to vary with the direction of motion, $\operatorname{sign}(v)$, the model (9.2) can be extended to

$$F_f = k_v v + k_c^+ \operatorname{sign}(v^+) + k_c^- \operatorname{sign}(v^-) \tag{9.3}$$

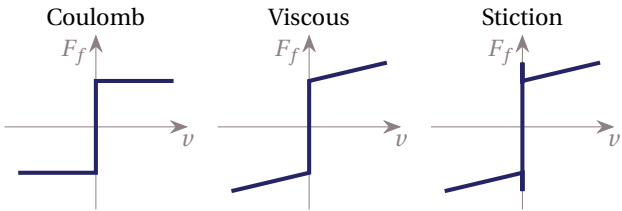

**Figure 9.1**    Illustrations of simple friction models.

where the sign operator is defined to be zero for $v = 0$, $v^+ = \max(0, v)$ and $v^- = \min(0, v)$.

It is commonly observed that the force needed to initiate movement from a resting position is higher than the force required to maintain a low velocity. This phenomenon, called stiction, is illustrated in Fig. 9.1. The friction for zero velocity and an external force $F_e$ can be modeled as

$$F_f = \begin{cases} F_e & \text{if } v = 0 \text{ and } |F_e| < k_s \\ k_s \operatorname{sign} F_e & \text{if } v = 0 \text{ and } |F_e| \geq k_s \end{cases} \tag{9.4}$$

where $k_s$ is the stiction friction coefficient. An external force greater than the stiction force will, according to model (9.4), cause an instantaneous acceleration and a discontinuity in the friction force.

The models above suffice for many purposes but can not explain several commonly observed friction-related phenomena, such as the Stribeck effect and dynamical behavior, etc. [Olsson et al., 1998]. To explain more complicated behavior, dynamical models such as the Dahl model [Dahl, 1968] and the LuGre model [De Wit et al., 1995] have been proposed.

Most proposed friction models include velocity-dependent effects, but no position dependence. A dependence upon position is however often observed, and may stem from, for instance, imperfect assembly, irregularities in the contact surfaces or application of lubricant, etc. [Armstrong et al., 1994]. Modeling of the position dependence is unfortunately nontrivial due to an often irregular relationship between the position and the friction force. Several authors have however made efforts in the area. Armstrong (1988) used accurate friction measurements to implement a look-up table for the position dependence and Huang et al. (1998) adaptively identified a sinusoidal position dependence.

More recent endeavors by Kruif and Vries (2002) used an Iterative Learning Control approach to learn a feedforward model including position-dependent friction terms.

In [Bittencourt and Gunnarsson, 2012], no significant positional dependence of the friction in a robot joint was found. However, a clear dependence upon the temperature of contact region was reported. To allow for temperature sensing, the grease in the gear box was replaced by an oil-based lubricant, which allowed for temperature sensing in the oil flow circuit.

A standard approach in dealing with systems with varying parameters is recursive identification during normal operation [Johansson, 1993]. Recursive identification of the models (9.1) and (9.2) could account for both position- and temperature dependence. Whereas straight forward in theory, it is often hard to perform in a robust manner in practical situations. Presence of external forces, accelerating motions, etc. require either a break in the adaptation, or an accurate model of the additional dynamics. Many control programs, such as time-optimal programs, never exhibit zero acceleration, and thus no chance for parameter adaptation.

To see why unmodeled dynamics cause a bias in the estimated parameters, consider the system

$$f = ma + f_{ext} + F_f \tag{9.5}$$

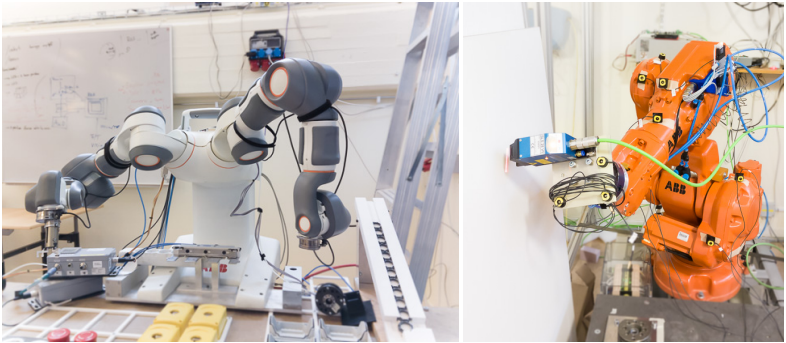

**Figure 9.2** Dual-arm robot and industrial manipulator IRB140 used for experimental verification of proposed models and identification procedures.

with mass $m$ and externally applied force $f_{ext}$. If we model this system as

$$\hat{f} = F_f \tag{9.6}$$

we effectively have the disturbance terms

$$f - \hat{f} = ma + f_{ext} \tag{9.7}$$

These terms do not constitute uncorrelated random noise with zero mean and will thus introduce a bias in the estimate. It is therefore of importance to obtain as accurate models as possible offline, where conditions can be carefully controlled to minimize the influence of unmodeled dynamics.

This chapter develops a model that incorporates positional friction dependence as well as an implicitly temperature-dependent term. The proposed additions can be combined or used independently as appropriate. Since many industrially relevant systems lack temperature sensing in areas of importance for friction modeling, a sensor-less approach is proposed. Both models are used for identification of friction in the joint of an industrial collaborative-style robot, see Fig. 9.2, and special aspects of position dependence are verified on a traditional industrial manipulator.

## 9.2 Models and Identification Procedures

This section first introduces a general identification procedure for friction models linear in the parameters, based on the least-squares method, followed by the introduction of a model that allows for the friction to vary with position. Third, a model that accounts for temperature-varying friction phenomena is introduced. Here, a sensor-less approach where the power loss due to friction is used as an input to a first-order system, is adopted.

As the models are equally suited for friction due to linear and angular movements, the terms force and torque are here used interchangeably.

## Least-squares friction identification

A standard model of the torques in rigid-body dynamical systems, such as industrial robots, is [Spong et al., 2006]

$$\tau = M(p)a + C(p,v)v + G(p) + F(v) \tag{9.8}$$

where $a = \dot{v} = \ddot{p}$ is the acceleration, $\tau$ the control torque, $M, C, G$ are matrices representing inertia-, Coriolis-, centrifugal- and gravitational forces and $F$ is a friction model. If a single joint at the time is operated, at constant velocity, Coriolis effects disappear [Spong et al., 2006] and

$$\left.\begin{array}{r} C(p,v) = 0 \\ a = 0 \end{array}\right\} \Rightarrow \tau = G(p) + F(v) \tag{9.9}$$

To further simplify the presentation, it is assumed that $G(p) = 0$. This can easily be achieved by either aligning the axis of rotation with the gravitational vector such that gravitational forces vanish, by identifying and compensating for a gravity model[1] or, as in [Bittencourt and Gunnarsson, 2012], performing a symmetric experiment with both positive and negative velocities and calculating the torque difference.

As a result of the discontinuity of the Coulomb model and the related uncertainty in estimating the friction force at zero velocity, datapoints where the velocity is not significantly different from zero must be removed from the dataset used for estimation. Since there is a large probability that these points will have the wrong sign, inclusion of these points might lead to severe bias in the estimate of the friction parameters.

## Estimation in multi-joint robots

When performing friction estimation for multi-link robots, such as a serial manipulator, one can often choose to perform experiments on individual joints one at a time. This approach, while structured and systematic, might require additional time to complete. Another alternative is to perform an experiment on all joints simultaneously. The simplifying assumption $G(q) = 0$ is then invalid and one must simultaneously consider full gravity model estimation.

The simple models described in Sec. 9.1 are all linear in the parameters and can be estimated efficiently using the methods described in Sec. 6.2.

***Gear coupling*** Friction occurs between all sliding surfaces. In an electrical motor with gears, this implies the presence of friction on both sides of the gear box as well as in the gear itself. In a simple gear, it is not possible, nor necessary, to distinguish between the sources of friction, only the sum is observable. However, many multi-link manipulators exhibit gear coupling between consecutive motors. This complicates friction estimation since the movement of a motor with coupled gears causes friction between parts that are also affected by the movement of the coupled motors. For instance, some manipulators have a triangular gear-ratio matrix for the spherical wrist. As an example, the gear-ratio matrix for the ABB IRB140 has the following structure

$$G = \begin{bmatrix} * & & & & & \\ & * & & & & \\ & & * & & & \\ & & & * & & \\ & & & * & * & \\ & & & * & * & * \end{bmatrix} \tag{9.10}$$

with the relations between relevant motor- and arm-side quantities given by

$$q_a = G q_m$$
$$\tau_a = G^{-1} \tau_m \tag{9.11}$$

---

[1] For a single joint, this simply amounts to appending the regressor matrix $\mathbf{A}$ with $\begin{bmatrix} \sin(p) & \cos(p) \end{bmatrix}$

This causes cross couplings between the arm-side and motor-side friction for the last three joints. For the first three joints, only the sum of arm- and motor-side friction is visible, but for joint 5 and 6, friction models of the motor-side friction for joint 4 and joints 4 and 5, respectively, are needed. Experiments on an ABB IRB2400 robot indicate that the friction on both sides of the motor is of roughly equal importance for the overall result, and special treatment of the wrist joints is therefore crucial.

For the example above, a friction model for joint 6 will thus have to contain terms dependent on the velocities of the fourth and fifth motors as well, e.g.:

$$F_{f6} = k_{c6} \operatorname{sign} \dot{q}_6 + k_{v6} \dot{q}_6 + k_{c5} \operatorname{sign} \dot{q}_5 + k_{v5} \dot{q}_5 + k_{c4} \operatorname{sign} \dot{q}_4 + k_{v4} \dot{q}_4 \tag{9.12}$$

## 9.3 Position-Dependent Model

As mentioned in Sec. 9.1, a positional, repeatable friction dependence is often observed in mechanical systems. This section extends the simple nominal models presented in Sec. 9.1 with position-dependent terms, where the position dependence is modeled with a basis-function expansion. Througout the chapter, we will make use of Gaussian basis functions, with the implicit assumption that the estimated function is smooth and differentiable. This choice is, however, not important for the development of the method, and the choice of basis functions should be made with considerations mentioned in Sec. 6.3. One such consideration is the discontinuity of Coulomb friction at zero velocity, which we handle explicitly by truncating half of the basis functions for positive velocities, and vice versa. We further explicitly estimate the mean Coulomb coefficient even when estimating position dependent models, and let the basis-function expansion estimate deviations around a simple nominal model.

We define the Gaussian RBF kernel $\kappa$ and the kernel vector $\phi$

$$\kappa(p, \mu, \sigma) = \exp\left(-\frac{(p-\mu)^2}{2\sigma^2}\right) \tag{9.13}$$

$$\phi(p) : (p \in \mathcal{P}) \mapsto \mathbb{R}^{1 \times K}$$

$$\phi(p) = \left[\kappa(p, \mu_1, \sigma), \cdots, \kappa(p, \mu_K, \sigma)\right] \tag{9.14}$$

where $\mu_i \in \mathcal{P}, i = 1, ..., K$ is a set of $K$ evenly spaced centers. For each input position $p \in \mathcal{P} \subseteq \mathbb{R}$, the kernel vector $\phi(p)$ will have activated (>0) entries for the kernels with centers close to $p$. Refer to Fig. 6.2 for an illustration of RBFs. The kernel vector is included in the regressor-matrix **A** from Sec. 6.2 such that if used together with a nominal, viscous friction model, **A** and the parameter vector $k$ are given by

$$\mathbf{A} = \begin{bmatrix} v_1 & \text{sign}(v_1) & \phi(p_1) \\ \vdots & \vdots & \vdots \\ v_N & \text{sign}(v_N) & \phi(p_N) \end{bmatrix} \in \mathbb{R}^{N \times (2+K)}, \ k = \begin{bmatrix} k_v \\ k_c \\ k_\kappa \end{bmatrix} \tag{9.15}$$

where $k_\kappa \in \mathbb{R}^K$ denotes the parameters corresponding to the kernel vector entries. The number of RBFs to include and the bandwidth $\sigma$ is usually chosen based on evidence maximization or cross validation [Murphy, 2012].

The position-dependent model can now be summarized as

$$F_f = F_n + \phi(p)k_\kappa \tag{9.16}$$

where $F_n$ is one of the nominal models from Sec. 9.1. The attentive reader might expect that for appropriate choices of basis functions, the regressor matrix **A** will be rank-deficient. This is indeed the case since the $\text{sign}(v)$ column lies in the span of $\phi$. However, the interpretation of the resulting model coefficients is more intuitive if the Coulomb level is included as a baseline around which the basis-function expansion models the residuals. To mitigate the issue of rank deficiency, one could either estimate the nominal model first, and then fit the BFE to the residuals, or include a slight ridge-regression penalty on $k_\kappa$.

The above method is valid for position-varying Coulomb friction. It is conceivable that the position dependence is affected by the velocity, in which case the model (9.16) will produce a sub-optimal result. The RBF network can, however, be designed to cover the space $(\mathcal{P} \times \mathcal{V}) \subseteq \mathbb{R}^2$. The inclusion of velocity dependence comes at the cost of an increase in the number of parameters from $K_p$ to $K_p K_v$, where $K_p$ and $K_v$ denote the number of basis-function centers in the position and velocity input spaces, respectively.

The expression for the RBF kernel will in this extended model assume the form

$$\kappa(x, \mu, \Sigma) = \exp\left(-\frac{1}{2}(x-\mu)^\mathsf{T}\Sigma^{-1}(x-\mu)\right) \tag{9.17}$$

where $x = \begin{bmatrix} p & v \end{bmatrix}^\top \in \mathcal{P} \times \mathcal{V}, \mu \in \mathcal{P} \times \mathcal{V}$ and $\Sigma$ is the covariance matrix determining the bandwidth. The kernel vector will be

$$\phi(x)\colon\ (x \in \mathcal{P} \times \mathcal{V}) \to \mathbb{R}^{1 \times (K_p K_v)}$$
$$\phi(x) = \left[\kappa(x, \mu_1, \Sigma), \cdots, \kappa(x, \mu_{K_p K_v}, \Sigma)\right] \tag{9.18}$$

This concept extends to higher dimensions, at the cost of an exponential growth in the number of model parameters.

## 9.4 Energy-Dependent Model

Friction is often observed to vary with the temperature of the contact surfaces and lubricants involved [Bittencourt and Gunnarsson, 2012]. Many systems of industrial relevance lack the sensors needed to measure the temperature of the contact regions, thus rendering temperature-dependent models unusable.

The main contributor to the rise in temperature that occurs during operation is heat generated by friction. This section introduces a model that estimates the generated energy, and also estimates its influence on the friction.

A simple model for the temperature change in a system with temperature $T$, surrounding temperature $T_s$, and power input $W$, is given by

$$\frac{dT(t)}{dt} = k_s\big(T_s - T(t)\big) + k_W W(t) \tag{9.19}$$

for some constants $k_s > 0, k_W > 0$. After the variable change $\Delta T(t) = T(t) - T_s$, and transformation to the Laplace domain, the model (9.19) can be written

$$\Delta T_c(s) = \frac{k_W}{s + k_s} W_c(s) \tag{9.20}$$

where the power input generated by friction losses is equal to the product of the friction force and the velocity

$$W(t) = |F_f(t)v(t)| \tag{9.21}$$

We propose to include the estimated power lost due to friction, and its influence on friction itself, in the friction model according to

$$F_f = F_n + \operatorname{sign}(v)E \tag{9.22}$$

$$E_c(s) = G(s)W_c(s) = \frac{\bar{k}_e}{1 + s\bar{\tau}_e} W_c(s) \tag{9.23}$$

where the friction force $F_f$ has been divided into the nominal friction $F_n$ and the signal $E$, corresponding to the influence of the thermal energy stored in the joint. The nominal model $F_n$ can be chosen as any of the models previously introduced, including (9.16). The energy is assumed to be supplied by the instantaneous power due to friction, $W$, and is dissipating as a first order system with time constant $\bar{\tau}_e$. A discrete representation is obtained after Zero-Order-Hold (ZOH) sampling [Wittenmark et al., 2002] according to

$$E_d(z) = H(z)W_d(z) = \frac{k_e}{z - \tau_e} W_d(z) \tag{9.24}$$

In the suggested model form, (9.22) to (9.24), the transfer function $H(z)$ incorporates both the notion of energy being stored and dissipated, as well as the influence of the stored energy on the friction.

Denote by $\hat{\tau}_n$ the output of the nominal model $F_n$. Estimation of the signal $E$ can now be done by rewriting (9.22) in two different ways

$$\hat{E} = (\tau - \hat{\tau}_n)\text{sign}(v) \tag{9.25}$$

$$F_n = \tau - \text{sign}(v)\hat{E} \tag{9.26}$$

### Estimating the model

The joint estimation of the parameters in the nominal model and in $H(z)$ in (9.24) can be carried out in a fixed-point iteration scheme. This amounts to iteratively finding an estimate $\hat{F}_n$ of the nominal model, using $\hat{F}_n$ to find an estimate $\hat{E}$ of $E$ according to (9.25), using $\hat{E}$ to estimate $H(z)$ in (9.24) and, using $H(z)$, filter $\hat{E} = H(z)W$. An algorithm for the estimation of all parameters in (9.22) to (9.24) is given in Algorithm 6. The estimation of $\hat{H}(z)$ in (9.24) can be done with, e.g., the Output Error Method [Ljung, 1987; Johansson, 1993] and the estimation of the nominal model is carried out using the LS procedure from Sec. 9.2.

---

**Algorithm 6** Estimation of the parameters and the signal $E$ in the energy-dependent friction model.

---

**Require:** Initial estimate $\hat{H}(z, k_e, \tau_e)$;
  **repeat**
    $\hat{E} \leftarrow \hat{H}(z)W$                         ▷ Filter $W$ through $\hat{H}(z)$;
    Update estimate of $F_n$ according to (9.26) using (6.5);
    Calculate $\hat{E}$ according to (9.25);
    Update $\hat{H}(z)$ using (9.24)            ▷ E.g., command oe() in Matlab;
  **until** Convergence

---

The proposed model suggests that the change in friction due to the temperature change occurs in the Coulomb friction. This assumption is always valid for the nominal model (9.1), and a reasonable approximation for the model (9.2) if $k_c \gg k_v v$ or if the system is both operated and identified in a small interval of velocities. If, however, the temperature change has a large effect on the viscous friction or on the position dependence, a 3D basis-function expansion can be performed in the space $\mathcal{P} \times \mathcal{V} \times \mathcal{E}$, $E \in \mathcal{E}$. This general model can handle arbitrary nonlinear dependencies between position, velocity and estimated temperature. The energy signal $E$ can then be estimated using a simple nominal model, and included in the kernel expansion for an extended model. Further discussion on this is held in Sec. 9.7.

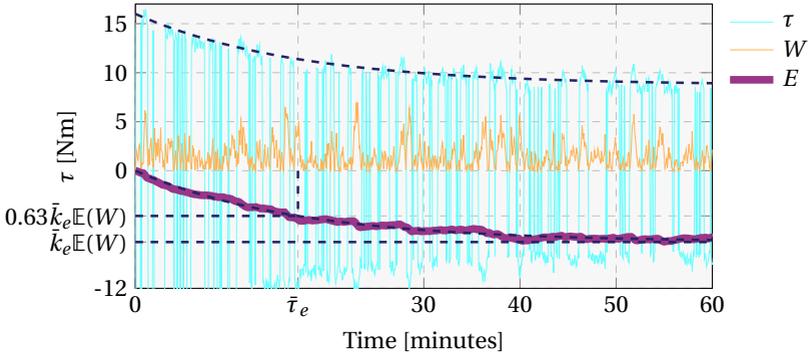

**Figure 9.3**   A realization of simulated signals. The figure shows how the envelope of the applied torque approximately decays as the signal $E$. Dashed, blue lines are drawn to illustrate the determination of initial guesses for the time constant $\bar{\tau}_e$ and the gain $\bar{k}_e$.

**Initial guess**

For this scheme to work, an initial estimate of the parameters in $H(z)$ is needed. This can be easily obtained by observing the raw torque data from an experiment. Consider for example Fig. 9.3, where the system (9.22) and (9.23) has been simulated. The figure depicts the torque signal as well as the energy signal $E$. The envelope of the torque signal decays approximately as the signal $E$, which allows for easy estimation of the gain $\bar{k}_e$ and the time constant $\bar{\tau}_e$. The time constant $\bar{\tau}_e$ is determined by the time it takes for the signal to reach $(1 - e^{-1}) \approx 63\,\%$ of its final value. Since $G(s)$ is essentially a low-pass filter, the output $E = G(s)W$ will approximately reach $E_\infty = G(0)\mathbb{E}(W) = \bar{k}_e\mathbb{E}(W)$ if sent a stationary, stochastic input $W$ with fast enough time constant $(\ll \bar{\tau}_e)$. Here, $\mathbb{E}(\cdot)$ denotes the statistical expectation operator and $E_\infty$ is the final value of the signal $E$. An initial estimate of the gain $\bar{k}_e$ can thus be obtained from the envelope of the torque signal as

$$\bar{k}_e \approx \frac{E_\infty}{\mathbb{E}(W)} \approx \frac{E_\infty}{\frac{1}{N}\sum_n W_n} \tag{9.27}$$

We refer to Fig. 9.3 for an illustration, where dashed guides have been drawn to illustrate the initial guesses.

The discrete counterpart to $G(s)$ can be obtained by discretization with relevant sampling time [Wittenmark et al., 2002].

## 9.5 Simulations

To analyze the validity of the proposed technique for estimation of the energy-dependent model, a simulation was performed. The system described by (9.22) and (9.23) was simulated to create 50 realizations of the relevant signals, and the proposed method was run for 50 iterations to identify the model parameters. The parameters used in the simulation are provided in Table 9.1. Initial guesses were chosen at random from the uniform distributions $\hat{\bar{k}}_e \sim \mathcal{U}(0, 3\bar{k}_e)$ $\hat{\bar{\tau}}_e \sim \mathcal{U}(0, 3\bar{\tau})$.

**Table 9.1**  Parameter values used in simulation. Values given on the format $x/y$ represent continuous/discrete values.

| Parameter | Value |
| --- | --- |
| $k_v$ | 5 |
| $k_c$ | 15 |
| $k_e$ | -3/-0.5 |
| $\tau_e$ | 10/0.9983 |
| Measurement noise $\sigma_\tau$ | 0.5 Nm |
| Sample time $h$ | 1 s |
| Duration | 3600 s |
| Iterations | 50 |

Figure 9.4 shows that the estimated parameters converge rapidly to their true values, and Fig. 9.5 indicates that the Root Mean Square output Error (RMSE) converges to the level of the added measurement noise. Figure 9.5 further shows that the errors in the parameter estimates, as defined by (9.28), were typically below 5 % of the parameter values.

$$\text{NPE} = \sqrt{\sum_{i=1}^{N_p} \left( \frac{\hat{x}_i - x_i}{|x_i|} \right)^2} \tag{9.28}$$

## 9.6 Experiments

The proposed models and identification procedures were applied to data from experiments with the dual-arm and the IRB140 industrial robots, see Fig. 9.2.

### Procedure

For IRB140, the first joint was used. The rest of the arms were positioned so as to minimize the moment of inertia. For the dual-arm robot, joint four in one of the arms was positioned such that the influence of gravity vanished.

A program that moved the selected joint at piecewise constant velocities between the two joint limits was executed for approximately 20 min. Torque-,

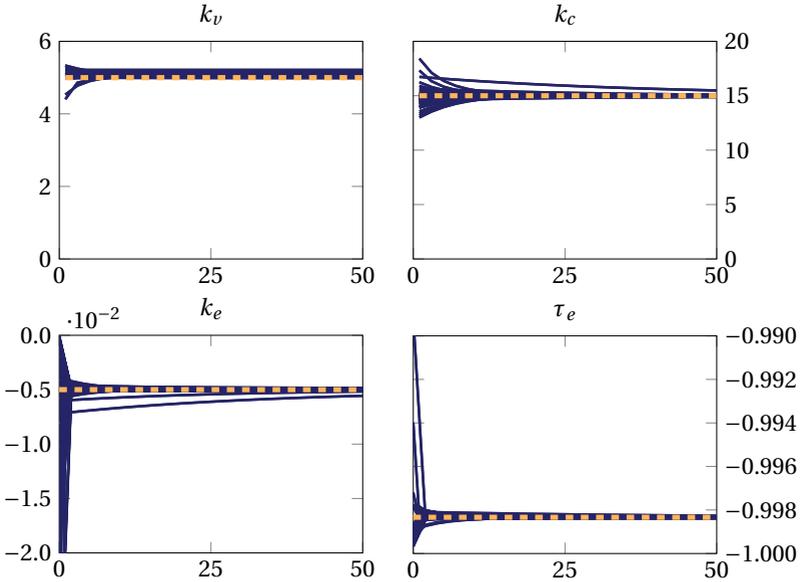

**Figure 9.4** Estimated parameters during 50 simulations. The horizontal axis displays the iteration number and the vertical axis the current parameter value. True parameter values are indicated with dashed lines.

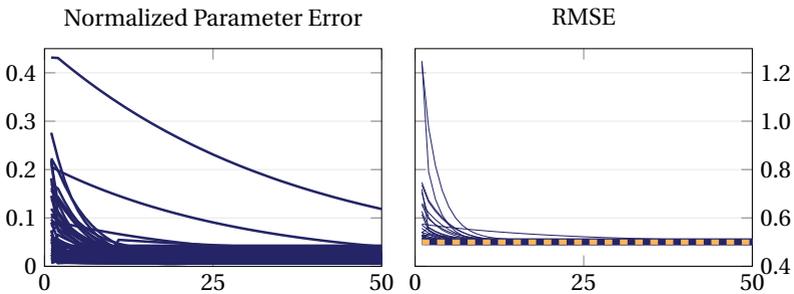

**Figure 9.5** Evolution of errors during the simulations, the horizontal axis displays the iteration number. The left plot shows normalized norms of parameter errors, defined in (9.28), and the right plot shows the RMS output error using the estimated parameters. The standard deviation of the added measurement noise is shown with a dashed line.

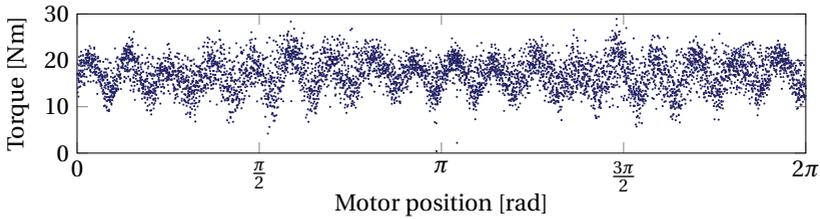

**Figure 9.6** Illustration of the torque dependence upon the motor position for the IRB140 robot.

velocity-, and position data were sampled and filtered at 250 Hz and subsequently sub-sampled and stored at 20 Hz, resulting in 25 000 data points. Points approximately satisfying (9.9) were selected for identification, resulting in a set of 16 000 data points.

***Nominal Model*** The viscous model (9.3) was fit using the ordinary LS procedure from Sec. 9.2. This model was also used as the nominal model in the subsequent fitting of position model (9.16) and energy model, (9.22) to (9.24).

***Position model*** For the position-dependent model, the number of basis functions and their bandwidth was determined using cross validation. A large value of $\sigma$ has a strong regularizing effect and resulted in a model that generalized well outside the training data. The model was fit using normalized basis functions as discussed in Sec. 6.3.

Due to the characteristics of the gear box and electrical motor in many industrial robots, there is a clear dependence not only on the arm position, but also on the motor position. Figure 9.6 shows the torque versus the motor position when the joint is operated at constant velocity. This is especially strong on the IRB140 and results are therefore illustrated for this robot. Both arm and motor positions are available through the simple relationship $p_{motor} = \text{mod}_{2\pi}(g \cdot p_{arm})$, where $g$ denotes the gear ratio. This allows for a basis-function expansion also in the space of motor positions. To illustrate this, $p_{motor}$ was expanded into $K_{p_m}K_v = 36 \times 6$ basis functions, corresponding to the periodicity observed in Fig. 9.6. The results for the model with motor-position dependence are reported separately. Further modeling and estimation of the phenomena observed in Fig. 9.6 is carried out in Chap. 10, where a spectral estimation technique is developed, motivated by the observation that the spectrum of the signal in Fig. 9.6 is modulated by the velocity of the motor.

To reduce variance in the estimated kernel parameters, all position-dependent models were estimated using ridge regression (Sec. 6.4), where a $L_2$-penalty was put on the kernel parameters. The strength of the penalty was determined using cross validation. All basis-function expansions were performed with normalized basis functions.

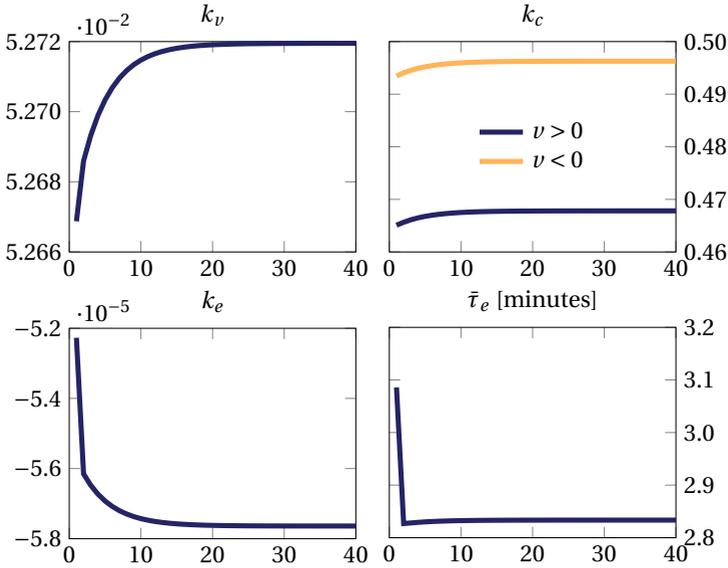

**Figure 9.7** Estimated parameters from experimental data. The horizontal axis displays the iteration number and the vertical axis displays the current parameter value.

**Table 9.2** Performance indicators for the three different models identified on the dual-arm robot.

|      | Nominal  | Position | Position + Energy |
| ---- | -------- | -------- | ----------------- |
| Fit  | 86.968   | 93.193   | 96.674            |
| FPE  | 3.63e-03 | 1.03e-03 | 2.65e-04          |
| RMSE | 6.03e-02 | 3.15e-02 | 1.54e-02          |
| MAE  | 4.71e-02 | 2.36e-02 | 1.22e-02          |

***Energy model*** The energy-dependent model was identified for the dual-arm robot using the procedure described in Algorithm 6. The initial guesses for $H(z)$ were $\bar{\tau}_e = 10\,\text{min}$ and $\bar{k}_e = -0.1$. The nominal model was chosen as the viscous friction model (9.3). Once the signal $E$ was estimated, a kernel expansion in the space $\mathcal{P} \times \mathcal{V} \times \mathcal{E}$ with $40 \times 6 \times 3$ basis functions was performed to capture temperature-dependent effects in both the Coulomb and viscous friction parameters.

### Results

The convergence of the model parameters is shown in Fig. 9.7. Figure 9.8 and Fig. 9.9 illustrate how the models identified for the dual-arm robot fit the exper-imental data. The upper plot in Fig. 9.8 shows an early stage of the experiment when the joint is cold. At this stage, the model without the energy term underestimates the torque needed, whereas the energy model does a better job. The lower plot shows a later stage of the experiment where the mean torque level is significantly lower. Here, the model without energy term is instead slightly overestimating the friction torque. The observed behavior is expected, since the model without energy dependence will fit the average friction level during the entire experiment. The two models correspond well in the middle of the experiments (not shown). Figure 9.9 illustrates the friction torque predicted by the estimated model as a function of position and velocity. The visible rise in the surface at large positive positions and positive velocities corresponds to the increase in friction torque observed in Fig. 9.8 at, e.g., time $t = 0.3$ min.

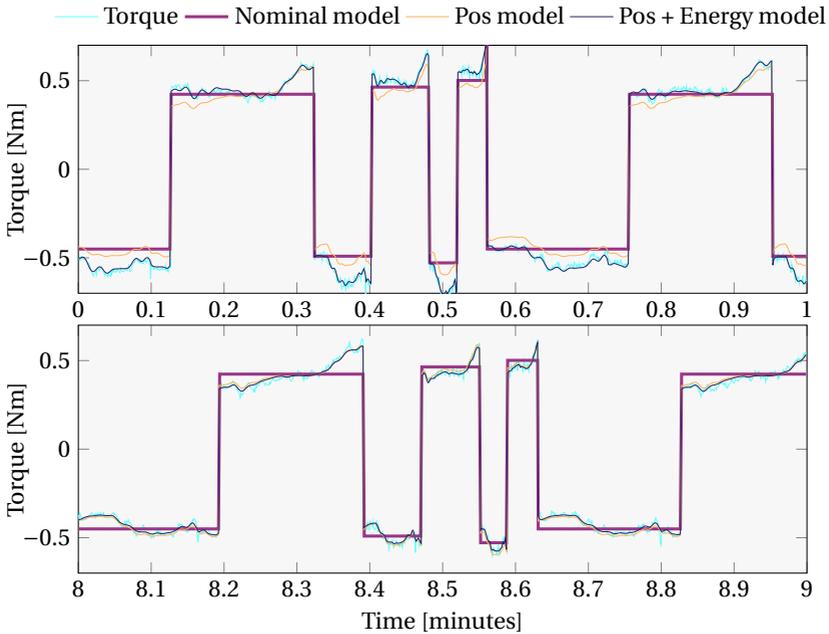

**Figure 9.8**    Model fit to experimental data (dual-arm). Upper plot shows an early stage of the experiment when the joint is cold. Lower plot a later stage, when the joint has been warmed up.

The nominal model (9.3), can not account for any of the positional effects and produces an overall, much worse fit than the position dependent models. Different measures of model fit for the three models are presented in Table 9.2 and Fig. 9.11 (Fit (%), Final Prediction Error, Root Mean Square Error, Mean Absolute Error). For definitions, see e.g., [Johansson, 1993].

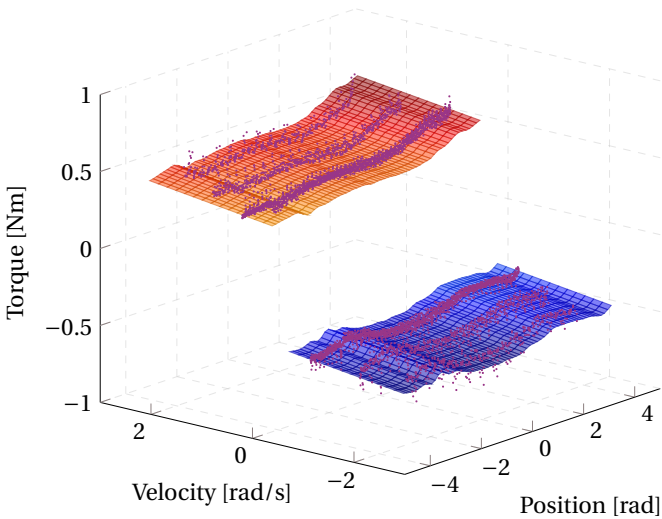

**Figure 9.9** Estimated position-dependent model for dual-arm (discontinuous surface) together with the datapoints used for estimation. A Coloumb + viscous model would consist of two flat surfaces, whereas the position-dependent model has uncovered a more complicated structure.

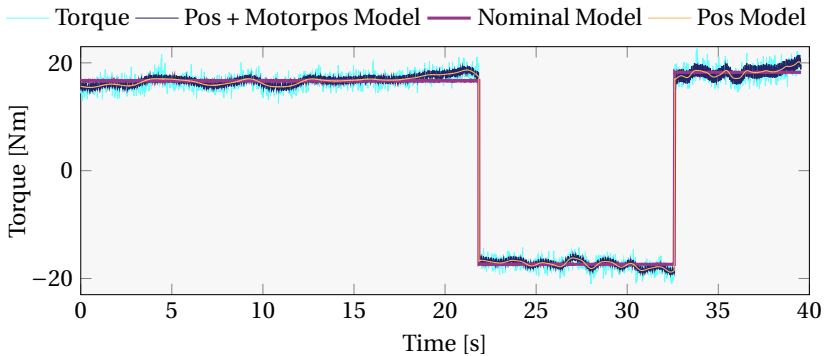

**Figure 9.10** Model fit including kernel expansion for motor position on IRB140. During $t = [0\,\mathrm{s}, 22\,\mathrm{s}]$, the joint traverses a full revolution of $2\pi$ rad. The same distance was traversed backwards with a higher velocity during $t = [22\,\mathrm{s}, 33\,\mathrm{s}]$. Notice the repeatable pattern as identified by the position-dependent models.

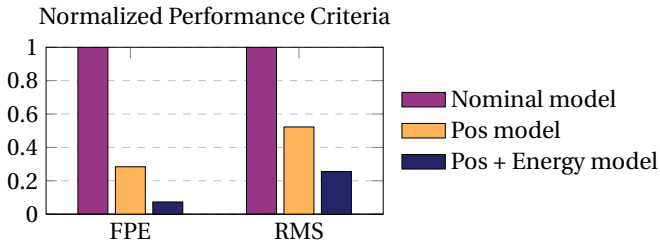

**Figure 9.11** Performance indicators for the identified models, dual-arm.

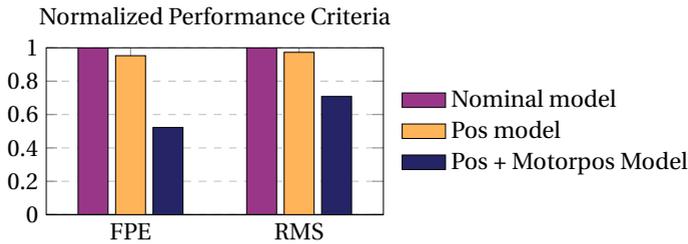

**Figure 9.12** Performance indicators for the identified models, IRB140.

For the IRB140, three models are compared. The nominal model (9.3), a model with a basis-function expansion in the space $\mathcal{P}_{arm}$, and a model with an additional basis-function expansion in the space $\mathcal{P}_{motor} \times \mathcal{V}$. The resulting model fits are shown in Fig. 9.10. What may seem like random measurement noise in the torque signal is in fact predictable using a relatively small set of parameters. Figure 9.12 illustrates that the large dependence of the torque on the motor position results in large errors. The inclusion of a basis-function expansion of the motor position in the model reduces the error significantly.

## 9.7 Discussion

The proposed models try to increase the predictive power of common friction models, and thereby increase their utility for model-based filtering, by incorporating position- and temperature dependence into the friction model. Systems with varying parameters can, in theory, be estimated with recursive algorithms, so called online identification. As elaborated upon in Sec. 9.1, online or observer-based identification of friction models is often difficult in practice due to the presence of additional dynamics or external forces. The proposed models are identified offline, during a controlled experiment, and are thus not subject to the problems associated with online identification. However, apart from the temperature-related parameters, all suggested models are linear in the parameters, and could be updated recursively using, for instance, the well-known recursive least-squares algorithm or the Kalman-smoothing algorithms in Chap. 7.

This paper makes use of standard and well-known models for friction, combined with a basis-function expansion to model position dependence. This choice was motivated by the large increase in model accuracy achieved for a relatively small increase in model complexity. Linear models are easy to estimate and the solution to the least-squares optimization problem is well understood. Depending on the intended use of the friction model, the most fruitful avenue to investigate in order to increase the model accuracy further varies. To the purpose of force estimation, accurate models of the stiction force are likely important. Stationary joints impose a fundamental limitation in the accuracy of the force estimate, and the maximum stiction force determines the associated uncertainty of the estimate. Preliminary work shows that the problem of indeterminacy of the friction force for static joints of redundant manipulators can be mitigated by superposition of a periodic motion in the nullspace of the manipulator Jacobian. Exploration of this remains an interesting avenue for future work.

Although outside the scope of this work, effects of joint load on the friction behavior can be significant [Bittencourt and Gunnarsson, 2012]. Such dependencies could be incorporated in the proposed models using the same RBF approach as for the incorporation of position dependence, i.e., through an RBF expansion in the joint load ($l \in \mathcal{L}$) dimension according to $\phi(x) : (x \in \mathcal{P} \times \mathcal{E} \times \mathcal{L}) \mapsto \mathbb{R}^{1 \times (K_p K_e K_l)}$, with $K_l$ basis-function centers along dimension $\mathcal{L}$. This strategy would capture possible position and temperature dependencies in the load-friction interaction.

The temperature-dependent part of the proposed model originates from the most simple possible model for energy storage, a generic first order differential equation. Since the generated energy is initially unknown, incorporating it in the model is not straight forward. We rely on the assumption that a simple initial friction model can be estimated without this effect and subsequently be used to estimate the generated energy loss. The energy loss estimated by this model can then be incorporated in a more complex model. Iterating this scheme was shown to converge in simulations, but depending on the conditions, the scheme might diverge. This might happen if, e.g., the friction varies significantly with temperature, where significantly is taken as compared to the nominal friction value at room temperature. In such situations, the initially estimated model will be far from the optimum, reducing the chance of convergence. In practice, this issue is easily mitigated by estimating the initial model only on data that comes from the joint at room temperature.

In its simplest form, the proposed energy-dependent model assumes that the change in friction occurs in the Coulomb friction level. This is always valid for the Coulomb model, and a reasonable approximation for the viscous friction model if $k_c \gg k_v v$ or if the system is both operated and identified in a small interval of velocities. If the viscous friction $k_v v$ is large, the approximation will be worse. This suggests modeling the friction as

$$F_f = k_v(E)\,v + k_c(E)\,\text{sign}\,(v) \tag{9.29}$$

where the Coulomb and viscous constants are seen as functions of the estimated energy signal $E$, i.e., a Linear Parameter-Varying model (LPV). To accomplish this, a kernel expansion including the estimated energy signal was suggested and evaluated experimentally.

Although models based on the internally generated power remove the need for temperature sensing in some scenarios, they do not cover significant variations in the surrounding temperature. The power generated in, for instance, an industrial robot is, however, often high enough to cause a much larger increase in temperature than the expected temperature variations of its surrounding [Bittencourt and Gunnarsson, 2012].

## 9.8  Conclusions

The modeling of both position and temperature dependence in systems with friction has been investigated. To model position varying friction, a basis-function expansion approach was adopted. It has been experimentally verified that taking position dependence into account can significantly reduce the model output error. It has also been reported that friction phenomena on both sides of a gearbox can be modeled using the proposed approach.

Further, the influence of an increase in temperature due to power generated by friction has been modeled and estimated. The proposed approach was based on a first-order temperature input-output model where the power generated by friction was used as input. The model together with the proposed identification procedure was shown to capture the decrease in friction seen in an industrial robot during a long-term experiment, this was accomplished without the need of temperature sensing.

# 10

# Spectral Estimation

## 10.1 Introduction

Spectral estimation refers to a family of methods that analyze the frequency contents of a sampled signal by means of decomposition into a linear combination of periodic basis functions. Armed with an estimate of the spectrum of a signal, it is possible to determine the distribution of power among frequencies, identify disturbance components and design filters, etc. [Stoica and Moses, 2005]. The spectrum also serves as a powerful feature representation for many classification algorithms [Bishop, 2006], e.g., by looking at the spectrum of a human voice recording it is often trivial to distinguish male and female speakers from each other, or tell if a saw-blade is in need of replacement by analyzing the sound it makes while cutting.

Standard spectral density-estimations techniques such as the discrete Fourier transform (DFT) exhibit several well-known limitations. These methods are typically designed for data sampled equidistantly in time or space. Whenever this fails to hold, typical approaches employ some interpolation technique in order to perform spectral estimation on equidistantly sampled data. Other possibilities include employing a method suitable for nonequidistant data, such as least-squares spectral analysis [Wells et al., 1985]. Fourier transform-based methods further suffer from spectral leakage due to the assumption that all sinusoidal basis functions are orthogonal over the data window [Puryear et al., 2012]. Least-squares spectral estimation takes the correlation of the basis functions into account and further allows for estimation of arbitrary/known frequencies without modification [Wells et al., 1985].

In some applications, the spectral content is varying with an external variable, for instance, a controlled input. As a motivating example, we consider the torque ripple induced by the rotation of an electrical motor. Spectral analysis of the torque signal is made difficult by the spectrum varying with the velocity of the motor, both due to the frequency of the ripple being directly proportional to the velocity, but also due to the properties of an electric DC-motor. A higher velocity both induces higher magnitude torque ripple, but also a higher filtering effect due to the inertia of the rotating parts. The effect of a sampling delay on the phase of the measured ripple is similarly proportional to the velocity.

Time-frequency analysis traditionally employs windowing techniques [Johansson, 1993] in order to reduce spectral leakage [Harris, 1978; Stoica and Moses, 2005], mitigate effects of non-stationarity, reduce the influence of ill-posed autocorrelation estimates [Stoica and Moses, 2005], and allow for time-varying spectral estimates [Puryear et al., 2012]. The motivating example considers estimation of the spectral content of a signal that is periodic over the space of angular positions $\mathcal{X}$. The spectral content will vary with time solely due to the fact that the velocity is varying with time. Time does thus not hold any intrinsic meaning to the modification of the spectrum, and the traditional windowing in time is no longer essential.

This chapter develops a spectral-estimation technique using basis-function expansions that allows the spectral properties (phase and amplitude) of the analyzed signal to vary with an auxiliary signal. Apart from a standard spectrum, functional relationships between the scheduling signal and the amplitude and phase of each frequency will be identified. We further consider the task of sparse spectral estimation and show how the proposed method extends also to this setting.

## 10.2  LPV Spectral Decomposition

In order to decompose the spectrum along an external dimension, we consider basis-function expansions, a topic that was introduced in detail in Sec. 6.3 and used for friction-modeling in Chap. 9. Intuitively, a basis-function expansion decomposes an intricate function or signal as a linear combination of simple basis functions. The Fourier transform can be given this interpretation, where an arbitrary signal is decomposed as a sum of complex-valued sinusoids. With this intuition, we aim for a method that allows decomposition of the spectrum of a signal along an external dimension, in LPV terminology called the scheduling dimension, $\mathcal{V}$.[1] If we consider a single sinusoid in the spectrum, the function decomposed by the basis-function expansion will thus be the complex-valued coefficient $k$ in $ke^{i\omega}$ as a function of the scheduling variable, $v$, i.e., $k = k(v)$. In the motivating example, $v$ is the angular velocity of the motor. Using complex-valued calculations, we simultaneously model the dependence of both amplitude and phase of a real frequency by considering the complex frequency. This parameterization will also result in an estimation problem that is linear in the parameters, which is not true for the problem of estimating phase and amplitude directly.

For additional details on the topic of basis-function expansions, the reader is referred to Sec. 6.3.

---

[1] We limit our exposition to $\mathcal{V} \subseteq \mathbb{R}$ for clarity, but higher dimensional scheduling spaces are possible.

**Least-squares identification of periodic signals**

Spectral estimation amounts to estimation of models of a signal $y$ on the form

$$y(n) = k_1 \sin(\omega n) + k_2 \cos(\omega n) \tag{10.1}$$

which are linear in the parameters $k$. Identification of linear models using the least-squares procedure was described in Sec. 6.2. The model (10.1) can be written in compact form by noting that $e^{i\omega} = \cos\omega + i\sin\omega$, which will be used extensively throughout the chapter to simplify notation.[2]

   We will now proceed to formalize the proposed spectral-decomposition method.

**Signal model**

Our exposition in this chapter will make use of Gaussian basis functions. The method is, however, not limited to this choice and extends readily to any other set of basis functions. A discussion on different choices is held in Sec. 6.3.

   We start by establishing some notation. Let $k$ denote the Fourier-series coefficients[3] of interest. The kernel activation vector $\phi(v_i) : (v \in \mathcal{V}) \mapsto \mathbb{R}^J$ maps the input to a set of basis-function activations and is given by

$$\phi(v_i) = \begin{bmatrix} \kappa(v_i, \theta_1) & \cdots & \kappa(v_i, \theta_J) \end{bmatrix}^\mathsf{T} \in \mathbb{R}^J \tag{10.2}$$

$$\kappa(v, \theta_j) = \kappa_j(v) = \exp\left(-\frac{(v-\mu)^2}{2\sigma^2}\right) \tag{10.3}$$

where $\kappa_j$ is a basis function parameterized by $\theta_j = (\mu_j, \sigma_j)$, $\mu \in \mathcal{V}$ is the center of the kernel and $\sigma^2$ is determining the width.

   Let $y$ denote the signal to be decomposed and denote the location of the sampling of $y_i$ by $x_i \in \mathcal{X}$. The space $\mathcal{X}$ is commonly time or space; in the motivating example of the electrical motor, $\mathcal{X}$ is the space of motor positions.[4] Let the intensities of a set of complex frequencies $i\omega \; \forall \; \omega \in \Omega$ be given by basis-function expansions along $\mathcal{V}$, according to

$$\hat{y}_i = \sum_{\omega \in \Omega} \sum_{j=1}^{J} k_{\omega,j} \kappa_j(v_i) e^{-i\omega x_i} = \sum_{\omega \in \Omega} k_\omega^\mathsf{T} \phi(v_i) e^{-i\omega x_i}, \quad k_\omega \in \mathbb{C}^J \tag{10.4}$$

The complex coefficients to be estimated, $k \in \mathbb{C}^{O \times J}$, $O = \mathrm{card}(\Omega)$, constitute the Fourier-series coefficients, where the intensity of each coefficient is decomposed over $\mathcal{V}$ through the BFE. This formulation reduces to the standard Fourier-style spectral model (10.5) in the case $\phi(v) \equiv 1$

$$\hat{y} = \sum_{\omega \in \Omega} k_\omega e^{-i\omega x} = \Phi k \tag{10.5}$$

where $\Phi = [e^{-i\omega_1 x} \dots e^{-i\omega_O x}]$. If the number $J$ of basis functions equals the number of data points $N$, the model will exactly interpolate the signal, i.e., $\hat{y} = y$. If in addition to $J = N$, the basis-function centers are placed at $\mu_j = v_j$, we obtain a Gaussian process regression interpretation where $\kappa$ is the covariance function. Owing to the numerical properties of the analytical solution of the least-squares problem, it is often beneficial to reduce the number of parameters significantly, so that $J \ll N$. If the chosen basis functions are suitable for the signal of interest, the error induced by this dimensionality reduction is small. In a particular case, the number of RBFs to include, $J$, and the bandwidth $\Sigma$ is usually chosen based on evidence maximization or cross validation [Murphy, 2012].

---

[2] Note that solving the complex LS problem using complex regressors $e^{i\omega}$ is not equivalent to solving the real LS problem using sin/cos regressors.

[3] We use the term Fourier-series coefficients to represent the parameters in the spectral decomposition, even if the set of basis functions are not chosen so as to constitute a true Fourier series.

[4] We note at this stage that $x \in \mathcal{X}$ can be arbitrarily sampled and are not restricted to lie on an equidistant grid, as is the case for, e.g., Fourier transform-based methods.

To facilitate estimation of the parameters in (10.4), we rewrite the model by stacking the regressor vectors in a regressor matrix **A**, see Sec. 10.2, such that

$$\mathbf{A}_{n,:} = \text{vec}\big(\phi(v_n)\Phi^\mathsf{T}\big)^\mathsf{T} \in \mathbb{C}^{O \cdot J}, n = 1 \dots N$$

We further define $\bar{\mathbf{A}}$ by expanding the regressor matrix into its real and imaginary parts

$$\bar{\mathbf{A}} = \begin{bmatrix} \Re\mathbf{A} & \Im\mathbf{A} \end{bmatrix} \in \mathbb{R}^{N \times 2OJ}$$

such that routines for real-valued least-squares problems can be used. The complex coefficients are, after solving the real-valued least-squares problem[5] retrieved as $k = k_\Re + i k_\Im$ where

$$[k_\Re^\mathsf{T} \quad k_\Im^\mathsf{T}]^\mathsf{T} = \underset{\bar{k}}{\arg\min} \left\| \bar{\mathbf{A}}\bar{k} - y \right\|$$

Since the purpose of the decomposition is spectral analysis, it is important to normalize the basis-function activations such that the total activation over $\mathcal{V}$ for each data point is unity. To this end, the expressions (10.4) are modified to

$$\hat{y} = \sum_{\omega \in \Omega} \sum_{j=1}^J k_{\omega,j} \bar{\kappa}_j(v) e^{-i\omega x} = \sum_{\omega \in \Omega} k_\omega^\mathsf{T} \bar{\phi}(v) e^{-i\omega x}$$

$$\bar{\kappa}_j(v) = \frac{\kappa_j(v)}{\sum_j \kappa_j(v)}, \quad \bar{\phi}(v) = \frac{\phi(v)}{\sum \phi(v)} \tag{10.6}$$

This ensures that the spectral content for a single frequency $\omega$ is a convex combination of contributions from each basis function in the scheduling dimension. Without this normalization, the power of the spectrum would be ill-defined and depend on an arbitrary scaling of the basis functions. The difference between a set of Gaussian functions and a set of normalized Gaussian functions is demonstrated in Fig. 10.1. The normalization performed in (10.6) can be viewed as the kernel function being made data adaptive by normalizing $\phi(v)$ to sum to one.

---

[5] See  Sec. 6.2 for details on the least-squares procedure.

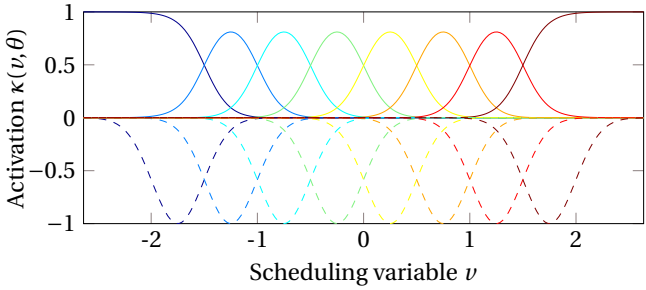

**Figure 10.1**  Gaussian (dashed) and normalized Gaussian (solid) windows. Regular windows are shown mirrored in the *x*-axis for clarity.

### Amplitude and phase functions

In spectral analysis, two functions of the Fourier-series coefficients are typically of interest, the amplitude and phase functions. These are easily obtained through elementary trigonometry and are stated here as a lemma, while a simple proof is deferred until Sec. 10.A:

LEMMA 3
Let a signal $y$ be composed by the linear combination $y = k_1 \cos(x) + k_2 \sin(x)$, then $y$ can be written on the form

$$y = A\cos(x - \varphi)$$

with

$$A = \sqrt{k_1^2 + k_2^2} \qquad \varphi = \arctan\left(\frac{k_2}{k_1}\right) \qquad \qquad \square$$

From this we obtain the following two functions for a particular frequency $\omega$

$$A(\omega) = |k_\omega| = \sqrt{\Re k_\omega^2 + \Im k_\omega^2}$$
$$\varphi(\omega) = \arg(k_\omega) = \arctan(\Im k_\omega / \Re k_\omega)$$

In the proposed spectral-decomposition method, these functions further depend on $\nu$, and are approximated by

$$A(\omega, \nu) = \left|\sum_{j=1}^{J} k_{\omega,j}\, \bar{\kappa}(\nu)\right| = \left|k_\omega^\top \bar{\phi}(\nu)\right| \tag{10.7}$$

$$\varphi(\omega, \nu) = \arg\left(\sum_{j=1}^{J} k_{\omega,j}\, \bar{\kappa}(\nu)\right) = \arg\left(k_\omega^\top \bar{\phi}(\nu)\right) \tag{10.8}$$

**Covariance properties**

We will now investigate and prove that (10.7) and (10.8) lead to asymptotically unbiased and consistent estimates of $A$ and $\varphi$, and will provide a strategy to obtain confidence intervals. We will initially consider a special case for which analysis is simple, whereafter we invoke the RBF universal approximation results of Park and Sandberg (1991) to show that the estimators are well motivated for a general class of functions. We start by considering signals on the form (10.9), for which unbiased and consistent estimates of the parameters are readily available:

PROPOSITION 3
Let a signal $y$ be given by

$$
\begin{aligned}
y &= a\cos(x) + b\sin(x) + e & e &\in \mathcal{N}(0, \sigma^2) \\
a &= \alpha^{\mathsf{T}}\phi & b &= \beta^{\mathsf{T}}\phi
\end{aligned}
\tag{10.9}
$$

with $\phi = \phi(v)$ and let $\hat{\alpha}$ and $\hat{\beta}$ denote unbiased estimates of $\alpha$ and $\beta$. Then

$$
\hat{A}(\hat{\alpha}, \hat{\beta}) = \sqrt{\left(\hat{\alpha}^{\mathsf{T}}\phi\right)^2 + \left(\hat{\beta}^{\mathsf{T}}\phi\right)^2}
\tag{10.10}
$$

has an expected value with upper and lower bounds given by

$$
A < \mathbb{E}\{\hat{A}\} < \sqrt{A^2 + \phi^{\mathsf{T}}\Sigma_{\alpha}\phi + \phi^{\mathsf{T}}\Sigma_{\beta}\phi}
\tag{10.11}
$$

***Proof.*** Since $\alpha$, $\beta$ and $e$ appear linearly in (10.9), unbiased and consistent estimates $\hat{\alpha}$ and $\hat{\beta}$ are available from the least-squares procedure (see Sec. 6.2). The expected value of $\hat{A}^2$ is given by

$$
\begin{aligned}
\mathbb{E}\{\hat{A}^2\} &= \mathbb{E}\left\{\left(\hat{\alpha}^{\mathsf{T}}\phi\right)^2 + \left(\hat{\beta}^{\mathsf{T}}\phi\right)^2\right\} \\
&= \mathbb{E}\left\{\left(\hat{\alpha}^{\mathsf{T}}\phi\right)^2\right\} + \mathbb{E}\left\{\left(\hat{\beta}^{\mathsf{T}}\phi\right)^2\right\}
\end{aligned}
\tag{10.12}
$$

We further have

$$
\begin{aligned}
\mathbb{E}\left\{\left(\hat{\alpha}^{\mathsf{T}}\phi\right)^2\right\} &= \mathbb{E}\{\hat{\alpha}^{\mathsf{T}}\phi\}^2 + \mathbb{V}\{\hat{\alpha}^{\mathsf{T}}\phi\} \\
&= \left(\alpha^{\mathsf{T}}\phi\right)^2 + \phi^{\mathsf{T}}\Sigma_{\alpha}\phi
\end{aligned}
\tag{10.13}
$$

where $\Sigma_{\alpha}$ and $\Sigma_{\beta}$ are the covariance matrices of $\hat{\alpha}$ and $\hat{\beta}$ respectively. Calculations for $\beta$ are analogous. From (10.12) and (10.13) we deduce

$$
\begin{aligned}
\mathbb{E}\{\hat{A}^2\} &= \left(\alpha^{\mathsf{T}}\phi\right)^2 + \left(\beta^{\mathsf{T}}\phi\right)^2 + \phi^{\mathsf{T}}\Sigma_{\alpha}\phi + \phi^{\mathsf{T}}\Sigma_{\beta}\phi \\
&= A^2 + \phi^{\mathsf{T}}\Sigma_{\alpha}\phi + \phi^{\mathsf{T}}\Sigma_{\beta}\phi
\end{aligned}
\tag{10.14}
$$

Now, due to Jensen's inequality, we have

$$
\mathbb{E}\{\hat{A}\} = \mathbb{E}\left\{\sqrt{\hat{A}^2}\right\} < \sqrt{\mathbb{E}\{\hat{A}^2\}}
\tag{10.15}
$$

which provides the upper bound on the expectation of $\hat{A}$. The lower bound is obtained by writing $\hat{A}$ on the form

$$\hat{A}(k) = \sqrt{\left(\hat{\alpha}^\mathsf{T}\phi\right)^2 + \left(\hat{\beta}^\mathsf{T}\phi\right)^2} = \left\|\hat{k}\right\| \tag{10.16}$$

with $\hat{k} = [\hat{\alpha}^\mathsf{T}\phi \quad \hat{\beta}^\mathsf{T}\phi]$. From Jensen's inequality we have

$$\mathbb{E}\{\hat{A}\} = \mathbb{E}\{\|\hat{k}\|\} > \|\mathbb{E}\{\hat{k}\}\| = \|k\| = A \tag{10.17}$$

which concludes the proof. $\qquad\square$

COROLLARY 2

$$\hat{A} = \sqrt{\left(\hat{\alpha}^\mathsf{T}\phi\right)^2 + \left(\hat{\beta}^\mathsf{T}\phi\right)^2} \tag{10.18}$$

is an asymptotically unbiased and consistent estimate of $A$.

*Proof.* Since the least-squares estimate, upon which the estimated quantity is based, is unbiased and consistent, the variances in the upper bound in (10.11) will shrink as the number of datapoints increases and both the upper and lower bounds will become tight, hence

$$\mathbb{E}\{\hat{A}\} \to A \quad \text{as} \quad N \to \infty$$

Analogous bounds for the phase function are harder to obtain, but the simple estimator $\hat{\varphi} = \arg(\hat{k})$ based on $\hat{k}$ obtained from the least-squares procedure is still asymptotically consistent [Kay, 1993].

Estimates using the least-squares method (6.5) are, under the assumption of uncorrelated Gaussian residuals of variance $\sigma^2$, associated with a posterior parameter covariance $\sigma^2(\mathbf{A}^\mathsf{T}\mathbf{A})^{-1}$. This will in a straightforward manner produce confidence intervals for a future prediction of $y$ as a linear combination of the estimated parameters. Obtaining unbiased estimates of the confidence intervals for the functions $A(v,\omega)$ and $\varphi(v,\omega)$ is made difficult by their nonlinear nature. We therefore proceed to establish an approximation strategy.

The estimated parameters $\hat{k}$ are distributed according to a complex-normal distribution $\mathcal{CN}(\Re z + i\Im z, \Gamma, C)$, where $\Gamma$ and $C$ are obtained through

$$\begin{aligned}
\Gamma &= \Sigma_{\Re\Re} + \Sigma_{\Im\Im} + i(\Sigma_{\Im\Re} - \Sigma_{\Re\Im}) \\
C &= \Sigma_{\Re\Re} - \Sigma_{\Im\Im} + i(\Sigma_{\Im\Re} + \Sigma_{\Re\Im}) \\
\Sigma &= \begin{bmatrix} \Sigma_{\Re\Re} & \Sigma_{\Re\Im} \\ \Sigma_{\Im\Re} & \Sigma_{\Im\Im} \end{bmatrix} = \sigma^2(\tilde{\mathbf{A}}^\mathsf{T}\tilde{\mathbf{A}})^{-1}
\end{aligned} \tag{10.19}$$

For details on the $\mathcal{CN}$-distribution, see, e.g., [Picinbono, 1996]. A linear combination of squared variables distributed according to a complex normal ($\mathcal{CN}$) distribution, is distributed according to a generalized $\chi^2$ distribution, a special case of the gamma distribution. Expressions for sums of dependent gamma-distributed variables exist, see, e.g., [Paris, 2011], but no expressions for the distribution of linear combinations of norms of Gaussian vectors, e.g., (10.7), are known to the author. In order to establish estimates of confidence bounds on the spectral functions, one is therefore left with high-dimensional integration or Monte-Carlo techniques. Monte-Carlo estimates will be used in the results presented in this chapter. The sampling from a $\mathcal{CN}$-distribution is outlined in Proposition 4, with a proof given in Sec. 10.A:

PROPOSITION 4
The vector

$$z = \Re\tilde{z} + i\Im\tilde{z} \in \mathbb{C}^D$$

where

$$\begin{bmatrix} \Re\tilde{z} \\ \Im\tilde{z} \end{bmatrix} = L \begin{bmatrix} \Re z \\ \Im z \end{bmatrix}, \quad \Re z, \Im z \sim \mathcal{N}(0, I) \in \mathbb{R}^D$$

and $\Sigma = LL^\mathsf{T}$ is a Cholesky decomposition of the matrix

$$\Sigma = \frac{1}{2} \begin{bmatrix} \Re(\Gamma + C) & \Im(-\Gamma + C) \\ \Im(\Gamma + C) & \Re(\Gamma - C) \end{bmatrix} \in \mathbb{R}^{2D \times 2D}$$

is a sample from the complex normal distribution $\mathcal{CN}(0, \Gamma, C)$. □

By sampling from the posterior distribution $p(k_\omega|y)$ and propagating the samples through the nonlinear functions $A(\omega, v)$ and $\varphi(\omega, v)$, estimates of relevant confidence intervals are easily obtained.

The quality of the estimate thus hinges on the ability of the basis-function expansion to approximate the given functions $a$ and $b$ in (10.9). Park and Sandberg (1991) provide us with the required result that establishes RBF expansions as universal function approximators for well-behaved functions.

**Line-spectral estimation**

In many application, finding a sparse spectral estimate with only a few nonzero frequency components is desired. Sparsity-promoting regularization can be employed when solving for the Fourier coefficients in order to achieve this. This procedure is sometimes referred to as line-spectral estimation [Stoica and Moses, 2005] or $L_1$-regularized spectral estimation. While this technique only requires the addition of a regularization term to the cost function in (6.5) on the form $\|k\|_1$, the resulting problem no longer has a solution on closed form, necessitating an iterative solver. Along with the standard periodogram and Welch spectral estimates, we compare $L_1$-regularized spectral estimation to the proposed approach on a sparse estimation problem in the next section. We further incorporate group-lasso regularization for the proposed approach. The group-lasso, described in Sec. 6.4, amounts to adding the term

$$\sum_{\omega \in \Omega} \|k_\omega\|_2 \tag{10.20}$$

to the cost function. We solve the resulting lasso and group-lasso regularized spectral-estimation problems using the ADMM algorithm [Parikh and Boyd, 2014].

## 10.3  Experimental Results

### Simulated signals

To assess the qualities of the proposed spectral-decomposition method, a test signal $y_t$ is generated as follows

$$y_t = \sum_{\omega \in \Omega} A(\omega, v) \cos\left(\omega x - \varphi(\omega, v)\right) + e \quad e \in \mathcal{N}(0, 0.1^2)$$
$$v_t = \text{linspace}(0, 1, N)$$
$$x = \text{sort}(\mathcal{U}(0, 10))$$

where $\Omega = \{4\pi, 20\pi, 100\pi\}$, the scheduling variable $v_t$ is generated as $N = 500$ equidistantly sampled points between 0 and 1 and $x$ is a sorted vector of uniform random numbers. The sorting is carried out for visualization purposes and for the Fourier-based methods to work, but this property is not a requirement for the proposed method. The functions $A$ and $\varphi$ are defined as follows

$$A(4\pi, v) = 2v^2$$
$$A(20\pi, v) = 2/(5v + 1)$$
$$A(100\pi, v) = 3e^{-10(v-0.5)^2}$$
$$\varphi(\omega, v) = 0.5 A(\omega, v) \tag{10.21}$$

where the constants are chosen to allow for convenient visualization. The signals $y_t$ and $v_t$ are visualized as functions of the sampling points $x$ in Fig. 10.2 and the functions $A$ and $\varphi$ together with the resulting estimates and confidence intervals using $J = 50$ basis functions are shown in Fig. 10.2. The traditional power spectral density can be calculated from the estimated coefficients as

$$P(\omega) = \left| \sum_{j=1}^{J} \hat{k}_{\omega, j} \right|^2 \tag{10.22}$$

and is compared to the periodogram and Welch spectral estimates in Fig. 10.3. This figure illustrates how the periodogram and Welch methods fail to clearly identify the frequencies present in the signal due to the dependence on the scheduling variable $v$. The LPV spectral method, however, correctly identifies all three frequencies present. Incorporation of $L_1$ regularization or group-lasso regularization introduces a bias. The $L_1$-regularized periodogram is severely biased, but manages to identify the three frequency components present. The difference between the LPV method and the group-lasso LPV method is small, where the regularization correctly sets all non-present frequencies to zero, but at the expense of a small bias.

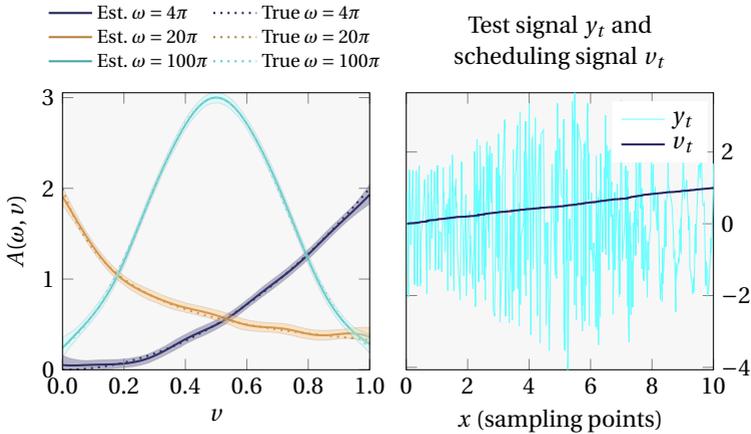

**Figure 10.2** Left: True and estimated functional dependencies with 95% confidence intervals. Right: Test signal with $N = 500$ datapoints. The signal contains three frequencies, where the amplitude and phase are modulated by the functions (10.21) depicted in the left panel. For visualization purposes, $v$ is chosen as an increasing signal. We can clearly see how the signal $y_t$ on the right has a higher amplitude when $v \approx 0.5$ and is dominated by a low frequency when $v \approx 1$, corresponding to the amplitude functions on the left.

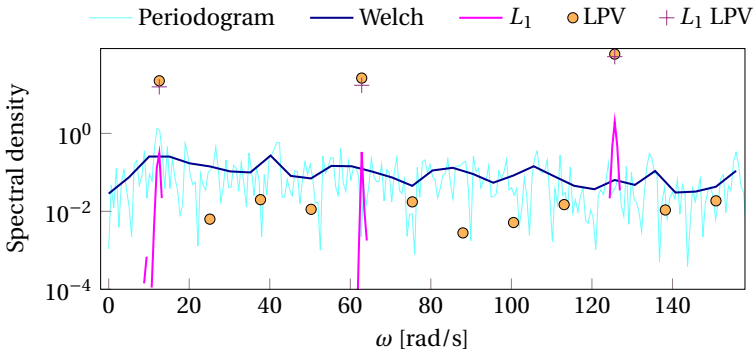

**Figure 10.3** Estimated spectra, test signal. The periodogram and Welch methods fail to identify the frequencies present in the signal due to the dependence on the scheduling variable $v$. The $L_1$-regularized periodogram correctly identifies the frequency components presents, but is severely biased. The LPV spectral method correctly identifies all three frequencies present and the group-lasso LPV method identifies the correct spectrum with a small bias and correctly sets all other frequencies to 0.

**Measured signals**

The proposed method was used to analyze measurements obtained from an ABB dual-arm robot (Fig. 9.2). Due to torque ripple and other disturbances, there is a velocity-dependent periodic signal present in the velocity control error, which will serve as the subject of analysis. The analyzed signal is shown in Fig. 10.4.

The influence of Coulomb friction on the measured signal is mitigated by limiting the support of half of the basis functions to positive velocities and vice versa. A total number of 10 basis functions was used and the model was identified with ridge regression. The regularization parameter was chosen using the L-curve method [Hansen, 1994]. The identified spectrum is depicted in Fig. 10.5, where the dominant frequencies are identified. These frequencies correspond well with a visual inspection of the data. Figure 10.5 further illustrates the result of applying the periodogram and Welch spectral estimators to data that has been sorted and interpolated to an equidistant grid. These methods correctly identify the main frequency, $4\,\mathrm{rev}^{-1}$, but fail to identify the lower-amplitude frequencies at $7\,\mathrm{rev}^{-1}$ and $9\,\mathrm{rev}^{-1}$ visible in the signal. The amplitude functions for the three strongest frequencies are illustrated in Fig. 10.6, where it is clear that the strongest frequency, $4\,\mathrm{rev}^{-1}$, has most of its power distributed over the lower-velocity datapoints, whereas the results indicate a slight contribution of frequencies at $7\,\mathrm{rev}^{-1}$ and $9\,\mathrm{rev}^{-1}$ at higher velocities, corresponding well with a visual inspection of the signal. Figure 10.6 also displays a histogram of the velocity values of the analyzed data. The confidence intervals are narrow for velocities present in the data, while they become wider outside the represented velocities.

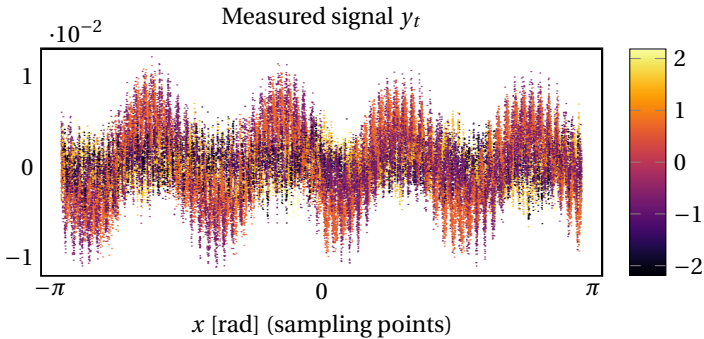

**Figure 10.4**   Measured signal as a function of sampling location, i.e., motor position. The color information indicates the value of the velocity/scheduling variable in each datapoint. Please note that this is not a plot of the measured data sequentially in time. This figure indicates that there is a high amplitude periodicity of $4\,\mathrm{rev}^{-1}$ for low velocities, and slightly higher frequencies but lower-amplitude signals at $7\,\mathrm{rev}^{-1}$ and $9\,\mathrm{rev}^{-1}$ for higher velocities.

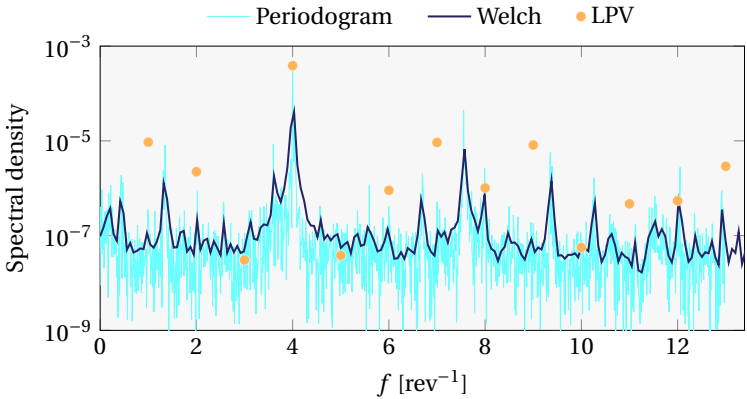

**Figure 10.5**  Estimated spectra, measured signal. The dominant frequencies are identified by the proposed method, while the Fourier-based methods correctly identify the main frequency, $4\,\mathrm{rev}^{-1}$, but fail to identify the lower-amplitude frequencies at $7\,\mathrm{rev}^{-1}$ and $9\,\mathrm{rev}^{-1}$ visible in the signal in Fig. 10.4.

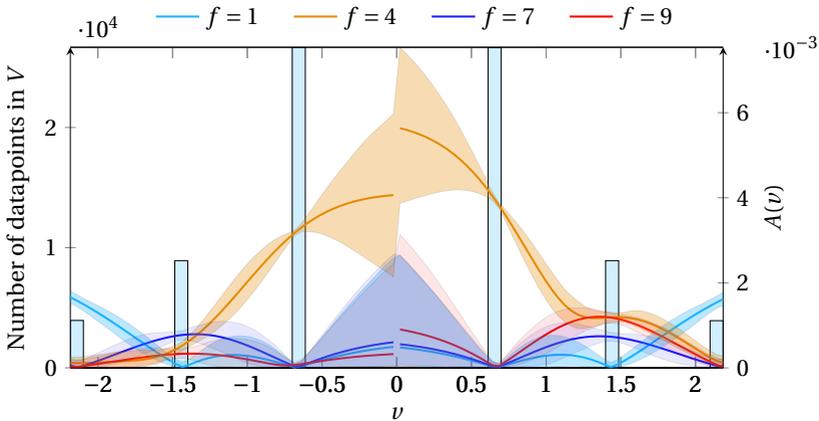

**Figure 10.6**  Estimated functional dependencies with 99% confidence intervals. The left axis and histogram illustrates the number of datapoints available at each velocity $v$. The right axis illustrate the estimated amplitude functions together with their confidence intervals.

## 10.4  Discussion

In this chapter, we make further use of basis-function expansions, this time in the context of spectral estimation. The common denominator is the desire to model a functional relationship where the function is low-dimensional and can have an arbitrary complicated form. The goal is to estimate how the amplitude and phase of sinusoidal basis functions that make up a signal vary with an auxiliary signal. Due to the phase variable entering nonlinearly, the estimation problem is rephrased as the estimation of linear parameter-varying coefficients of sines and cosines of varying frequency. The amplitude and phase functions are then calculated using nonlinear transforms of the estimated coefficients. While it was shown that the simple estimators of the amplitude and phase functions are biased, this bias vanishes as the number of datapoints increases.

From the expression for the expected value of the amplitude function

$$A < \mathbb{E}\left\{\hat{A}\right\} < \sqrt{A^2 + \phi^\mathsf{T}\Sigma_\alpha\phi + \phi^\mathsf{T}\Sigma_\beta\phi} \tag{10.23}$$

we see that the bias vanishes as $\Sigma_\alpha$ and $\Sigma_\beta$ are reduced. Further insight into this inequality can be gained by considering the scalar, nonlinear transform

$$f(x) = |x|, \quad x \sim \mathcal{N}(\mu, \sigma^2) \tag{10.24}$$

If $\mu$ is several standard deviations away from zero, the nonlinear aspect of the absolute-value function will have negligible effect. When $\mu/\sigma$ becomes smaller, say less than 2, the effect starts becoming significant. Hence, if the estimated coefficients are significantly different from zero, the bias is small. This is apparent also from the figures indicating the estimated functional relationship with estimated confidence bounds. For areas where data is sparse, the confidence bounds become wider and the estimate of the mean seems to be inflated in these areas.

The leakage present in the standard Fourier-based methods is usually undesired. The absence of leakage might, however, be problematic when the number of estimated frequencies is low, and the analyzed signal contains a very well defined frequency. If this frequency is not included in the set of basis frequencies, the absence of leakage might lead to this component being left unnoticed. Introduction of leakage is technically straightforward, but best practices for doing so remain to be investigated.

## 10.5  Conclusions

This chapter developed a spectral-estimation method that can decompose the spectrum of a signal along an external dimension, which allows estimation of the amplitude and phase of the frequency components as functions of the external variable. The method is linear in the parameters which allows for straightforward calculation of the spectrum through solving a set of linear equations. The method does not impose limitations such as equidistant sampling, does not suffer from leakage and allows for estimation of arbitrary chosen frequencies. The closed-form calculation of the spectrum requires $\mathcal{O}(J^3O^3)$ operations due to the matrix inversion associated with solving the LS-problem, which serves as the main drawback of the method if the number of frequencies to estimate is large (the product $JO$ greater than a few thousands). For larger problems, an iterative solution method must be employed. Iterative solution further enables the addition of a group-lasso regularization, which was shown to introduce sparsity in the estimated spectrum.

Implementations of all methods and examples discussed in this chapter are made available in [*LPVSpectral.jl*, B.C., 2016].

## Appendix A. Proofs

***Proof.*** Lemma 3
The amplitude $A$ is given by two trigonometric identities

$$A\cos(x - \varphi) = A\cos(\varphi)\cos(x) + A\sin(\varphi)\sin(x) \qquad (10.25)$$

$$= k_1\cos(x) + k_2\sin(x) \qquad (10.26)$$

$$k_1^2 + k_2^2 = A^2(\cos(\varphi)^2 + \sin(\varphi)^2) = A^2 \qquad (10.27)$$

and the phase $\varphi$ by

$$\arctan\left(\frac{k_2}{k_1}\right) = \arctan\left(\frac{A\sin\varphi}{A\cos\varphi}\right) = \varphi$$

where $k_1 = A\cos(\varphi), \quad k_2 = A\sin(\varphi)$ is identified from (10.25). $\qquad \square$

***Proof.*** Proposition 4
Let $v^\mathsf{T} = \begin{bmatrix} x^\mathsf{T} & y^\mathsf{T} \end{bmatrix}$. The mean and variance of $\tilde{v} = Lv$ is given by

$$\mathbb{E}\{\tilde{v}\} = L\mathbb{E}\{v\} = 0$$
$$\mathbb{E}\{\tilde{v}\tilde{v}^\mathsf{T}\} = \mathbb{E}\{Lvv^\mathsf{T}L^\mathsf{T}\} = LIL^\mathsf{T} = \Sigma$$

The complex vector $z = x + iy \in \mathbb{C}^D$ composed of the elements of $v$ is then $\mathcal{CN}(0, \Gamma, C)$-distributed according to [Picinbono, 1996, Proposition 1]. $\qquad \square$

# 11

# Model-Based Reinforcement Learning

In this chapter, we will briefly demonstrate how the models developed in Part I of the thesis can be utilized for reinforcement-learning purposes. As alluded to in Sec. 5.2, there are many approaches to RL, some of which learn a value function, some a policy directly and some that learn the dynamics of the system. While value-function-based methods are very general and can be applied to a very wide range of problems without making many assumptions, they are terribly inefficient and require a vast amount of interaction with the environment in order to learn even simple policies. Model-based methods trade off some of the generality by making more assumptions, but in return offer greater data efficiency. In line with the general topic of this thesis, we will demonstrate the use of the LTV models and identification algorithms from Chap. 7 together with the black-box models from Chap. 8 in an example of model-based reinforcement learning. The learning algorithm used is derived from [Levine and Koltun, 2013], where it was shown to have impressive data-efficiency when applied to physical systems. In this chapter, we show how the data-efficiency can be improved further by estimating the dynamical models using the methods from Chap. 7.

## 11.1  Iterative LQR—Differential Dynamic Programming

Trajectory optimization refers simply to an optimization problem with sequential structure and dynamical equations among the constraints. As an example, consider the problem of finding a sequence of joint torques that moves a robot arm from one pose to another while minimizing a weighted sum of the required time and energy. In solving this problem, the dynamics of the robot have to be satisfied by any feasible trajectory, and it is thus reasonable to include the dynamical equations as constraints in the optimization problem.

In Sec. 5.1 we briefly mentioned that the LQR algorithm [Todorov and Li, 2005] can be used for trajectory optimization with nonlinear systems and nonquadratic cost functions by linearizing the system and approximating the cost function with a quadratic. The solution will, due to the approximations, not be exact, and the procedure must thus be iterated, where new approximations are obtained around the previous solution trajectory. This algorithm can be used for model-based, trajectory centric reinforcement learning by, after each optimization pass, performing a rollout on the real system and updating the system model using the new data. We will begin with an outline of the algorithm [Todorov and Li, 2005] and then provide a detailed description on how we use it for model-based reinforcement learning.

**The algorithm**

The standard LQR algorithm uses dynamic programming to find the optimal linear, time-varying feedback gain along a trajectory of an, in general, linear time-varying system. To extend this to the nonlinear, non-Gaussian setting, we linearize the system and cost function $c$ along a trajectory $\tau_i = \{\hat{x}_t, \hat{u}_t\}_{t=1}^{T}$, where $i$ denotes the learning iteration starting at 1 and $s$ denotes the superstate $[x^\top\, u^\top]^\top$:

$$x^+ = Ax + Bu \tag{11.1}$$

$$c(x, u) = s^\top c_s + \frac{1}{2} s^\top c_{ss} s \tag{11.2}$$

where subscripts denote partial derivatives. Using the linear quadratic model (11.1), the optimal state- and state-action value functions $V$ and $Q$ can be calculated recursively starting at $t = T$ using the expressions

$$Q_{ss} = c_s + \hat{f}_s^\top V_{xx}^+ \hat{f}_s \tag{11.3}$$

$$Q_s = c_s + \hat{f}_s^\top V_x^+ \tag{11.4}$$

$$V_{xx} = Q_{xx} - Q_{ux}^\top Q_{uu}^{-1} Q_{ux} \tag{11.5}$$

$$V_x = Q_x - Q_{ux}^\top Q_{uu}^{-1} Q_u \tag{11.6}$$

where all quantities are given at time $t$ and $\cdot^+$ denotes a quantity at $t+1$. Subscripts $Q_s$ and $Q_{ux}$, etc., denote partial derivatives $\nabla_s Q$ and $\nabla_{ux} Q$, respectively. The calculation of value functions is started at time $t = T$ with $V_{T+1} = 0$. Given the calculated value functions, the optimal update to the control signal trajectory, $k$, and feedback gain, $K$, are given by

$$K = -Q_{uu}^{-1} Q_{ux} \tag{11.7}$$

$$k = -Q_{uu}^{-1} Q_u \tag{11.8}$$

These are used to update the nominal trajectory $\tau$ as

$$u_{i+1} = u_i + k_i \tag{11.9}$$

$$C(x) = K\bar{x} \tag{11.10}$$

where $\bar{x} = x - \hat{x}$ denotes deviations from the nominal trajectory. The update of the state trajectory in $\tau$ is done by forward simulation of the system using $u_{i+1}$ as input. In the reinforcement-learning setting, this corresponds to performing an experiment on the real system.

**Stochastic controller**

For learning purposes, it might be beneficial to let the controller be stochastic in order to more efficiently explore the statespace around the current trajectory. A Gaussian controller

$$p(u|x) = \mathcal{N}(K\bar{x}, \Sigma) \tag{11.11}$$

can be calculated with the above framework, where the choice $\Sigma = Q_{uu}^{-1}$ was shown by Levine and Koltun (2013) to optimize a maximum entropy cost function

$$p(\tau) = \underset{p(u|x)}{\arg\min} \, \mathbb{E}\{c(\tau)\} - \mathcal{H}(p(\tau)) \tag{11.12}$$

$$\text{subject to} \quad p(x^+|x, u) = \mathcal{N}(Ax + Bu, \Sigma_f) \tag{11.13}$$

To explore with Gaussian noise with a covariance matrix of $Q_{uu}^{-1}$ makes intuitive sense. It allows us to add large noise in the directions where the value function does not increase much. Directions where the value function seems to increase sharply, however, should not be explored as much as this would lead the optimization to high-cost areas of the state-space.

The term $\Sigma_f$ in (11.13) corresponds to the uncertainty in the model of $f$ along the trajectory. If $f$ is modeled as an LTV model and estimated using the methods in Chap. 7, this term can be derived from quantities calculated by the Kalman smoothing algorithm, making these identification methods a natural fit to the current setting.

**Staying close to the last trajectory**

In a reinforcement-learning setting, the model under which the algorithm optimizes, $\hat{f}$, is estimated from a small amount of data. Optimization can in general lead arbitrarily far away from the starting point, i.e., the initial trajectory. A nonlinear model or LTV model estimated around a trajectory is not likely to accurately represent the dynamics of the system in regions of the state space far away from where it was estimated. To mitigate this issue and make sure that the optimization procedure does not suggest trajectories too far away from the trajectory along which the model was estimated, a constraint can be added to the optimization problem. The constraint suggested by Levine and Abbeel (2014) takes the form

$$D_{\text{KL}}\big(p(\tau) \,||\, \hat{p}(\tau)\big) \leq \epsilon \tag{11.14}$$

where

$$p(\tau) = \left( p(x_0) \prod_t p(x_{t+1}|x_t, u_t) p(u_t|x_t) \right) \tag{11.15}$$

and $\hat{p}(\tau)$ denotes the previous trajectory distribution. The addition of (11.14) allows the resulting constrained optimization problem to be solved efficiently using a small modification to the algorithm, reminiscent of the Lagrangian method. The simplicity of the resulting problem is a product of the linear-Gaussian nature of the models involved.

## 11.2 Example—Reinforcement Learning

In this example, we use methods from Chap. 7 to identify LTV dynamics models for reinforcement learning. We will consider the system used for simulations in Chap. 8, the pendulum on a cart. Owing to the nonlinear nature of the pendulum dynamics, linear approximations of the dynamics in the upward (initial) position have an unstable pole, and imaginary poles in the downward (final) position. This simple system is well understood and often used as a reinforcement learning benchmark. The traditional goal of the benchmark is to find a policy that is able to swing up the pendulum from its downward hanging position to the unstable equilibrium where the pendulum is upright. We will consider the reverse task of optimally dampen the swinging pendulum after it has fallen from its upright position, since this task allows us to use a trajectory-based RL algorithm. We can thus compare the results to those obtained by using an optimal control algorithm equipped with the ground-truth system model. This task is also robust to the choice of cost function. We simply penalize states and control signals quadratically and constrain the control signal to $\pm 10$.

To find an optimal dampening policy, we employ a reinforcement learning framework inspired by Levine and Koltun (2013), summarized in Algorithm 7. In

---

**Algorithm 7** A simple reinforcement learning algorithm using trajectory optimization.

---

   **repeat**
       Perform rollout
       Fit dynamics model
       Optimize trajectory and controller
   **until** Convergence

---

the third step of Algorithm 7, we employ the iLQR algorithm with a bound on the KL divergence between two consecutive trajectory distributions. To incorporate control signal bounds, we employ a variant of iLQR due to Tassa et al. (2014). Our implementation of iLQR, allowing control signal contraints and constraints on the KL divergence between trajectory distributions, is made available at [*Differential-DynamicProgramming.jl*, B.C., 2016].

We compare three different models; the ground-truth system model, an LTV model obtained by solving (7.5) using the Kalman smoothing algorithm, and an LTI model. The total cost over $T = 400$ time steps is shown as a function of learning iteration in Fig. 11.1. The figure illustrates how the learning procedure reaches the optimal cost of the ground-truth model when an LTV model is used, whereas when using an LTI model, the learning diverges. The figure further illustrates that if the LTV model is fit using a prior (Sec. 7.5), the learning speed is increased. We obtain the prior in two different ways. In one experiment, we fit a neural network model on the form $x^+ = g(x, u) + x$ with the same structure and parameters as in Sec. 8.6, with tangent-space regularization. After every rollout, the black-box model is refit using all available data up to that point.[1] No uncertainty estimate is available from such a model, and we thus heuristically decay the covariance of the prior obtained from this model as the inverse of the learning iteration. This prior information is easily incorporated into the learning of the LTV model using the method outlined in Sec. 7.5, and the precision of the prior acts as a hyper parameter weighing the LTV model against the neural network model. In the second experiment with priors, we pretrain a neural network model of the system in a standard system-identification fashion, and use this model as a prior without updating it with new data during the learning iterations. The figure indicates that use of a successively updated prior model is beneficial to convergence, but it takes a few trajectories of data before the difference is noticeble. If the prior model is pretrained the benefit is immediate, but failure to update the prior model hampers convergence towards the end of learning. This behavior indicates that it was beneficial to update the model with data following the state-visitation distribution of the policy under optimization, as this focuses the attention of the neural network, allowing it to accurately represent the system in relevant areas of the state space. In taking this experiment further, a Bayesian prior model, allowing calculation of the true posterior over the Jacobian for use in (7.16), would be beneficial. Example trajectories found by the learning procedure using different models are illustrated in Fig. 11.2. Due to the constraints on the control signal, all algorithms converge to similar policies of bang-bang type.

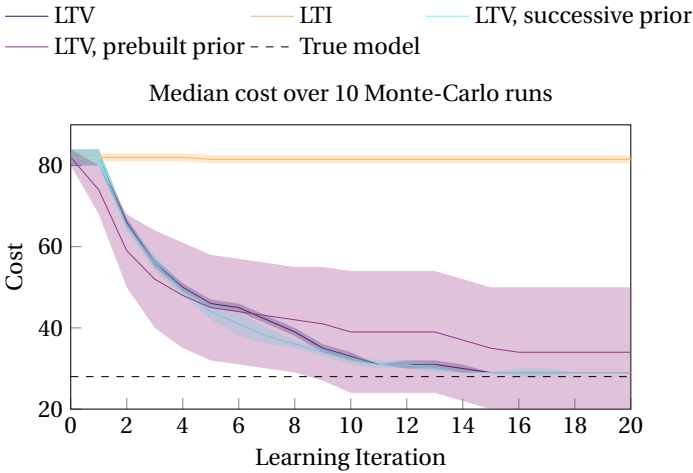

**Figure 11.1**   Reinforcement learning example. Three different model types are used to iteratively optimize the trajectory of a pendulum on a cart. Colored bands show standard deviation around the median. Linear expansions of the dynamics are unstable in the upward position and stable in the downward position. The algorithm thus fails with an LTI model. A standard LTV model learns the task well and reaches the optimal cost in approximately 15 learning iterations. Building a successive DNN model for use as prior to the LTV model improves convergence slightly. Using a DNN model from a priori identification as prior improves convergence at the onset of training when the amount of information obtained about the system is low. Interestingly, the figure highlights how use of this model hampers performance near convergence. This illustrates that successively updating the DNN model with new data along the current trajectory improves the model in the areas of the state space important to the task. The optimal control algorithm queried the true system model a total of 52 times.

---

[1] Retraining in every iteration is computationally costly, but ensures that the model does not converge to a minimum only suitable for the first seen trajectories.

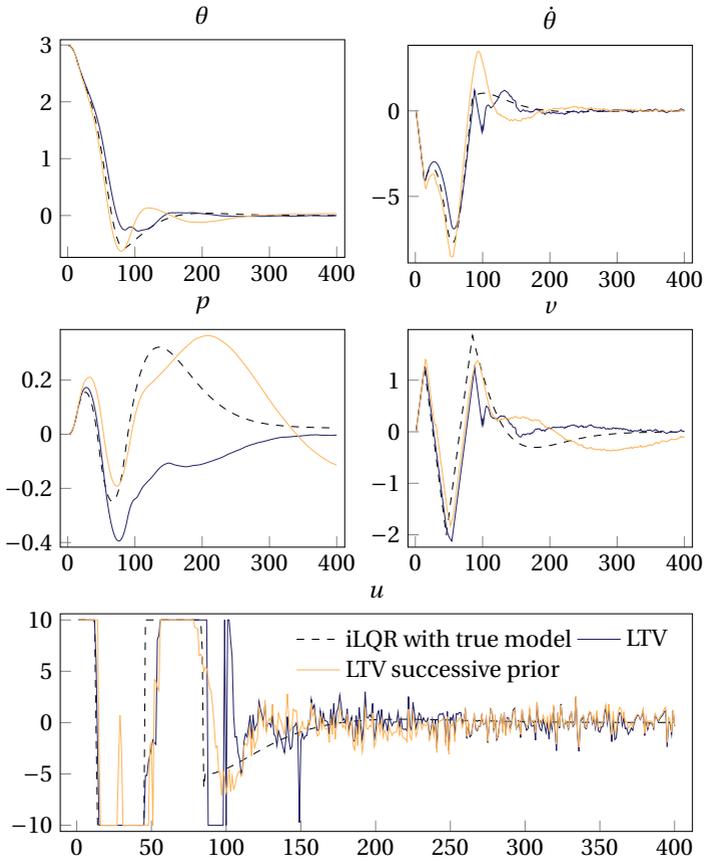

**Figure 11.2** Optimized trajectories of pendulum system after learning using three different methods. The state consists of angle $\theta$, angular velocity $\dot{\theta}$, position $p$ and velocity $v$. DDP denotes the optimal control procedure with access to the true system model. The noise in the control signal trajectories of the RL algorithms is added for exploration purposes.

Although a simple example, Fig. 11.1 illustrates how an appropriate choice of model class can allow a trajectory-based reinforcement learning problem to be solved using very few experiments on the real process. We emphasize that one iteration of Algorithm 7 corresponds to one experiment on the process. Levine and Koltun (2013), from where inspiration to this approach was drawn, require a handful of experiments on the process per learning iteration and policy update, just to fit an adequate dynamics model.

**Part II**

# Robot State Estimation

# 12

# Introduction—Friction Stir Welding

Friction stir welding (FSW) is becoming an increasingly popular joining technique capable of producing stronger joints than fusion welding, allowing for a reduction of material thickness and weight of the welded components [Midling et al., 1998; De Backer, 2014]. Conventional, custom-made FSW machines of gantry type are built to support the large forces inherent in the FSW process. The high stiffness required has resulted in expensive and inflexible machinery, which has limited the number of feasible applications of FSW as well as the adaptation of FSW as a joining technique [De Backer, 2014]. Recently, the use of robotic manipulators in FSW applications has gained significant interest due to the lower cost compared to conventional FSW machinery as well as the much increased flexibility of an articulated manipulator [De Backer, 2014; Guillo and Dubourg, 2016]. The downsides of the use of robots include the comparatively low stiffness, which causes significant deflections during welding, with a lower quality weld as result.

A typical approach adopted to reduce the uncertainty introduced by deflections is stiffness/compliance modeling [Guillo and Dubourg, 2016; Lehmann et al., 2013; De Backer and Bolmsjö, 2014]. This amounts to finding models of the joint deflections $\Delta q$ or of the Cartesian deflections $\Delta x$ on one of the forms

$$\Delta q = C_j(\tau) \tag{12.1}$$

$$\Delta X = C_C(\mathfrak{f}) \tag{12.2}$$

where $\tau$ and $\mathfrak{f}$ are the joint torques and external forces, respectively, $X$ is some notion of Cartesian pose and $C$ denotes some, possibly nonlinear, compliance function. The corresponding inverse relations are typically referred to as stiffness models. Robotic compliance modeling has been investigated by many authors, where the most straightforward approach is based on linear models obtained by measuring the deflections under application of known external loads. To avoid the dependence on expensive equipment capable of accurately measuring the deflections, techniques such as the clamping method have been proposed [Bennett et al., 1992; Lehmann et al., 2013; Sörnmo, 2015; Olofsson, 2015] for the identification of models on the form (12.1). This approach makes the assumption that deflections only occur in the joints, in the direction of movement. Hence, deflections occurring in the links or in the joints orthogonally to the movement cause model errors, limiting the resulting accuracy of the model obtained [Sörnmo, 2015]. In [Guillo and Dubourg, 2016], the use of arm-side encoders was investigated to allow for direct measurement of the joint deflections. As of today, arm-side encoders are not available in the vast majority of robots, and the modification required to install them is yet another obstacle to the adaptation of robotic FSW. The method further suffers from the lack of modeling of link- and orthogonal joint deflections.

Cartesian models like (12.2) have been investigated in the FSW context by [De Backer, 2014; Guillo and Dubourg, 2016; Abele et al., 2008]. The proposed Cartesian deflection models are local in nature and not valid globally. This requires separate models to be estimated throughout the workspace, which is time consuming and limits the flexibility of the setup.

Although the use of compliance models leads to a reduction of the uncertainty introduced by external forces, it is difficult to obtain compliance models accurate enough throughout the entire workspace. This fact serves as the motivation for complementing the compliance modeling with sensor-based feedback. Sensor-based feedback is standard in conventional robotic arc and spot welding, where the crucial task of the feedback action is to align the weld torch with the seam along the transversal direction, with the major uncertainty being the placement of the work pieces. During FSW, however, the uncertainties in the robot pose are significant, while the tilt angle of the tool in addition to its position is of great importance [De Backer et al., 2012]. This requires a state estimator capable of estimating accurately at least four DOF, with slightly lower performance required in the tool rotation axis and the translation along the weld seam. Conventional seam-tracking sensors are capable of measuring 1-3 DOF only [Nayak and Ray, 2013; Gao et al., 2012], limiting the information available to a state estimator and thus maintaining the need for, e.g., compliance modeling.

Motivated by the concerns raised above, we embark on developing calibration methods, a state estimator and a framework for simulation of robotic seam tracking under the influence of large external process forces. Chapter 13 develops methods for calibration of 6 DOF force/torque sensors and a seam-tracking laser sensor, while a particle-filter based state-estimator and simulation framework is developed in Chap. 14. Notation, coordinate frames and variables used in this part of the thesis are listed in Table 12.1.

## 12.1 Kinematics

Kinematics refer to the motion of objects without concerns for any forces involved (as opposed to dynamics). This section will briefly introduce some notation and concepts within kinematics that will become important in subsequent chapters, and simultaneously introduce the necessary notation. Most of the discussion will focus around the representation and manipulation of coordinate frames.

**Rotation and transformation matrices**

*SO*(3) denotes the special orthogonal group 3 and is the set of all 3D orientation matrices. Matrices in *SO*(3) have determinant one and orthogonal columns and rows with norm one, $R^\mathsf{T} R = R R^\mathsf{T} = I$ [Spong et al., 2006].

*SE*(3) denotes the special Euclidean group 3 and is the set of all 3D rigid body transformation matrices [Spong et al., 2006]. Transformation matrices in *SE*(3) have the form

$$T_C^D \in SE(3) = \begin{bmatrix} R_C^D & t_C^D \\ \mathbf{0}_{1\times 3} & 1 \end{bmatrix} \tag{12.3}$$

where the rotation matrix $R \in SO(3)$ and the translation vector $t \in \mathbb{R}^3$. $T_C^D$ denotes the rigid-body transformation from frame *D* to frame *C*, such that $\mathbf{p}_C = T_C^D \mathbf{p}_D$ for a point $\mathbf{p}_D = \begin{bmatrix} p_D^\mathsf{T} & 1 \end{bmatrix}^\mathsf{T} = \begin{bmatrix} x & y & z & 1 \end{bmatrix}_D^\mathsf{T}$ and $\mathbf{v}_C = T_C^D \mathbf{v}_D$ for a vector on the form $\mathbf{v}_D = \begin{bmatrix} x & y & z & 0 \end{bmatrix}_D^\mathsf{T}$. The inverse of a transformation matrix is given by

$$T = \begin{bmatrix} R & t \\ \mathbf{0} & 1 \end{bmatrix} \in SE(3), \quad T^{-1} = \begin{bmatrix} R^\mathsf{T} & -R^\mathsf{T} t \\ \mathbf{0} & 1 \end{bmatrix} \tag{12.4}$$

We say that **p** is the *homogeneous* form of *p*, which allows $Rp + t$ to be represented compactly as $T\mathbf{p}$.

**Table 12.1**   Definition and description of coordinate frames and variables.

| | | |
|---|---|---|
| $\mathcal{RB}$ | | Robot base frame. |
| $\mathcal{T}$ | | Tool frame, attached to the (TCP). |
| $\mathcal{S}$ | | Sensor frame, specified according to Fig. 14.1. |
| $q$ | $\in \mathbb{R}^n$ | Joint Coordinate |
| $X$ | $\in SE(3)$ | Tool pose (State) |
| $\tau$ | $\in \mathbb{R}^n / \mathbb{R}^3$ | Joint torque or external torque |
| $\mathfrak{f}$ | $\in \mathbb{R}^6$ | External force/torque wrench |
| $m$ | $\in \mathbb{R}^2$ | Laser measurement in $\mathcal{S}$ |
| $m_a$ | $\in \mathbb{R}^1$ | Laser angle measurement in $\mathcal{S}$ |
| $e$ | $\in \mathbb{R}^2$ | Measurement error |
| $T_A^B$ | $\in SE(3)$ | Transformation matrix from $\mathcal{B}$ to $\mathcal{A}$ |
| $F_k(q)$ | $\in SE(3)$ | Robot forward kinematics at pos. $q$ |
| $J(q)$ | $\in \mathbb{R}^{6\times n}$ | Manipulator Jacobian at pos. $q$ |
| $\langle s \rangle$ | $\in so(3)$ | Skew-symmetric matrix with parameters $s \in \mathbb{R}^3$ |
| $\hat{a}$ | | Estimate of variable $a$ |
| $a^+$ | | $a$ at the next sample instant |
| $\bar{a}$ | | Reference for variable $a$ |
| $a_{i:j}$ | | Elements $i, i+1, ..., j$ of $a$ |
| $T^\vee$ | $\in \mathbb{R}^6$ | The twist coordinate representation of $T$ |

For additional details on kinematics and transforms in the robotics context, the reader is referred to, e.g., [Murray et al., 1994; Spong et al., 2006].

# 13

# Calibration

The field of robotics offers a wide range of calibration problems, the solutions to which are oftentimes crucial for the accuracy or success of a robotic application. For our purposes, we will loosely define *calibration* as the act of finding a transformation of a measurement from the space it is measured in, to another space in which it is more useful to us. For example, a sensor measuring a quantity in its intrinsic coordinate system must be calibrated for us to know how to interpret the measurements in an external coordinate system, such as that of the robot or its tool. This chapter will describe a number of calibration algorithms developed in order to make use of various sensors in the seam-tracking application, but their usefulness outside this application will also be highlighted.

The calibration problems considered in this chapter are all geometric in nature and related to the kinematics of a mechanical system. In kinematic calibration, we estimate the kinematic parameters of a structure. Oftentimes, and in all cases considered here, we have a well defined kinematic equation defining the kinematic structure, but this equation contains unknown parameters. To calibrate those parameters, one has to devise a method to collect data that allow us to solve for the parameters of interest. Apart from solving the equations given the collected data, the data collection itself is what sets different calibration methods apart from each other, and from system identification in general. Many methods require special-purpose equipment, either for precise perturbations of the system, or for precise measurements of the system. Use of special-purpose equipment limits the availability of a calibration method. We therefore propose calibration algorithms that solve for the desired kinematic parameters without any special-purpose equipment, making them widely applicable.

## 13.1 Force/Torque Sensor Calibration

A 6 DOF force/torque sensor is a device capable of measuring the complete wrench of forces and torques applied to the sensor. They are commonly mounted on the tool flange of a manipulator to endow it with force/torque sensing capabilities, useful for, e.g., accurate control in contact situations.

In order to make use of a force/torque sensor, the rotation matrix $R_{TF}^{S}$ between the tool flange and the sensor coordinate systems, the mass $m$ held by the force sensor at rest, and the translational vector $r \in \mathbb{R}^3$ from the sensor origin to the center of mass are required. Methods from the literature typically involve fixing the force/torque sensor in a jig and applying known forces/torques to the sensor [Song et al., 2007; Chen et al., 2015]. In the following, we will develop and analyze a calibration method that only requires movement of the sensor attached to the tool flange in free air, making it very simple to use.

**Method**

The relevant force and torque equations are given by

$$f_S = R_S^{TF} R_{TF}^{RB}(mg) \tag{13.1}$$

$$\tau_S = R_S^{TF} \langle r \rangle R_{TF}^{RB}(mg) \tag{13.2}$$

where $g$ is the gravity vector given in the robot base-frame and $f, \tau$ are the force and torque measurements, respectively, such that $\mathfrak{f} = [f^\mathsf{T} \ \tau^\mathsf{T}]^\mathsf{T}$. At first glance, this is a hard problem to solve. The equation for the force relation does not appear to allow us to solve for both $m$ and $R_S^{TF}$, the constraint $R \in SO(3)$ is difficult to handle, and the equation for the torque contains the nonlinear term $R_S^{TF} \langle r \rangle$. Fortunately, however, the problem can be separated into two steps, and the constraint $R_S^{TF} \in SO(3)$ will allow us to distinguish $R_S^{TF}$ from $m$.

A naive approach to the stated calibration problem is to formulate an optimization problem where $R$ is parameterized using, e.g., Euler angles. A benefit of this approach is its automatic and implicit handling of the constraint $R \in SO(3)$. One then minimizes the residuals of (13.1) and (13.2) with respect to all parameters using a local optimization method. This approach is, however, prone to convergence to local optima and is hard to conduct in practice.

Instead, we start by noting that multiplying a matrix with a scalar only affects its singular values, but not its singular vectors, $mR = U(mS)V^\mathsf{T}$. Thus, if we solve a convex relaxation to the problem and estimate the product $mR$, we can recover $R$ by projecting $mR$ onto $SO(3)$ using the procedure in Sec. 6.1. Given $R$ we can easily recover $m$. Equation (13.1) is linear in $mR$ and the minimization step can readily be conducted using the least-squares approach of Sec. 6.2. To facilitate this estimation, we write (13.1) on the equivalent form

$$(f_S^\mathsf{T} \otimes I_3) \operatorname{vec}(mR) = R_{TF}^{RB} g \tag{13.3}$$

where $\operatorname{vec}(mR) \in \mathbb{R}^9$ is a vector of parameter to be estimated.

Once $R_S^{TF}$ and $m$ are estimated using measured forces only, we can estimate $r$ using the torque relation by noting that

$$\tau_S = R_S^{TF} \langle r \rangle R_{TF}^{RB}(mg) \tag{13.4}$$

$$R_{TF}^{S} \tau_S = \langle R_{TF}^{RB}(mg) \rangle r \tag{13.5}$$

where the second equation is linear in the unknown parameter-vector $r$.

When solving a relaxed problem, there is in general no guarantee that a good solution to the original problem will be found. To verify that relaxing the problem does not introduce any numerical issues or problems in the presence of noise, etc., we introduce a second algorithm for finding $R_S^{TF}$. If the mass $m$ is known in advance, the problem can be reformulated using the Cayley transform [Tsiotras et al., 1997] and a technique similar to the attitude estimation algorithm found in [Mortari et al., 2007] can be used to solve for the rotation matrix, without constraints or relaxations.

The Cayley transform of a matrix $R \in SO(3)$ is given by

$$R = (I + \Sigma)^{-1}(I - \Sigma) = (I - \Sigma)(I + \Sigma)^{-1} \tag{13.6}$$

where $\Sigma = \langle s \rangle$ is a skew-symmetric matrix of the Cayley-Gibbs-Rodrigues parameters $s \in \mathbb{R}^3$ [Tsiotras et al., 1997]. Applying the Cayley transform to the force relation yields

$$f_S = R_S^{TF} R_{TF}^{RB}(mg) \tag{13.7}$$

$$f_S = (I + \Sigma)^{-1}(I - \Sigma) R_{TF}^{RB}(mg) \tag{13.8}$$

$$(I + \Sigma) f_S = (I - \Sigma) R_{TF}^{RB}(mg) \tag{13.9}$$

$$\Sigma \big( f_S + R_{TF}^{RB}(mg) \big) = -\big( f_S - R_{TF}^{RB}(mg) \big) \tag{13.10}$$

$$\Big[ \Sigma = \langle s \rangle, \quad \langle a \rangle b = \langle b \rangle a \Big]$$

$$\langle f_S + R_{TF}^{RB}(mg) \rangle s = -\big( f_S - R_{TF}^{RB}(mg) \big) \tag{13.11}$$

which is a linear equation in the parameters $s$ that can be solved using the standard least-squares procedure. The least-squares solution to this problem was, however, found during experiments to be very sensitive to measurement noise in $f_S$. This is due to the fact that $f_S$ appears not only in the dependent variable on the right-hand side, but also in the regressor $\langle f_S + R_{TF}^{RB}(mg) \rangle$. This is thus an *errors-in-variables* problem for which the solution is given by the *total least-squares* procedure [Golub and Van Loan, 2012], which we make use of in the following evaluation.

### Numerical evaluation

The two algorithms, the relaxation-based and the Cayley-transform based, were compared on the problem of finding $R_S^{TF}$ by simulating a force-calibration scenario where a random $R_S^{TF}$ and 100 random poses $R_{RB}^{TF}$ were generated. In one simulation, we let the first algorithm find $R_S^{TF}$ and $m$ with an error in the initial estimate of $m$ by a factor of 2, while the Cayley algorithm was given the correct mass. The results, depicted in the left panel of Fig. 13.1 indicate that the two methods performed on par with each other. The figure shows the error in the estimated rotation matrix as a function of the added measurement noise in $f_S$. In a second experiment, depicted in the right panel of Fig. 13.1, we started both algorithms with a mass estimate with 10 % error. Consequently, the Cayley algorithm performed significantly worse

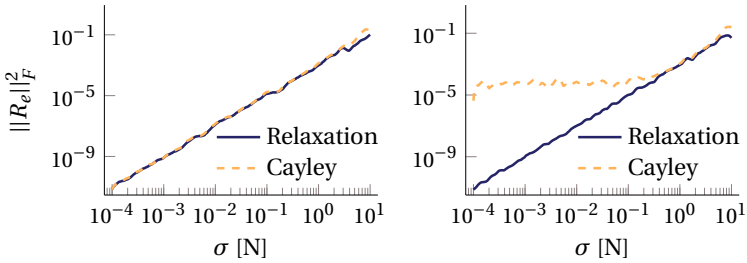

**Figure 13.1** The error in the estimated rotation matrix is shown as a function of the added measurement noise for two force-calibration methods, relaxation based and Cayley-transform based. On the left, the relaxation-based method was started with an initial mass estimate $m_0 = 2m$ whereas the Cayley-transform based method was given the correct mass. On the right, both algorithms were given $m_0 = 1.1m$

for low noise levels, while the difference was negligible when the measurement noise was large enough to dominate the final result.

The experiment showed that not only does the relaxation-based method perform on par with the unconstrained Cayley-transform based method, it also allows us to estimate the mass, reducing the room for potential errors due to an error in the mass estimate. It is thus safe to conclude that the relaxation-based algorithm is superior to the Cayley algorithm in all situations. Implementations of both algorithms are provided in [*Robotlib.jl*, B.C., 2015].

## 13.2 Laser-Scanner Calibration

Laser scanners have been widely used for many years in the field of robotics. A large group of laser scanners, such as 2D laser range finders and laser stripe profilers, provide accurate distance measurements confined to a plane. By moving either the scanner or the scanned object, a 2D laser scanner can be used to build a 3D representation of an object or the environment. To this purpose, laser scanners are commonly mounted on mobile platforms, aerial drones or robots.

This section considers the calibration of such a sensor, and as a motivating example, we consider a wrist-mounted laser scanner for robotic 3D scanning and weld-seam-tracking applications. The method does, however, work equally well in any situation where a similar sensor is mounted on a drone or mobile platform, as long as the location of the platform is known in relation to a fixed coordinate system. We will use the term robot to refer to any such system, and the robot coordinate system to refer to either the robot base coordinate system, or the coordinate system of a tracking device measuring the location of the tool flange. To relate the measurements of the scanner to the robot coordinate system, the rigid transformation between the scanner coordinate system and the tool flange

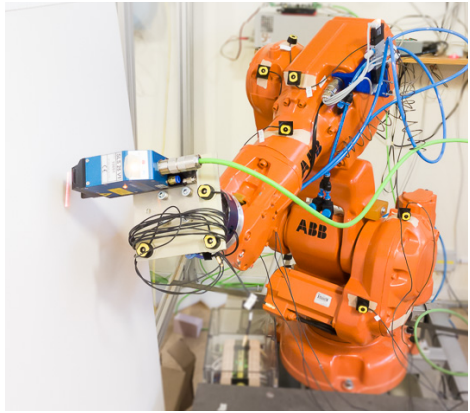

**Figure 13.2**  ABB IRB140 used for experimental verification. The sensor (blue) is mounted on the wrist and is plane of the laser light is intersecting a white flat surface.

of the robot, $T_{TF}^S$, is needed.

A naive approach to the stated calibration problem is to make use of the 4/5/6-point tool calibration routines commonly found in industrial robot systems. This amounts to several times positioning the origin of the sensor at a fixed point in the workspace, with varying orientation. These methods suffer from the fact that the origin and the axes of the sensor coordinate system are invisible to the operator, which must rely on visual feedback from both the workspace and a computer monitor simultaneously. Further, the accuracy of these methods is very much dependent on the skill of the operator and data collection for even a small amount of points is very tedious.

Sensor manufacturers use special-purpose calibration objects and jigs that are machined to high precision [Meta Vision Systems, 2014; SICK IVP, 2011], a method largely unavailable to a user.

Other well known algorithms for eye-to-hand calibration include [Daniilidis, 1999; Tsai and Lenz, 1989; Horaud and Dornaika, 1995], which are all adopted for calibration of a wrist-mounted camera using a calibration pattern such as a checkerboard pattern. A laser scanner is fundamentally different in the information it captures and can not determine the pose of a checkerboard pattern. This must be considered by the calibration algorithm employed.

The proposed method will make use of data collected from planar surfaces as a very simple form of calibration object. Kinematic calibration of robotic manipulator using planar constraints in various formats has been considered before. Zhuang et al. (1999), propose a method that begins with an initial estimate of the desired parameters, which is improved with a non-linear optimization algorithm. They also discuss observability issues related to identification using planar constraints. The method focuses on improving parameter estimates in the kine-matic model of the robot, and convergence results are therefore only presented for initial guesses very close to their true values (0.01mm/0.01°). Initial estimates this accurate are very hard to obtain unless a very precise CAD model of all parts involved is available.

Zhang and Pless (2004) found the transformation between a camera and a laser range finder using a checkerboard pattern and used computer vision to estimate the location of the calibration planes. With the equations of the calibration planes known, the desired transformation matrix was obtained from a set of linear equations.

Planar constraints have also been considered by Ikits and Hollerbach (1997) who employed a non-linear optimization technique to estimate the kinematic parameters. The method requires careful definition of the planes and can not handle arbitrary frame assignments.

A wrist mounted sensor can be seen as an extension of the kinematic chain of the robot. Initial guesses can be poor, especially if based on visual estimates. This section presents a method based solely on solving linear sets of equations. The method accepts a very crude initial estimate of the desired kinematic parameters, which is refined in an iterative procedure. The placement of the calibration planes is assumed unknown, and their locations are found automatically together with the desired transformation matrix. The exposition will be focused on sensors measuring distances in a plane, but extensions to the proposed method to 3D laser scanners such as LIDARs and 1D point lasers will be provided towards the end.

### Preliminaries

Throughout this section, the kinematic notation presented in Sec. 12.1 will be used. The normal of a plane from which measurement point $i$ is taken, given in frame $A$, will be denoted $n_A^i$.

A plane is completely specified by

$$n^\mathsf{T} p = d, \quad \|n\|_2 = 1 \tag{13.12}$$

where $d$ is the orthogonal distance from the origin to the plane, $n$ the plane normal and $p$ is any point on the plane.

### Laser-scanner characteristics

The laser scanner consists of an optical sensor and a laser source emitting light in a plane that intersects a physical plane in a line, see Fig. 13.2. The three dimensional location of a point along the projected laser line may be calculated by triangulation, based on the known geometry between the camera and the laser emitter. A single measurement from the laser scanner typically yields the coordinates of a large number of points in the laser plane. Alternatively, a measurement consists of a single point and the angle of the surface, which is easily converted to two points by sampling a second point on the line through the given point. Comments on statistical considerations for this sampling are provided in Sec. 13.A.

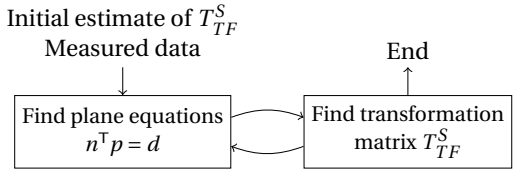

**Figure 13.3** Illustration of the two-step, iterative method.

### Method

The objective of the calibration is to find the transformation matrix $T_{TF}^S \in SE(3)$ that relates the measurements of the laser scanner to the coordinate frame of the tool flange of the robot.

The kinematic chain of a robot will here consist of the transformation between the robot base frame and the tool flange $T_{RB}^{TF_i}$, given by the forward kinematics[1] in pose $i$, and the transformation between the tool flange and the sensor $T_{TF}^S$. The sensor, in turn, projects laser light onto a plane with unknown equation. A point observed by the sensor can be translated to robot base frame by

$$\mathbf{p}_{RB_i} = T_{RB}^{TF_i} T_{TF}^S \mathbf{p}_{S_i} \tag{13.13}$$

where $i$ denotes the index of the pose.

To find $T_{TF}^S$, an iterative two-step method is proposed, which starts with an initial guess of the matrix. In each iteration, the equations for the planes are found using eigendecomposition, whereafter a set of linear equations is solved for an improved estimate of the desired transformation matrix. The scheme, illustrated in Fig. 13.3, is iterated until convergence.

***Finding the calibration planes***   Consider initially a set of measurements, $\mathcal{P}_S = [p_1, ..., p_{N_p}]_S$, gathered from a single plane. The normal can be found by Principal Component Analysis (PCA), which amounts to performing an eigendecomposition of the covariance matrix $C$ of the points [Pearson, 1901]. The eigenvector corresponding to the smallest eigenvalue of $C$ will be the desired estimate of the plane normal.[2] To this purpose, all points are transformed to a common frame, the robot base frame, using (13.13) and the current estimate of $T_{TF}^S$.

To fully specify the plane equation, the center of mass $\mu$ of $\mathcal{P}_{RB}$ is calculated. The distance $d$ to the plane is then calculated as the length of the projection of the vector $\mu$ onto the plane normal

$$d = \left\| \bar{n}(\bar{n}^{\mathsf{T}}\mu) \right\| \tag{13.14}$$

where $\bar{n}$ is a normal with unit length given by PCA. This distance can be encoded into the normal by letting $\|n\| = d$. The normal is then simply found by

$$n = \bar{n}(\bar{n}^{\mathsf{T}}\mu) \tag{13.15}$$

---

[1] Or alternatively, given by an external tracking system.

[2] This eigenvalue will correspond to the mean squared distance from the points to the plane.

This procedure is repeated for all measured calibration planes and results in a set of normals that will be used to find the optimal $T_{TF}^{S}$.

**Solving for $T_{TF}^{S}$**   All measured points should fulfill the equation for the plane they were obtained from. This means that for a single point $p$, the following must hold

$$\bar{n}^{\mathsf{T}} p = d \Longleftrightarrow n^{\mathsf{T}} p = \|n\|^2 \tag{13.16}$$

A measurement point obtained from the sensor in the considered setup should thus fulfill the following set of linear equations

$$\mathbf{p}_{RB_i} = T_{RB}^{TF_i} T_{TF}^{S} \mathbf{p}_{S_i} \tag{13.17}$$

$$n^{\mathsf{T}} p_{RB_i} = \|n\|^2 \tag{13.18}$$

$$\mathbf{p}_{S_i} = \begin{bmatrix} p_{S_i}^{\mathsf{T}} & 1 \end{bmatrix}^{\mathsf{T}} = \begin{bmatrix} x_{S_i} & y_{S_i} & z_{S_i} & 1 \end{bmatrix}^{\mathsf{T}} \tag{13.19}$$

where bold-face notation denotes a point expressed in homogeneous coordinates according to (13.19). Without loss of generality, the points $p_S$ can be assumed to lie in the plane $z_S = 0$. As a result, the third column in $T_{TF}^{S}$ can not be solved for directly. The constraints on $R_{TF}^{S}$ to belong to $SO(3)$, will however allow for reconstruction of the third column in $R_{TF}^{S}$ from the first two columns.

Let $\tilde{T}$ denote the remainder of $T_{TF}^{S}$ after removing the third column and the last row. The linear equations (13.17)-(13.18) can be expressed as

$$A_i k = \|n_i\| - q_i \tag{13.20}$$

where $k = \text{vec}(\tilde{T}) \in \mathbb{R}^{9 \times 1}$ consists of the stacked columns of $\tilde{T}$ and

$$A_i = \begin{bmatrix} n_i^{\mathsf{T}} R_{RB}^{TF_i} x_{S_i} & n_i^{\mathsf{T}} R_{RB}^{TF_i} y_{S_i} & n_i^{\mathsf{T}} R_{RB}^{TF_i} \end{bmatrix} \qquad \in \mathbb{R}^{1 \times 9} \tag{13.21}$$

$$q_i = n_i^{\mathsf{T}} p_{RB}^{TF_i} \qquad \in \mathbb{R} \tag{13.22}$$

Since (13.17) and (13.18) are linear in the parameters, all elements of $T_{TF}^{S}$ can be extracted into $k$, and $A_i$ can be obtained by performing the matrix multiplications in (13.17) and (13.18) and identifying terms containing any of the elements of $k$. Terms which do not include any parameter to be identified are associated with $q_i$. The final expressions for $A_i$ and $q_i$ given above can then be obtained by identifying matrix-multiplication structures among the elements of $A_i$ and $q_i$.

Equation (13.20) does not have a unique solution. A set of at least nine points gathered from at least three planes is required in order to obtain a unique solution to the vector $k$. This can be obtained by stacking the entries in (13.20) according to

$$\mathbf{A} k = \mathbf{Y}, \quad \mathbf{A} = \begin{bmatrix} A_1 \\ A_2 \\ \vdots \\ A_{N_p} \end{bmatrix}, \quad \mathbf{Y} = \begin{bmatrix} \|n_1\| - q_1 \\ \|n_2\| - q_2 \\ \vdots \\ \|n_{N_p}\| - q_{N_p} \end{bmatrix} \tag{13.23}$$

The resulting problem is linear in the parameters, and the vector $k^*$ of parameters that minimizes

$$k^* = \underset{k}{\arg\min} \left\| \mathbf{A}k - \mathbf{Y} \right\|_2 \tag{13.24}$$

can then be obtained from the least-squares procedure of Sec. 6.2. We make a note at this point that while solving (13.24) is the most straightforward way of obtaining an estimate, the problem contains errors in both **A** and **Y** and is therefore of errors-in-variables type and a candidate for the total least-squares procedure. We discuss the two solution methods further in Sec. 13.B.

Since $k$ only contains the first two columns of $R_{TF}^S$, the third column is formed as

$$R_3 = R_1 \times R_2 \tag{13.25}$$

where $\times$ denotes the cross product between $R_1$ and $R_2$, which produces a vector orthogonal to both $R_1$ and $R_2$. When solving (13.24) we are actually solving a problem with the same relaxation as the one employed in Sec. 13.1, since the constraints $R_1^\mathsf{T} R_2 = 0$, $\|R_1\| = \|R_2\| = 1$ are not enforced. The resulting $R_{TF}^S$ will thus in general not belong to $SO(3)$. The closest valid rotation matrix can be found by projection onto $SO(3)$ according to the procedure outlined in Sec. 6.1.

This projection will change the corresponding entries in $k^*$ and the resulting coefficients will no longer solve the problem (13.24). A second optimization problem can thus be formed to re-estimate the translational part of $k$, given the orthogonalized rotational part. Let $k$ be decomposed according to

$$k = \begin{bmatrix} \tilde{R}^* & p \end{bmatrix} \quad \tilde{R}^* \in \mathbb{R}^{1 \times 6}, \ p \in \mathbb{R}^{1 \times 3} \tag{13.26}$$

and denote by $\mathbf{A}_{n:k}$ columns $n$ to $k$ of **A**. The optimal translational vector, given the orthonormal rotation matrix, is found by solving the following optimization problem

$$\tilde{\mathbf{Y}} = \mathbf{Y} - \mathbf{A}_{1:6} \tilde{R}^* \tag{13.27}$$

$$p^* = \underset{p}{\arg\min} \left\| \mathbf{A}_{7:9} p - \tilde{\mathbf{Y}} \right\|_2 \tag{13.28}$$

***Final refinement***   As noted by Zhang and Pless (2004), solving an optimization problem like (13.24) is equivalent to minimizing the algebraic distance between the matrix, parameterized by $k$, and the data. There is no direct minimization of the distances from measurements to planes involved. Given the result from the above procedure as initial guess, any suitable, iterative minimization strategy can be employed to further minimize a cost function on the form

$$J(T_{TF}^S) = \sum_{i=1}^{N_p} (n_i^\mathsf{T} p_{RB_i}(T_{TF}^S) - \|n_i\|)^2 \tag{13.29}$$

which is the sum of squared distances from the measurement points to the planes. Here, $p_{RB_i}$ is seen as a function of $T_{TF}^S$ according to (13.17).

The simulation experiments presented in the next section did not show any significant improvement to the estimated transformation matrix from this additional refinement, and we thus refrain from exploring this topic further.

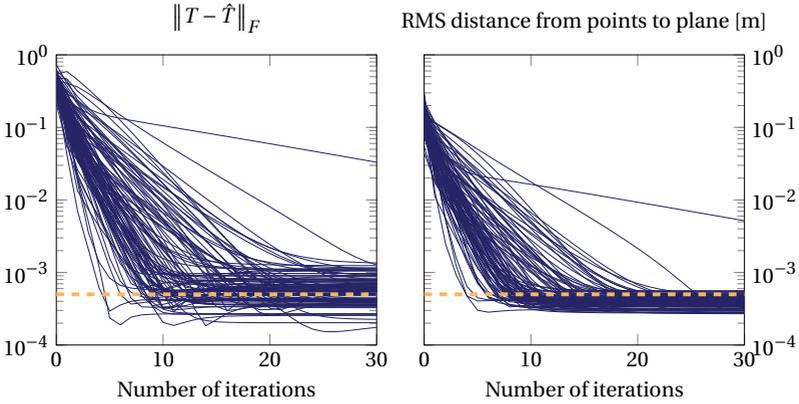

**Figure 13.4** Convergence results for simulated data during 100 realizations, each with 10 poses sampled from each of 3 planes. Measurement noise level $\sigma = 0.5$mm is marked with a dashed line. On the left, the Frobenius norm between the true matrix and the estimated, on the right, the RMS distance between measurement points and the estimated plane.

### Results

The performance of the method was initially studied in simulations, which allows for a comparison between the obtained estimate and the ground truth. The simulation study is followed by experimental verification using a real laser scanner mounted on the wrist of an industrial manipulator.

***Simulations*** To study the convergence properties of the proposed approach, a simulation study was conducted. A randomly generated $T_{TF}^{S}$ was used together with measurements from a set of random poses. The initial guess of $T_{TF}^{S}$ was chosen as the true matrix corrupted with an error distributed uniformly, with $x, y, z \sim \mathcal{U}(-200\text{mm}, 200\text{mm})$ and roll, pitch, yaw$\sim \mathcal{U}(-30°, 30°)$. The measurements were obtained from three orthogonal planes and corrupted with Gaussian white noise with standard deviation $\sigma = 0.5$mm.

Figure 13.4 illustrates the convergence for 100 realizations of the described procedure. Most realizations converged to the true matrix within 15 iterations. Analysis shows that careful selection of poses results in faster convergence. The random pose-selection strategy employed in the simulation study suffers the risk of co-linearity between measurement poses, which slows down convergence.

Figure 13.5 illustrates the final results in terms of the accuracy in both the translational and rotational part of the estimate of $T_{TF}^{S}$.

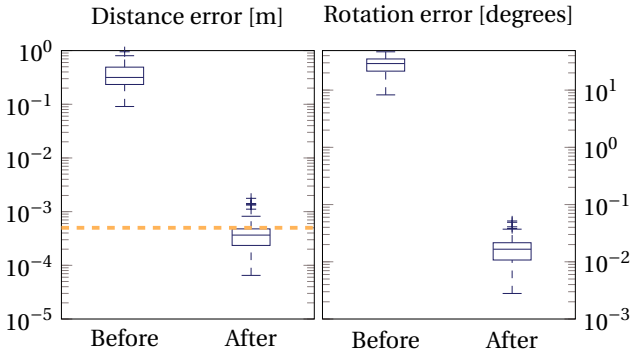

**Figure 13.5**   Errors in $T_{TF}^{S}$ before and after calibration for 100 realizations with 30 calibration iterations. For each realization, 10 poses were sampled from each of 3 planes. The measurement noise level $\sigma = 0.5$mm is marked with a dashed line. On the left, the translational error between the true matrix and the estimated, on the right, the rotational error.

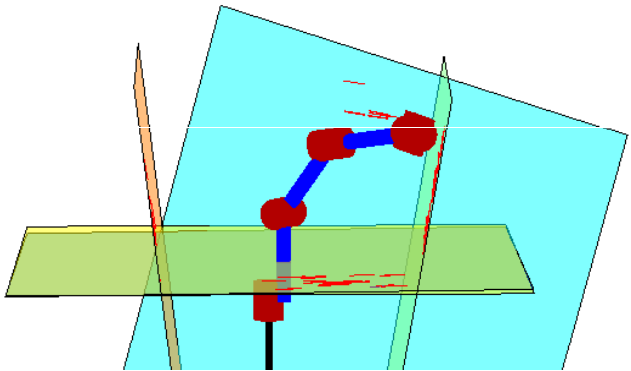

**Figure 13.6**   A visualization of the reconstructed planes used for data collection. The planes were placed so as to be close to orthogonal to each other, surrounding the robot. Figure 13.2 presents a photo of the setup.

***Experiments***   Experimental verification of the proposed method was conducted with an ABB IRB140 robot equipped with a Meta SLS 25 [Meta Vision Systems, 2014] weld seam-tracking sensor, see Fig. 13.2. A flat whiteboard was placed on different locations surrounding the robot, see Fig. 13.6, and several measurements of each plane were recorded.

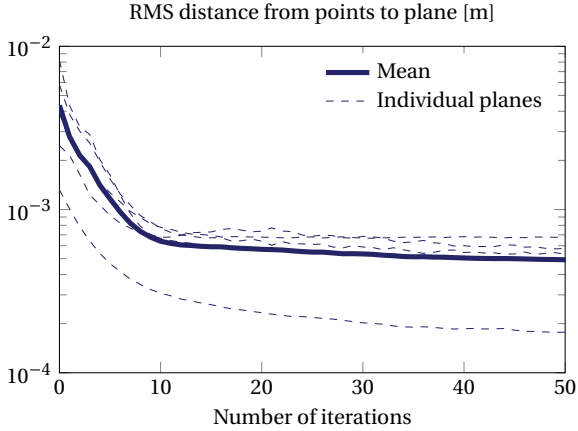

**Figure 13.7**   Convergence results for experimental data gathered from 5 planes. The RMS distance between measurement points and the estimated planes are shown together with the mean over all planes.

The algorithm was started with the initial guess

$$
T_{TF}^S = \begin{bmatrix} 1 & 0 & 0 & 0 \\ 0 & 1 & 0 & 0.15 \\ 0 & 0 & 1 & 0.15 \\ 0 & 0 & 0 & 1 \end{bmatrix}
\tag{13.30}
$$

and returned the final estimate

$$
T_{TF}^S = \begin{bmatrix} 0.9620 & 0.2710 & 0.0010 & 0.0850 \\ -0.2710 & 0.9620 & -0.0240 & 0.1170 \\ -0.0070 & 0.0230 & 1.0000 & 0.1610 \\ 0 & 0 & 0 & 1 \end{bmatrix}
\tag{13.31}
$$

The translational part of the initial guess was obtained by estimating the distance from the tool flange to the origin of the laser scanner, whereas the rotation matrix was obtained by estimating the projection of the coordinate axes of the scanner onto the axes of the tool flange.[3]

The convergence behavior, illustrated in Fig. 13.7, is similar to that in the simulation, and the final error was on the same level as the noise in the sensor data. A histogram of the final errors is shown in Fig. 13.8, indicating a symmetric but heavy-tailed distribution. We remark that if a figure like Fig. 13.8 indicate the presence of outliers or errors with a highly non-Gaussian distribution, one could consider alternative distance metrics in (13.24), such as the $L_1$ norm, for a more robust performance.

---

[3] The fact that the initial estimate of the rotation matrix was the identity matrix is a coincidence.

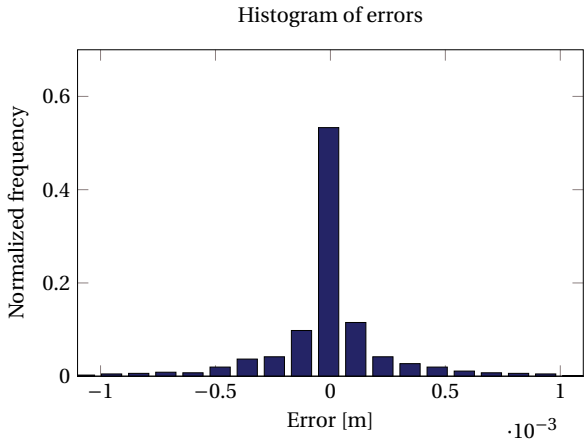

**Figure 13.8** Histogram of errors $\mathbf{Y} - \mathbf{A}w^*$ for the experimental data.

### Discussion

The calibration method described is highly practically motivated. Calibration is often tedious and an annoyance to the operator of any technical equipment. The method described tries to mitigate this problem by making use of data that is easy to acquire. In its simplest form, the algorithm requires some minor bookkeeping in order to associate the recorded points with the plane they were collected from. An extension to the algorithm that would make it even more accessible is automatic identification of planes using a RANSAC [Fischler and Bolles, 1981] or clustering procedure.

While the method was shown to be robust against large initial errors in the estimated transform, the effect of a nonzero curvature of the planes from which the data is gathered remains to be investigated.

In practice, physical limitations may limit the poses from which data can be gathered, i.e., neither the sensor nor the robot must collide with the planes. The failure of the collected poses to adequately span the relevant space may make the algorithm sensitive to errors in certain directions in the space of estimated parameters. Fortunately, however, directions that are difficult to span when gathering calibration data are also unlikely to be useful when the calibrated sensor is deployed. The found transformation will consequently be accurate for relevant poses that were present in the calibration data.

### Conclusions

This section has presented a robust, iterative method composed of linear sub problems for the kinematic calibration of a 2D laser sensor. Large uncertainties in the initial estimates are handled and the estimation error converges to below the level of the measurement noise. The calibration routine can be used for any type of laser sensor that measures distances in a plane, as long as the forward kinematics is known, such as when the sensor is mounted on the flange of an industrial robot or on a mobile platform or aerial drone, tracked by an external tracking system. Extensions to other kinds of laser sensors are provided in Sections 13.C and 13.D.

An implementation of the proposed method is made available in [*Robotlib.jl*, B.C., 2015].

## Appendix A. Point Sampling

Some laser sensors for weld-seam tracking do not provide all the measured points, but instead provide the location of the weld seam and the angle of the surface. In this situation one must sample a second point along the line implied by the provided measurements in order for the proposed algorithm to work. The sampling of the second point, $p_2$, is straight forward, but entails a trade-off between noise sensitivity and numerical accuracy. Given measurements of a point and an angle, $p_m, \alpha_m$, corrupted with measurement noise $e_p, e_\alpha$, respectively, $p_2$ can be calculated as

$$p_2 = p_m + \gamma l \tag{13.32}$$

$$l = \begin{bmatrix} \cos \alpha_m \\ \sin \alpha_m \end{bmatrix} \tag{13.33}$$

$$\alpha_m = \alpha + e_\alpha, \quad e_\alpha \sim \mathcal{N}(0, \sigma_\alpha^2) \tag{13.34}$$

$$p_m = p + e_p, \quad e_p \sim \mathcal{N}(0, \Sigma_p) \tag{13.35}$$

where $\gamma$ is a constant determining the offset between $p_m$ and $p_2$. If this offset is very small, the condition number of $\mathbf{A}$ will be large and the estimation accuracy will be impacted. If $\gamma$ is chosen very large, the variance in $p_m$ and $p_2$ will be very different. The variance of $p_2$ is given by

$$\mathbb{E}\left\{p_2 p_2^\mathsf{T}\right\} = \mathbb{V}\left\{p_2\right\} = \mathbb{V}\left\{p_m + \gamma l\right\} = \mathbb{V}\left\{p_m\right\} + \gamma^2 \mathbb{E}\left\{l l^\mathsf{T}\right\} \tag{13.36}$$

$$\approx \Sigma_p + \gamma^2 \sigma_\alpha^2 \left|\nabla_\alpha l \nabla_\alpha l^\mathsf{T}\right| \tag{13.37}$$

$$= \Sigma_p + \gamma^2 \sigma_\alpha^2 \left|\begin{bmatrix} \sin^2 \alpha_m & -\sin \alpha_m \cos \alpha_m \\ -\sin \alpha_m \cos \alpha_m & \cos^2 \alpha_m \end{bmatrix}\right| \tag{13.38}$$

where (13.36) holds if $\mathbb{E}\left\{e_\alpha e_p\right\} = 0$. We thus see that the variance in the sampled point will be strictly larger than the variance in the measured point, and we should ideally trust this second point less when we form the estimate of both the plane equations and $T_{TF}^S$. Since the original problem is of errors-in-variables type, the optimal solution to the problem with unequal variances is given by the weighted total least-squares procedure [Fang, 2013]. Experiments outlined in the Sec. 13.B indicated that as long as $\gamma$ was chosen on the same scale as $\|p_S\|$, taking this uncertainty into account had no effect on the fixed point to which the algorithm converged.

## Appendix B. Least-Squares vs. Total Least-Squares

In the main text, we only made a brief comment on the potential methods of estimating $T_{RB}^{TF}$ given the matrices **A** and **Y**. While the least-squares procedure corresponds to the ML estimate for a model where only the right-hand side **Y** is corrupted with Gaussian noise, this does not hold true when also **A** is corrupted with noise. In the present context we have

$$A_i k = \|n_i\| - n_i^\mathsf{T} p_{RB}^{TF_i} \tag{13.39}$$

$$A_i = \begin{bmatrix} n_i^\mathsf{T} R_{RB}^{TF_i} x_{S_i} & n_i^\mathsf{T} R_{RB}^{TF_i} y_{S_i} & n_i^\mathsf{T} R_{RB}^{TF_i} \end{bmatrix} \tag{13.40}$$

where significant measurement errors appear in $x_S$ and $y_S$. The variables $n_i$ are estimated using the current estimate of $T_{TF}^S$ and will thus be corrupted by both estimation errors and measurement errors. Unfortunately, estimating the variance of this error is non-trivial, and correct application of the total least-squares procedure is thus hard. We propose two alternative strategies:

*Alt. 1*   An approximate strategy is obtained if we assume that the variance of $x_S$ and $y_S$ is negligible in comparison to the variance in $n_i$. We further make the simplifying assumption that $n_i$ is normally distributed with the covariance matrix $\Sigma_{n_i}$. To obtain an estimate of $\Sigma_{n_i}$ one can, e.g., perform statistical bootstrapping. This approach requires as many bootstrapping procedures as the number of planes data is sampled from.

*Alt. 2*   Another strategy is obtained by estimating the complete covariance matrix $\Sigma_{Ay}$ of $\begin{bmatrix} A_i & y_i \end{bmatrix}^\mathsf{T}$ using statistical bootstrapping [Murphy, 2012]. This approach makes less assumptions than the first approach. While conceptually simple, this approach requires $N_p$ bootstrapping procedures and is thus more computationally expensive. One can, however, run the complete algorithm until convergence using estimates based on regular LS, and first after convergence switch to the more accurate WTLS procedure to improve accuracy.

### Evaluation

The first strategy was implemented and tested on the simulated data from Sec. 13.2. We used 500 bootstrap samples and 15 iterations in the WTLS algorithm. No appreciable difference in fixed points was detected between solving for $T_{TF}^S$ using the standard LS procedure and the WTLS procedure. We thus concluded that the considerable additional complexity involved in estimating the covariance matrices of the data and solving the optimization problem using the more complicated WTLS algorithm is not warranted. The evaluation can be reproduced using code available in [*Robotlib.jl*, B.C., 2015] and the WTLS algorithm implementation is made available in [*TotalLeastSquares.jl*, B.C., 2018].

## Appendix C. Calibration of Point Lasers

In this section, we briefly consider the changes required to the proposed algorithm for sensors measuring the distance to a single point only.

We can for a point laser only determine the equation for the laser line, as opposed to before when we could find the laser light plane and thus the entire calibration matrix in $SE(3)$.

Once more, data is collected from three planar surfaces. The required modifications to the proposed algorithm are listed below

**Eq. (13.19)** We now assume, without loss of generality, that the laser point lies along the line $y_S = z_S = 0$. As a result, the second and third columns of $T_{TF}^S$ can not be solved for. These two vectors can be set to zero.

**Eq. (13.20)** The truncated vector $k \in \mathbb{R}^6$ will now consist of the first column of $R_{TF}^S$ and the translation vector $p_{TF}^S$.

**Eq. (13.21)** The three middle elements of $A_i$, corresponding to $n_i^\intercal R_{RB}^{TF_i} y_{S_i}$ are removed.

**Orthogonalization** The orthogonalization procedure reduces to the normalization of the first three elements of $k$ to have norm one.

The rest of the algorithm proceeds according to the original formulation.

## Appendix D. Calibration of 3D Lasers and LIDARs

The proposed algorithm can be utilized also for 3D distance sensors, such as LIDARs. This type of sensor provides richer information about the environment, and thus allows a richer set of calibration algorithms to be employed. We make no claims regarding the effectiveness of the proposed algorithm in this scenario, and simply note the adjustments needed to employ it.

To use the proposed algorithm, we modify it according to

**Eq. (13.19)** We no longer assume that the laser line lies in the plane $z_S = 0$. As a result, the full matrix $T_{TF}^S$ can now be solved for immediately, without the additional step of forming $R_3 = R_1 \times R_2$.

**Eq. (13.20)** The vector $k \in \mathbb{R}^{12}$ will now consist of all the columns of $R_{TF}^S$ and the translation vector $p_{TF}^S$.

**Eq. (13.21)** The regressor will now consist of

$$A_i = \begin{bmatrix} n_i^\intercal R_{RB}^{TF_i} x_{S_i} & n_i^\intercal R_{RB}^{TF_i} y_{S_i} & n_i^\intercal R_{RB}^{TF_i} z_{S_i} & n_i^\intercal R_{RB}^{TF_i} \end{bmatrix} \in \mathbb{R}^{1 \times 12}$$

To make use of this algorithm in practice, one has to consider the problem of assigning measured points to the correct planes. This can be hard when employing 3D sensors with a large field-of-view. Potential strategies include pre-estimation of planes in the sensor coordinate system using, e.g., the RANSAC algorithm, or manual segmentation.

# 14

# State Estimation for FSW

In this chapter, we will consider the problem of state estimation in the context of friction stir welding. The state estimator we develop is able to incorporate measurements from the class of laser sensors that was considered in the previous chapter, as well as compliance models and force-sensor measurements. As alluded to in Chap. 12, the FSW process requires accurate control of the full 6 DOF pose of the robot relative to the seam. The problem of estimating the pose of the welding tool during welding is both nonlinear and non-Gaussian, motivating a state estimator beyond the standard Kalman filter. The following sections will highlight the unique challenges related to state estimation associated with friction stir welding and propose a particle-filter based estimator, a method that was introduced in Sec. 4.2.

We also develop a framework for seam-tracking simulation in Sec. 14.2, where the relation between sources of error and estimation performance is analyzed. Through geometric reasoning, we show that some situations call for additional sensing on top of what is provided by a single laser sensor. The framework is intended to assist the user in selection of an optimal sensor configuration for a given seam, where sensor configurations vary in, e.g., the number of sensors applied and their distance from the tool center point (TCP). The framework also helps the user tune the state estimator, a problem which is significantly harder for a particle-filter based estimator compared to a Kalman filter.

## 14.1  State Estimator

We will start our exposition by choosing a state representation, describing the probability density functions used in the state transition and measurement update steps as well as cover some practical implementation details.

A natural state to consider in robotics is the set of joint angles, $q$, and their velocities, $\dot{q}$. In the context of robotic machining in general, and FSW in particular, deflections in the kinematic structure due to process forces unfortunately invalidate the joint angles as an accurate description of the robot pose. The proposed state estimator will therefore work in the space $SE(3)$, represented as $4 \times 4$ transformation matrices, which further allows for a natural inclusion of sensor measurements that naturally occupy the same Cartesian space as the tool pose we are ultimately interested in estimating. Although subject to errors, the sensor information available from the robot is naturally transformed to $SE(3)$ by means of the forward kinematics function $F_k(q)$. This information will be used to increase the observability in directions of the state-space where the external sensing leaves us blind.

The velocities and accelerations present during FSW are typically very low and we therefore chose to not include velocities in the state to be estimated. This reduces the state dimension and computational burden significantly, while maintaining a high estimation performance.

### Preliminaries

This section briefly introduces a number of coordinate frames and variables used in the later description of the method. For a general treatment of coordinate frames and transformations, see [Murray et al., 1994].

The following text will reference a number of different coordinate frames. We list them all in Table 12.1 and provide a visual guide to relate them to each other in Fig. 14.1. Table 12.1 further introduces a number of variables and some special notation that will be referred to in the following description. All Cartesian-space variables are given in the robot base frame $\mathcal{RB}$ unless otherwise noted.

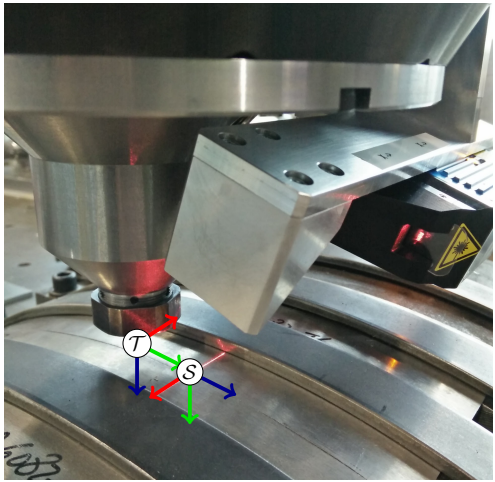

**Figure 14.1** Coordinate frames $(x, y, z)$ = (red, green, blue). The origin of the sensor frame $\mathcal{S}$ is located in the laser plane at the desired seam intersection point. The tool frame is denoted by $\mathcal{T}$.

**Nominal trajectory**

Before we describe the details of the state estimator, we will establish the concept of the *nominal trajectory*. In the linear-Gaussian case, the reference trajectory of a control system is of no importance while estimating the state, this follows from (4.24), which shows that the covariance of the state estimate is independent of the control signal. In the present context, however, we make use of the reference trajectory for two purposes. 1.) It provides prior information regarding the state transition. This lets us bypass a lot of modeling complexity by assuming that the robot controller will do a good job following the velocities specified in the reference trajectory. We know, however, that the robot controller will follow this reference with a potentially large position error, due to deflections, etc., outlined above. 2.) The reference trajectory provides the *nominal seam geometry* needed to determine the likelihood of a measurement from the laser sensor given a state hypothesis $\hat{X}$.

To get a suitable representation of the nominal trajectory used to propagate the particles forward, we can, e.g., perform a simulation of the robot program using a simulation software, often provided by the robot manufacturer. This procedure eliminates the need to reverse engineer the robot path planner. During the simulation, a stream of joint angles is saved, which, when run through the forward kinematics, returns the nominal trajectory in Cartesian space. The simulation framework outlined in the following sections provides a number of methods for generating a nominal trajectory for simulation experiments.

The nominal trajectory will consist of a sequence of joint coordinates $\{\bar{q}_t\}_{t=1}^{N}$ which, if run through the forward kinematic function, yields a sequence of points specifying the nominal seam geometry $\{p_t = F_k(\bar{q}_t)\}_{t=1}^{N}$.

**Density functions**

To employ a particle filter for state estimation, specification of a statistical model of the state evolution and the measurement generating process is required, see Sec. 4.2. This section introduces and motivates the various modeling choices in the form of probability density functions used in the proposed particle filter.

***State transition***

$$p(X^+|X, \mathfrak{f}) \tag{14.1}$$

The state-transition function is a model of the state at the next time instance, given the state at the current time instance. We model the mean of the state-transition density (14.1) using the robot reference trajectory. The reference trajectory is generated by, e.g., the robot controller or FSW path planner, and consists of a sequence of poses which the robot should traverse. We assume that a tracking controller can make corrections to this nominal trajectory, based on the state estimates from the state estimator.

We denote by $T^+$ the incremental transformation from $F_k(\bar{q})$ to $F_k(\bar{q}^+)$ such that

$$F_k(\bar{q}^+) = T^+ F_k(\bar{q})$$

The mean of the state transition density is thus given by

$$\mu\{p(X^+|X,\mathfrak{f})\} = T^+ = F_k(\bar{q}^+)\,F_k(\bar{q})^{-1}$$

The shape of the density should encode the uncertainty in the update of the robot state from one sample to another. For a robot moving in free space, this uncertainty is usually very small. Under the influence of varying external process forces, however, significant uncertainty is introduced [De Backer, 2014; Sörnmo, 2015; Olofsson, 2015]. Based on this assumption, we may choose a density where the width is a function of the process force. For example, we may chose a multivariate Gaussian distribution and let the covariance matrix be a function of the process force.

***Robot measurement update***

$$p(q,\mathfrak{f}|X) \tag{14.2}$$

When the robot is subject to large external forces applied at the tool, the measurements provided by the robot will not provide an accurate estimate of the tool pose through the forward-kinematics function. If a compliance model $C_j(\tau)$ is available, we may use it to reduce the uncertainty induced by kinematic deflections, a topic explored in detail in [Lehmann et al., 2013; Sörnmo, 2015; Olofsson, 2015]. We thus choose the following model for the mean of the robot measurement density (14.2)

$$\mu\{p(q,\mathfrak{f}|X)\} = F_k(q + C_j(\tau)) \tag{14.3}$$

The uncertainty in the robot measurement comes from several sources. The joint resolvers/encoders are affected by noise, which is well modeled as a Gaussian random variable. When Gaussian errors, $e_q$, in the joint measurements are propagated through the linearized forward-kinematics function, the covariance matrix $\Sigma_C$ of the resulting Cartesian-space errors $e_C$ is obtained by approximating $e_q = dq$ as

$$\begin{aligned}
q_m &= q + e_q = q + dq \\
e_q &\sim \mathcal{N}(0,\Sigma_q) \\
e_C &\sim \mathcal{N}(0,J\Sigma_q J^{\mathsf{T}})
\end{aligned}$$

where $q_m$ is the measured value. The corresponding Cartesian-space covariance matrix is given by

$$e_C = \frac{d\langle F_k(q)\rangle^\vee}{dq}\, dq = J\, dq = J\, e_q$$

$$\Sigma_C = \mathbb{E}\left\{e_C\, e_C^\mathsf{T}\right\} = \mathbb{E}\left\{J\, e_q\, e_q^\mathsf{T} J^\mathsf{T}\right\} = J\,\mathbb{E}\left\{e_q\, e_q^\mathsf{T}\right\} J^\mathsf{T}$$

where the approximation $J(q + e_q) \approx J(q)$ has been made. The twist coordinate representation $\langle F_k(q)\rangle^\vee$ is obtained by taking the logarithm of the transformation matrix $\log(F_k(q))$, which produces a twist $\xi \in se(3)$, and the operation $\xi^\vee \in \mathbb{R}^6$ returns the twist coordinates [Murray et al., 1994].[1]

Except for the measurement noise $e_q$, the errors in the robot measurement update density are not independent between samples. The error in both the forward kinematics and the compliance model is configuration dependent. Since the velocity of the robot is bounded, the configuration will change slowly and configuration-dependent errors will thus be highly correlated in time. The standard derivation of the particle filter relies on the assumption that the measurement errors constitute a sequence of independent, identically distributed (i.i.d.) random variables. Independent measurement errors can be averaged between samples to obtain a more accurate estimate, which is not possible with correlated errors, where several consecutive measurements all suffer from the same error.

Time-correlated errors are in general hard to handle in the particle filtering framework and no systematic way to cope with this problem has been found. One potential approach is to incorporate the correlated error as a state to be estimated [Evensen, 2003; Åström and Wittenmark, 2013a]. This is feasible only if there exist a way to differentiate between the different sources of error, something which in the present context would require additional sensing. State augmentation further doubles the state dimension, with a corresponding increase in computational complexity.

Since only a combination of the tracking error, the kinematic error and the dynamic error is measurable, we propose to model the time-correlated uncertainties as a uniform random variable with a width $d$ chosen as the maximum expected error. When performing the measurement update with the densities of several perfectly correlated uniform random variables, the posterior distribution equals the prior distribution. The uniform distribution is thus invariant under the measurement update. We illustrate this in Fig. 14.2, where the effect of the measurement update is displayed for a hybrid between the Gaussian and uniform distributions.

The complete robot measurement density function with the above modeling choices, (14.2), is formed by the convolution of the densities for a Gaussian, $p_G$, and a uniform, $p_U$, random variable, according to

$$p(q, \mathring{f}|X) = \int_{\mathbb{R}^k} p_U(x - y)\, p_G(y)\, dy \tag{14.4}$$

---

[1] If the covariance of the measurements $q_m$ are is obtained on the motor-side of the gearbox, the Cartesian-space covariance will take the form $\Sigma_C = JG\,\mathbb{E}\left\{e_q\, e_q^\mathsf{T}\right\} G^\mathsf{T} J^\mathsf{T}$ where $G$ is the gear-ratio matrix of (9.11).

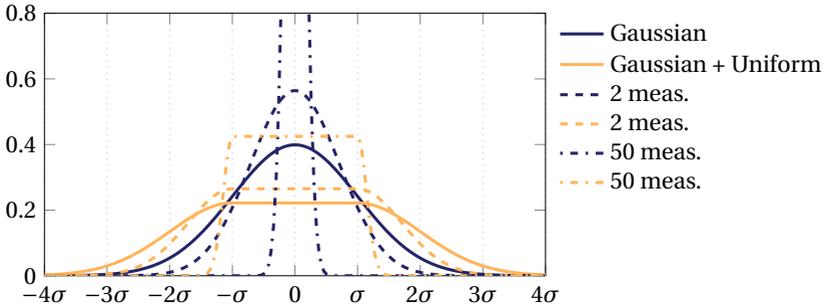

**Figure 14.2** Illustration of measurement densities and the posterior densities after several performed measurement updates. The figure illustrates the difference between a Gaussian distribution, for which the variance is reduced by each additional measurement update, and the proposed hybrid distribution with $d = \sigma$, for which the uniform part maintains its uncertainty after a measurement update.

where $k$ is the dimensionality of the state $x$. This integral has no closed-form solution, but can technically be evaluated numerically. Instead of evaluating (14.4), which is computationally expensive and must be done for every particle at every time step, we propose the following approximation

$$p(q, \mathfrak{f}|X) \approx \begin{cases} C & \text{if } |\Delta x| \leq d \\ C \exp\left(-\frac{(|\Delta x| - d)^2}{2\sigma^2}\right) & \text{if } |\Delta x| > d \end{cases} \tag{14.5}$$

with $\Delta x$ taken to be the element-wise difference between the positional coordinates of $X$ and a mean vector $\mu \in \mathbb{R}^3$, $\Delta x = x - \mu$, and the normalization constant

$$C = \frac{1}{\sqrt{2\pi}\sigma + 2d}$$

This approximation is a hybrid between the Gaussian and uniform distributions and reduces to the Gaussian distribution as the width of the uniform part $d \to 0$, and reduces to the uniform distribution as $\sigma \to 0$. Equation (14.5) is given for the one-dimensional case and one possible extension to higher dimensions is given

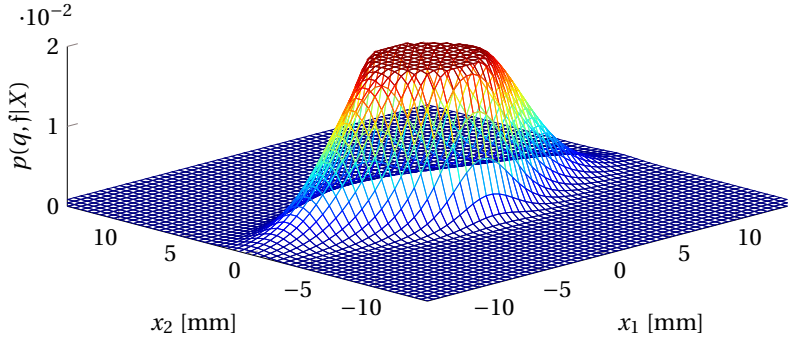

**Figure 14.3** Illustration of the multivariate version of the robot measurement density, (14.6).

by

$$p(q, \mathfrak{f}|X) = \begin{cases} D & \text{if } \left\| \Delta x \right\|_2 \leq d \\ D \exp\left(-\frac{1}{2}\delta x^\mathsf{T} \Sigma^{-1} \delta x\right) & \text{if } \left\| \Delta x \right\|_2 > d \end{cases} \tag{14.6}$$

$$\delta x = \left(1 - \frac{d}{\left\| \Delta x \right\|_2}\right) \Delta x$$

$$D = \frac{1}{(2\pi)^{\frac{k}{2}} \sqrt{\det(\Sigma)} + V(d,k)}$$

where $k$ is the state dimension and $V(d,k)$ is the volume of a $k$-dimensional sphere with radius $d$.

The univariate distribution, and the posterior distribution after several fused measurements, are shown in Fig. 14.2. An illustration of the multivariate case with

$$\Sigma = \begin{bmatrix} 4 & 0 \\ 0 & 1 \end{bmatrix}, \quad d = 3$$

is shown in Fig. 14.3.

The kinematic deflections that cause the uncertainty in the robot measurement are proportional to the external forces applied to the tool. The width of the uniform random variable $d = d(\mathfrak{f})$ is therefore chosen as a function of the process force

$$d(\mathfrak{f}) = d_0 + k_d \left\| \mathfrak{f} \right\|$$

where $d_0$ is a nominal uncertainty chosen with respect to the maximum absolute positioning error of the robot in the relevant work-space volume and $k_d \left\| \mathfrak{f} \right\|$ reflects the increase in uncertainty with the magnitude of the process force.

The discussion on the errors associated with the robot measurements is continued in more detail in Sec. 14.4.

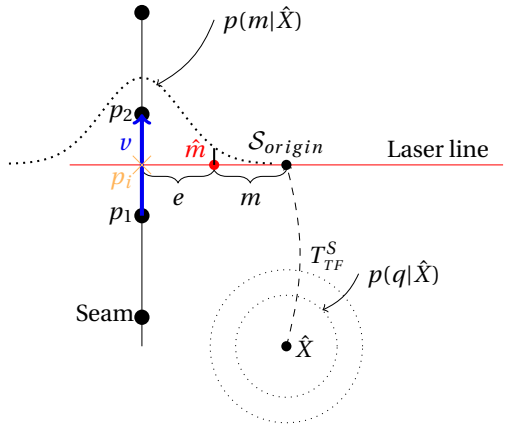

**Figure 14.4** Illustration of the relations between a particle $\hat{X}$ (TCP hypothesis), its belief about the location of the laser line and the laser measurement $m$ (14.7). Particles for which the distance, $e$, between the measurement hypothesis $\hat{m}$ and the seam intersection point $p_i$ is small, in terms of the distribution $p(m|X)$, are more likely to be correct estimates of the current state $X$. The points $p_1$ and $p_2$ are found by searching for the seam points closest to $\hat{m}$.

### *Laser sensor measurement update*

$$p(m|X)$$

A model for the measurement of the laser sensor is straightforward to describe, but harder to evaluate since it involves evaluating the distance between a measurement location and the nominal seam, i.e., determining how likely obtaining a particular measurement is given a state hypothesis and the prespecified seam geometry. Given a state hypothesis $\hat{X}$, the corresponding hypothesis for the intersection point between the seam and the laser plane, $\hat{m}$, is calculated using $T_{TF}^S$ according to

$$\hat{m} = (\hat{X} T_{TF}^S)_{1:3,4} + \begin{bmatrix} m \\ 0 \end{bmatrix} \tag{14.7}$$

If the prespecified seam geometry is given as a sequence of points $p_t$, we can evaluate the distance $e$ between $\hat{m}$ and the seam by performing a search for the closest nominal trajectory points. The error $e$ is then calculated as the projection of $\hat{m}$ onto the seam along the laser plane. The projected point $p_i$ lies in the laser plane on the line $v$ between the closest seam points on each side on the laser plane, $p_1$ and $p_2$, refer to Fig. 14.4 for an illustration. Formally, the intersection point $p_i$ must satisfy the following two equations

$$\left. \begin{array}{r} p_i = p_1 + \gamma v \\ 0 = n^\mathsf{T}(p_i - \hat{m}) \end{array} \right\} \Rightarrow \gamma = \frac{n^\mathsf{T}(\hat{m} - p_1)}{n^\mathsf{T} v}$$

where *n* is the normal of the laser light plane.

The mean of $p(m|X)$ is thus equal to

$$\mu\{p(m|X)\} = p_i$$

and the shape should be chosen to reflect the error distribution of the laser sensor, here modeled as a normal distribution according to

$$p(m|X) = (2\pi)^{-\frac{3}{2}}|\Sigma|^{-\frac{1}{2}}\exp\left(-\frac{1}{2}e^{\mathsf{T}}\Sigma^{-1}e\right), \quad e = \hat{m} - p_i$$

Many seam-tracking sensors are capable of measuring also the angle of the weld surface around the normal of the laser plane. An angle measurement is easily compared to the corresponding angle hypothesis of a particle using standard roll, pitch, yaw calculations. Using the convention in Fig. 14.1, the angle around the normal of the laser plane corresponds to the yaw angle. Roll and pitch angles are unfortunately not directly measurable by this type of sensor. If, however, a sensor with two or more laser planes is used, it is possible to estimate the full orientation of the sensor. This will be analyzed further in Sec. 14.2.

***Reduction of computational time*** The evaluation of $p(m|X)$ can be computationally expensive due to the search procedure. We can reduce this cost by reducing the number of points to search over. This can be achieved by approximating the trajectory with a piecewise affine function. Since the intersection point between the nominal seam line and the laser light plane is calculated, this does not affect the accuracy of the evaluation of $p(m|X)$ much. To this end, we solve the following convex optimization problem

$$\begin{aligned}
\underset{z,w}{\text{minimize}} \quad & \|y - z\|_F^2 + \lambda \sum_{t=1}^{N-2}\sum_{j=1}^{3}|w_{t,j}| \\
\text{subject to} \quad & \|y - z\|_\infty \le \epsilon \\
& w_{t,j} = z_{t,j} - 2z_{t+1,j} + z_{t+2,j}
\end{aligned} \tag{14.8}$$

where $y \in \mathbb{R}^{N \times 3}$ are the positions of the nominal trajectory points, $z$ is the approximation of $y$, and $\epsilon$ is the maximum allowed approximation error. The nonzero elements of $w$ will determine the location of the knots in the piecewise affine approximation and $\lambda$ will influence the number of knots.[2]

The proposed optimization problem does not incorporate constraints on the orientation error of the approximation. This error will, however, be small if the trajectory is smooth with bounded curvature and a constraint is put on the error in the translational approximation, as in (14.8).

Optimization problem (14.8) can be seen as a multivariable trend-filtering problem, a topic which was discussed in greater detail in Sec. 6.4.

---

[2] $w_t = z_t - 2z_{t+1} + z_{t+2}$ is a discrete second-order differentiation of $z$.

## 14.2   Simulation Framework

The PF algorithm (Algorithm 1), detailed in the previous sections, has been implemented in an open-source framework, publicly available [*PFSeamTracking.jl*, B.C. et al., 2016]. The framework provides, apart from the state estimator, convenience methods for plotting, trajectory generation, optimization, simulation of laser-, joint-, and force sensor readings and perturbations due to process forces and kinematic model errors as well as tools for visualization of particle distributions.

### Visualization

An often time-consuming part during the implementation of a particle filtering framework is the tuning of the filter parameters. Due to the highly nonlinear nature of the present filtering problem, this is not as straightforward as in the Kalman-filtering scenario. A poorly tuned Kalman filter manifests itself as either too noisy, or too slow. A poorly tuned particle filter may, however, suffer from catastrophic failures such as mode collapse or particle degeneracy [Gustafsson, 2010; Thrun et al., 2005; Rawlings and Mayne, 2009].

To identify the presence of mode collapse or particle degeneracy and to assist in the tuning of the filter, we provide a visualization tool that displays the true trajectory as traversed by the robot together with the distribution of the particles, as well as each particle's hypothesis measurement location. An illustrative example is shown in Fig. 14.5, where one dimension in the filter state is shown as a function of time in a screen shot of the visualizer.

To further aid the tuning of the filter, we perform several simulations in parallel with nominal filter parameters perturbed by samples from a prespecified distribution and perform statistical significance tests to determine the parameters of most importance to the result for a certain sensor/trajectory configuration. Figure 14.6 displays the statistical significance of various filter parameters for a certain trajectory and sensor configuration. The color coding indicates the log(P)-values for the corresponding parameters in a linear model predicting the errors in Cartesian directions and orientation. As an example, the figure indicates that the parameter $\sigma_{W2}$, corresponding to the orientation noise in the state update, has a significant impact on the errors in all Cartesian directions. The sign and value of the underlying linear model can then serve as a guide to fine tuning of this parameter.

## 14.3   Analysis of Sensor Configurations

One of the main goals of this work was to enable analysis of the optimal sensor configuration for a given seam geometry. On the one hand, not all seam geometries allow for accurate estimation of the state in all directions, and on the other hand, not all seam geometries require accurate control in all directions. The optimal sensor configuration depends heavily on the amount of features present in the trajectories, where a feature is understood as a localizable detail in the trajectory. The estimation performance is also critically dependent on the number of laser light planes that intersect the seam. A single laser sensor can measure three degrees of freedom, two translations and one orientation. The remaining three DOFs are in general not observable. This is illustrated in the planar case in Fig. 14.7. All particles lying on a capsule manifold, generated by the spherical movement around the measurement location, together with a sliding motion along the seam, are equally likely given the measurement. A second measurement eliminates the spherical component of the capsule, leaving only the line corresponding to the sliding motion along the seam unobservable. The unobservable subspace left when two or more laser planes are used can only be reduced by features in the seam, breaking the line symmetry. One example of a reduction in uncertainty due to a feature in the trajectory is illustrated in Fig. 14.5. The forward kinematics measurement from the robot will, however, ensure that the uncertainty stays bounded within a region of the true state.

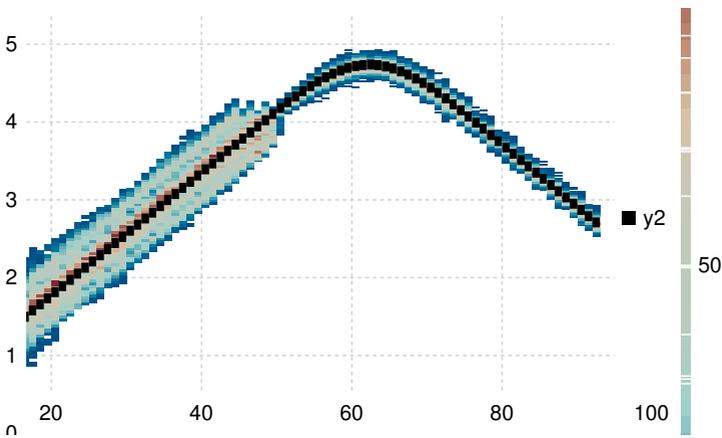

**Figure 14.5** Visualization of a particle distribution as a function of time during a simulation. The black line indicates the evolution of one coordinate of the true state as a function of the time step and the heatmap illustrates the density of the particles. This figure illustrates how the uncertainty of the estimate is reduced as a feature in the trajectory becomes visible for the sensor at time step 50. The sensor is located slightly ahead of the tool, hence, the distribution becomes narrow slightly before the tool reaches the feature. The feature is in this case a sharp bend in the otherwise straight seam.

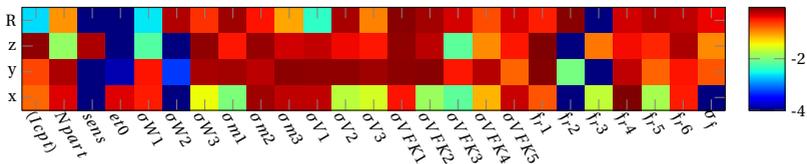

**Figure 14.6** An illustration of how the various parameters in the software framework can be tuned. By fitting linear models, with tuning parameters as factors, that predict various errors as linear combinations of parameter values, parameters with significant effect on the performance can be identified using the log(P)-values (color coded). The $x$-axis indicates the factors and the $y$-axis indicates the predicted errors in orientation and translation. The parameters are described in detail in the software framework.

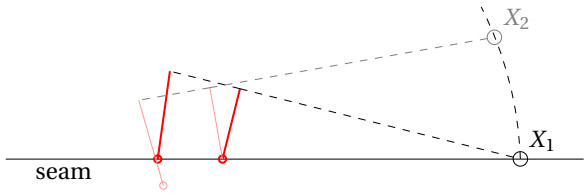

**Figure 14.7** A sensor with a single laser stripe is not capable of distinguishing between wrong translation and wrong orientation. The two hypotheses $X_1$, $X_2$ both share the closest measurement point on the seam. The second laser stripe invalidates the erroneous hypothesis $X_2$ which would have the second measurement point far from the seam. Without the second laser stripe it is clear that the available sensor information can not distinguish $X_1$ and $X_2$ from each other.

To illustrate the importance of seam geometry for the estimation performance, we consider Fig. 14.8, where the resulting errors for two trajectory types and several sensor configurations (0,1,2 sensors) are displayed. The trajectories referred to in the figure are generated as follows. The $xy$-trajectory lies entirely in the $xy$-plane of the tool frame $\mathcal{T}$, with a linear movement of 200 mm along the $y$-direction and a smooth, 20 mm amplitude, triangle-wave motion in the $x$-direction. The $yz$-trajectory lies in the $yz$-plane, with a linear movement of 200 mm along the $y$-direction and a 100 mm amplitude, sinusoidal, motion along the $z$-direction. The trajectories are depicted in Fig. 14.9. It is clear that the type of trajectory is important for the resulting estimation error, in this case, the filter was tuned for trajectory type $xy$.

Figure 14.8 illustrates the difficulties in determining the translation along the direction of movement when no features are present, as well as the benefit of sensor feedback in the measurable dimensions. The provided visualization tools assist in re-tuning the filter for a new trajectory, and can suggest optimal configurations of the available seam-tracking sensors.

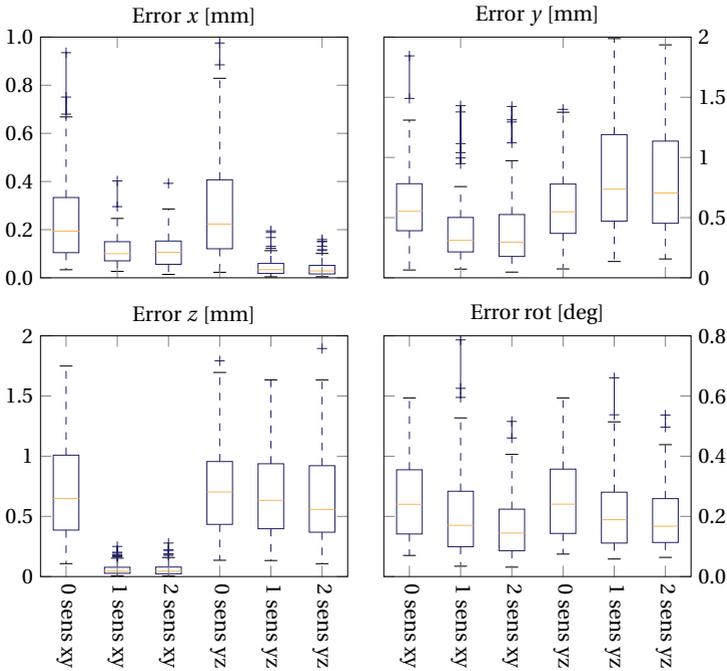

**Figure 14.8** Error distributions for various sensor configurations (0-2 sensors) and two different trajectory types (xy,yz). In both trajectory cases, *y* is the major movement direction along which the laser sensors obtain little or no information. The same filter parameters, tuned for the *xy*-trajectory, were used in all experiments.

## 14.4 Discussion

We have discussed several sources of kinematic uncertainties. To improve the performance of a state estimator, one can consider two approaches. Reducing uncertainties through modeling, or by introduction of additional sensing. Along the first avenue, we note that the kinematic model of the robot used in the forward kinematics calculations is often inaccurate, and errors in the absolute positioning accuracy of an industrial robot can often be in the order of 1 mm or more [Mooring et al., 1991; Nubiola and Bonev, 2013], even after additional, costly calibration performed by the robot manufacturer [ABB, 2005]. To characterize this uncertainty without performing a full kinematic calibration is usually hard, since it is a nonlinear function of the errors in link lengths, offsets etc. in the kinematic model. Possibilities include modeling this uncertainty as a Gaussian distribution with a variance corresponding to the average error in the considered work-space volume, or as a uniform distribution with a width corresponding to the maximum error. A number for the maximum error in the forward kinematics under no load is usually provided by the robot manufacturer, or can be obtained using, e.g., an external optical tracking system.

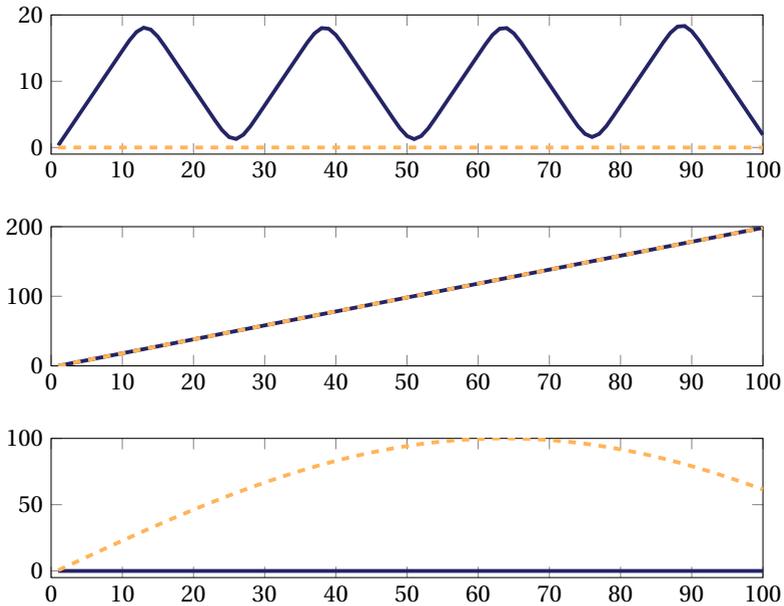

**Figure 14.9** Trajectories $xy$ (solid) and $yz$ (dashed). Distance [mm] along each axis $(x, y, z)$ is depicted as a function of time step.

A major source of uncertainty is compliance in the structure of the robot. Deflections in the robot joints and links caused by large process forces result in an uncertainty in the measured tool position. This problem can be mitigated by a compliance model, $C_j(\tau)$ in (14.3), reducing the uncertainty to the level of the model uncertainty [Lehmann et al., 2013]. Although several authors have considered compliance modeling, the large range of possible seam geometries and the large range of possible process forces make finding a globally valid, sufficiently accurate compliance model very difficult.

The reduction of uncertainty through additional sensing offers the possibility of reducing the remaining errors greatly, potentially eliminating the need for a state estimator altogether. Sensors capable of measuring the full 6DOF pose, such as optical tracking systems, are unfortunately very expensive. They further require accurate measurements also of the workpiece, potentially placing additional burden on the operator. Relative sensing, such as the laser sensors considered in this work, directly measure the relevant distance between the seam and the tool. Unfortunately, they suffer from a number of weaknesses. They can for obvious reasons not measure the location of the seam at the tool center point, and must thus measure the seam at a different location where it is visible. In a practical scenario, this might cause the sensor to measure the seam up to 50 mm from the TCP. Since the location of the TCP relative to the seam must be inferred through geometry from this sensor measurement, the full 6 DOF pose becomes relevant, even if does not have to be accurately controlled. A second weakness of the considered relative sensing is the lack of observability along the seam direction. While dual laser sensors allow measuring more degrees of freedom than a single laser sensor, no amount of laser sensors can infer the position along a straight seam.

The proposed state estimator tries to infer as much as possible about the state by requiring knowledge of the seam geometry. This is necessary for the estimator to know what sensor measurements to expect. The particle filter maintains a representation of the full filtering density, it is thus possible to determine in simulation whether or not the uncertainty is such that the worst-case tracking performance is sufficient. The uncertainty is in general highest along the direction of movement since no sensor information is available in this direction. Fortunately, however, this direction is also the direction with lowest required tracking accuracy. Situations that require higher tracking accuracy in this direction luckily coincide with the situations that allow for higher estimation accuracy, when a feature is present in the seam. An example of this was demonstrated in Fig. 14.5.

While the proposed framework is intended for simulation in order to aid the design of a specialized state estimator, some measures were taken to reduce the computational time and at the same time reducing the number of parameters the operator has to tune. The most notable such measure was the choice to not include velocities in the state. The velocities typically present in the FSW context are fairly low, while forces are high. The acceleration in the transverse direction can thus be high enough to render the estimation of velocities impossible on the short time-scale associated with vibrations in the process. The bandwidth of the controller is further far from enough for compensation to be feasible. In the directions along the seam, the velocity is typically well controlled by the robot controller apart from during the transient occurring when contact is established. Once again, the bandwidth is not sufficient to compensate for errors occurring at the frequencies present during the transient.

Lastly, the method does not include estimation of errors in the location of the work piece. Without assumptions on either the error in the work-piece location or the error in the forward kinematics of the robot, these two sources of error can not be distinguished. Hence, augmenting the state with a representation of the work-piece error will not be fruitful. If significant variation in work-piece placement is suspected we instead propose to add a scanning phase prior to welding. This would allow for using the laser sensor to, under no load, measure the location of sufficiently many points along the seam to be able to estimate the location of the work piece in the coordinate system of the robot. This procedure, which could be easily automated, would compensate for errors in both work-piece placement and the kinematic chain of the robot.

## 14.5 Conclusions

We have suggested a particle-filter based state estimator capable of estimating the full 6 DOF pose of the tool relative to the seam in a seam-tracking scenario. Sensor fusion is carried out between the robot internal measurements, propagated through a forward kinematics model with large uncertainties due to the applied process forces, and measurements from a class of seam-tracking laser sensors. We have highlighted some of the difficulties related to state estimation where accurate measurements come in a reduced-dimensional space, together with highly uncertain measurements of the full state space, where the uncertainties are highly correlated in time.

The presented framework is available as open-source [*PFSeamTracking.jl*, B.C. et al., 2016] and the algorithm has been successfully implemented at The Welding Institute (TWI) in Sheffield, UK, and is capable of executing in approximately 1000 Hz using 500 particles on a standard desktop PC.

# Conclusions and Future Work

We have presented a wide range of problems and methods within estimation for physical systems. Common to many parts of the thesis is the use of ideas from machine learning to solve classical identification problems. Many of the problems considered could, in theory, be solved by gathering massive amounts of data and training a deep neural network. Instead, we have opted for developing methods that make use of prior knowledge where available, and flexibility where not. This has resulted in practical methods that require a practical amount of data. The proof of this has in many cases been provided by experimental application on physical systems, the very systems that motivated the work.

Looking forward, we see robust uncertainty quantification as a very interesting and important direction for future work. Some of the developed methods have a probabilistic interpretation and lend themselves well to maximum a posteriori inference, in restricted settings. Lifting this restriction is straightforward in theory but most often computationally challenging. Work on approximate methods has recently made great strides in the area, but the field requires further attention before its application as a robust technology.

# Bibliography

ABB (2005). *Absolute accuracy: industrial robot option.* URL: `https://library.e.abb.com/public/0f879113235a0e1dc1257b130056d133/Absolute%20Accuracy%20EN_R4%20US%2002_05.pdf` (visited on 2018-11-22).

Abele, E., S. Rothenbücher, and M. Weigold (2008). "Cartesian compliance model for industrial robots using virtual joints". *Production Engineering* **2**:3, pp. 339–343.

Amos, B. and J. Z. Kolter (2017). *Optnet: differentiable optimization as a layer in neural networks.* eprint: `arXiv:1703.00443`.

Armstrong, B. (1988). "Friction: experimental determination, modeling and compensation". In: *Robotics and Automation, Proc. 1988 IEEE Int. Conf. Pennsylvania*, pp. 1422–1427.

Armstrong, B., P. Dupont, and C. C. De Wit (1994). "A survey of models, analysis tools and compensation methods for the control of machines with friction". *Automatica* **30**:7, pp. 1083–1138.

Åström, K. and B. Wittenmark (2013a). *Computer-Controlled Systems: Theory and Design, Third Edition.* Dover Publications, Minola, NY. ISBN: 9780486284040.

Åström, K. J. (2012). *Introduction to stochastic control theory.* Courier Corporation, New York.

Åström, K. J. and B. Wittenmark (2013b). *Adaptive control.* Courier Corporation, New York.

Åström, K. J. and R. M. Murray (2010). *Feedback systems: an introduction for scientists and engineers.* Princeton University Press, New Jersey.

Åström, K. J. and B. Wittenmark (2011). *Computer-controlled systems: theory and design.* Dover, New York.

Bagge Carlson, F. (2015). *Robotlib.jl.* Dept. Automatic Control, Lund University, Sweden. URL: `https://github.com/baggepinnen/Robotlib.jl`.

Bagge Carlson, F. (2016a). *BasisFunctionExpansions.jl.* Dept. Automatic Control, Lund University, Sweden. URL: `https://github.com/baggepinnen/BasisFunctionExpansions.jl`.

Bagge Carlson, F. (2016b). *DifferentialDynamicProgramming.jl*. Dept. Automatic Control, Lund University, Sweden. URL: https : / / github . com / baggepinnen/DifferentialDynamicProgramming.jl.

Bagge Carlson, F. (2016c). *LPVSpectral.jl*. Dept. Automatic Control, Lund University, Sweden. URL: https://github.com/baggepinnen/LPVSpectral.jl.

Bagge Carlson, F. (2017). *LTVModels.jl*. Dept. Automatic Control, Lund University, Sweden. URL: https://github.com/baggepinnen/LTVModels.jl.

Bagge Carlson, F. (2018a). *JacProp.jl*. Dept. Automatic Control, Lund University, Sweden. URL: https://github.com/baggepinnen/JacProp.jl.

Bagge Carlson, F. (2018b). *LowLevelParticleFilters.jl*. Dept. Automatic Control, Lund University, Sweden. URL: https : / / github . com / baggepinnen / LowLevelParticleFilters.jl.

Bagge Carlson, F. (2018c). *TotalLeastSquares.jl*. Dept. Automatic Control, Lund University, Sweden. URL: https : / / github . com / baggepinnen / TotalLeastSquares.jl.

Bagge Carlson, F. and M. Haage (2017). *YuMi low-level motion guidance using the Julia programming language and Externally Guided Motion Research Interface*. Technical report TFRT-7651. Department of Automatic Control, Lund University, Sweden.

Bagge Carlson, F., R. Johansson, and A. Robertsson (2015a). "Six DOF eye-to-hand calibration from 2D measurements using planar constraints". In: *Int. Conf. Intelligent Robots and Systems (IROS), Hamburg*. IEEE.

Bagge Carlson, F., R. Johansson, and A. Robertsson (2018a). "Tangent-space regularization for neural-network models of dynamical systems". *arXiv preprint arXiv:1806.09919*.

Bagge Carlson, F. and M. Karlsson (2016a). *DynamicMovementPrimitives.jl*. Dept. Automatic Control, Lund University, Sweden. URL: https://github.com/ baggepinnen/DynamicMovementPrimitives.jl.

Bagge Carlson, F. and M. Karlsson (2016b). *PFSeamTracking.jl*. Dept. Automatic Control, Lund University, Sweden. URL: https : / / github . com / baggepinnen/PFSeamTracking.jl.

Bagge Carlson, F., M. Karlsson, A. Robertsson, and R. Johansson (2016). "Particle filter framework for 6D seam tracking under large external forces using 2D laser sensors". In: *Int. Conf. Intelligent Robots and Systems (IROS), Daejeong, South Korea*.

Bagge Carlson, F., A. Robertsson, and R. Johansson (2015b). "Modeling and identification of position and temperature dependent friction phenomena without temperature sensing". In: *Int. Conf. Intelligent Robots and Systems (IROS), Hamburg*. IEEE.

Bagge Carlson, F., A. Robertsson, and R. Johansson (2017). "Linear parameter-varying spectral decomposition". In: *2017 American Control Conf (ACC), Seattle*.

Bagge Carlson, F., A. Robertsson, and R. Johansson (2018b). "Identification of LTV dynamical models with smooth or discontinuous time evolution by means of convex optimization". In: *IEEE Int. Conf. Control and Automation (ICCA), Anchorage, AK*.

Bao, Y., L. Tang, and D. Shah (2017). "Robotic 3d plant perception and leaf probing with collision-free motion planning for automated indoor plant phenotyping". In: *2017 ASABE Annual International Meeting*. American Society of Agricultural and Biological Engineers, p. 1.

Bellman, R. (1953). *An Introduction to the Theory of Dynamic Programming*. R-245. Rand Corporation, Santa Monica, CA.

Bellman, R. (1961). "On the approximation of curves by line segments using dynamic programming". *Communications of the ACM* **4**:6, p. 284.

Bellman, R. and R. Roth (1969). "Curve fitting by segmented straight lines". *Journal of the American Statistical Association* **64**:327, pp. 1079–1084.

Bennett, D. J., J. M. Hollerbach, and P. D. Henri (1992). "Kinematic calibration by direct estimation of the jacobian matrix". In: *Proceedings. IEEE International Conference on Robotics and Automation, Nice*. Pp. 351–357.

Bertsekas, D. P., D. P. Bertsekas, D. P. Bertsekas, and D. P. Bertsekas (2005). *Dynamic programming and optimal control*. Vol. 1. 3. Athena scientific Belmont, MA.

Bezanson, J., A. Edelman, S. Karpinski, and V. B. Shah (2017). "Julia: a fresh approach to numerical computing". *SIAM Review* **59**:1, pp. 65–98.

Bishop, C. M. (2006). *Pattern Recognition and Machine Learning*. Springer, New York.

Bittencourt, A. C. and S. Gunnarsson (2012). "Static friction in a robot joint - modeling and identification of load and temperature effects". *Journal of Dynamic Systems, Measurement, and Control* **134**:5.

Botev, A., H. Ritter, and D. Barber (2017). *Practical Gauss-Newton optimisation for deep learning*. eprint: arXiv:1706.03662.

Boyd, S. and L. Vandenberghe (2004). *Convex optimization*. Cambridge University Press, Cambridge, UK.

Bristow, D. A., M. Tharayil, and A. G. Alleyne (2006). "A survey of iterative learning control". *IEEE Control Systems* **26**:3, pp. 96–114.

Bugmann, G. (1998). "Normalized gaussian radial basis function networks". *Neurocomputing* **20**:1-3, pp. 97–110. ISSN: 0925-2312.

Chalus, M. and J. Liska (2018). "Calibration and using a laser profile scanner for 3d robotic welding". *International Journal of Computational Vision and Robotics* **8**:4, pp. 351–374.

Chen, D., A. Song, and A. Li (2015). "Design and calibration of a six-axis force/-torque sensor with large measurement range used for the space manipulator". *Procedia Engineering* **99**:1, pp. 1164–1170.

Costa, O. L. V., M. D. Fragoso, and R. P. Marques (2006). *Discrete-time Markov jump linear systems*. Springer Science & Business Media, London.

Dahl, P. (1968). *A solid friction model*. Tech. rep. TOR-0158 (3107-18)-1. Aerospace Corp, El Segundo, CA.

Daniilidis, K. (1999). "Hand-eye calibration using dual quaternions". *The International Journal of Robotics Research* **18**:3, pp. 286–298.

De Backer, J. (2014). *Feedback Control of Robotic Friction Stir Welding*. PhD thesis. ISBN 978-91-87531-00-2, University West, Trollhättan, Sweden.

De Backer, J. and G. Bolmsjö (2014). "Deflection model for robotic friction stir welding". *Industrial Robot: An International Journal* **41**:4, pp. 365–372.

De Backer, J., A.-K. Christiansson, J. Oqueka, and G. Bolmsjö (2012). "Investigation of path compensation methods for robotic friction stir welding". *Industrial Robot: An International Journal* **39**:6, pp. 601–608.

De Wit, C. C., H. Olsson, K. J. Åström, and P. Lischinsky (1995). "A new model for control of systems with friction". *Automatic Control, IEEE Trans. on* **40**:3, pp. 419–425.

Eckart, C. and G. Young (1936). "The approximation of one matrix by another of lower rank". *Psychometrika* **1**:3, pp. 211–218.

Eggert, D. W., A. Lorusso, and R. B. Fisher (1997). "Estimating 3-d rigid body transformations: a comparison of four major algorithms". *Machine Vision and Applications* **9**:5-6, pp. 272–290.

Evensen, G. (2003). "The ensemble Kalman filter: theoretical formulation and practical implementation". *Ocean Dynamics* **53**:4, pp. 343–367. ISSN: 1616-7341. DOI: 10.1007/s10236-003-0036-9.

Fang, X. (2013). "Weighted total least squares: necessary and sufficient conditions, fixed and random parameters". *Journal of geodesy* **87**:8, pp. 733–749.

Fischler, M. A. and R. C. Bolles (1981). "Random sample consensus: a paradigm for model fitting with applications to image analysis and automated cartography". *Communications of the ACM* **24**:6, pp. 381–395.

Gao, X., D. You, and S. Katayama (2012). "Seam tracking monitoring based on adaptive kalman filter embedded elman neural network during high-power fiber laser welding". *Industrial Electronics, IEEE Transactions on* **59**:11, pp. 4315–4325.

Gershman, S. J. and D. M. Blei (2011). *A tutorial on Bayesian nonparametric models*. eprint: arXiv:1106.2697.

Glad, T. and L. Ljung (2014). *Control theory*. CRC press, Boca Raton, Florida.

Goldstein, A. A. (1964). "Convex programming in hilbert space". *Bulletin of the American Mathematical Society* **70**:5, pp. 709–710.

Golub, G. H. and C. F. Van Loan (2012). *Matrix computations*. Vol. 3. Johns Hopkins University Press, Baltimore.

Goodfellow, I., Y. Bengio, and A. Courville (2016). *Deep Learning*. http://www.deeplearningbook.org. MIT Press, Cambridge MA.

Guillo, M. and L. Dubourg (2016). "Impact & improvement of tool deviation in friction stir welding: weld quality & real-time compensation on an industrial robot". *Robotics and Computer-Integrated Manufacturing* **39**, pp. 22–31.

Gustafsson, F. (2010). "Particle filter theory and practice with positioning applications". *Aerospace and Electronic Systems Magazine, IEEE* **25**:7, pp. 53–82.

Hansen, P. C. (1994). "Regularization tools: a matlab package for analysis and solution of discrete ill-posed problems". *Numerical algorithms* **6**:1, pp. 1–35. URL: http://www2.compute.dtu.dk/~pcha/Regutools/RTv4manual.pdf (visited on 2017-01).

Harris, F. J. (1978). "On the use of windows for harmonic analysis with the discrete fourier transform". *Proceedings of the IEEE* **66**:1, pp. 51–83.

He, K., X. Zhang, S. Ren, and J. Sun (2015). *Deep residual learning for image recognition*. eprint: arXiv:1512.03385.

He, K., X. Zhang, S. Ren, and J. Sun (2016). "Deep residual learning for image recognition". In: *IEEE Conf. on Computer Vision and Pattern Recognition, Las Vegas*, pp. 770–778.

Hjort, N. L., C. Holmes, P. Müller, and S. G. Walker (2010). *Bayesian nonparametrics*. Vol. 28. Cambridge University Press, Cambridge, UK.

Hochreiter, S. and J. Schmidhuber (1997). "Long short-term memory". *Neural computation* **9**:8, pp. 1735–1780.

Horaud, R. and F. Dornaika (1995). "Hand-eye calibration". *The International Journal of Robotics Research* **14**:3, pp. 195–210.

Huang, P.-Y., Y.-Y. Chen, and M.-S. Chen (1998). "Position-dependent friction compensation for ballscrew tables". In: *Control Applications, 1998. Proc. 1998 IEEE Int. Conf., Trieste, Italy*. Vol. 2, pp. 863–867.

Ikits, M. and J. Hollerbach (1997). "Kinematic calibration using a plane constraint". In: *Robotics and Automation, 1997. Proceedings., 1997 IEEE International Conference on, Pittsburgh*. Vol. 4, 3191–3196 vol.4. DOI: 10.1109/ROBOT.1997.606774.

Innes, M. (2018). "Flux: elegant machine learning with julia". *Journal of Open Source Software*. DOI: 10.21105/joss.00602.

Ioffe, S. and C. Szegedy (2015). "Batch normalization: accelerating deep network training by reducing internal covariate shift". In: *International Conference on Machine Learning, Lille*, pp. 448–456.

Ionides, E. L., C. Bretó, and A. A. King (2006). "Inference for nonlinear dynamical systems". *Proceedings of the National Academy of Sciences* **103**:49, pp. 18438–18443.

Johansson, R. (1993). *System modeling & identification*. Prentice-Hall, Englewood Cliffs, NJ.

Julialang (2017). *Julia standard library*. URL: http://docs.julialang.org/en/stable/stdlib/linalg/ (visited on 2017-01).

Karl, M., M. Soelch, J. Bayer, and P. van der Smagt (2016). *Deep variational Bayes filters: unsupervised learning of state space models from raw data*. eprint: arXiv: 1605.06432.

Karlsson, M., F. Bagge Carlson, A. Robertsson, and R. Johansson (2017). "Two-degree-of-freedom control for trajectory tracking and perturbation recovery during execution of dynamical movement primitives". In: *20th IFAC World Congress, Toulouse*.

Kay, S. M. (1993). *Fundamentals of statistical signal processing, volume I: estimation theory*. Prentice Hall, Englewood Cliffs, NJ.

Khalil, H. K. (1996). "Nonlinear systems". *Prentice-Hall, New Jersey* **2**:5.

Kim, S.-J., K. Koh, S. Boyd, and D. Gorinevsky (2009). "$\ell_1$ trend filtering". *SIAM review* **51**:2, pp. 339–360.

Kingma, D. and J. Ba (2014). "Adam: a method for stochastic optimization". *arXiv preprint arXiv:1412.6980*.

Kruif, B. J. de and T. J. de Vries (2002). "Support-vector-based least squares for learning non-linear dynamics". In: *Decision and Control, 2002, Proc. IEEE Conf., Las Vegas*. Vol. 2, pp. 1343–1348.

Lehmann, C., B. Olofsson, K. Nilsson, M. Halbauer, M. Haage, A. Robertsson, O. Sörnmo, and U. Berger (2013). "Robot joint modeling and parameter identification using the clamping method". In: *7th IFAC Conference on Manufacturing Modelling, Management,and Control*. Saint Petersburg, Russia, pp. 843–848.

Lennartson, B., R. H. Middleton, and I. Gustafsson (2012). "Numerical sensitivity of linear matrix inequalities using shift and delta operators". *IEEE Transactions on Automatic Control* **57**:11, pp. 2874–2879.

Levine, S. and P. Abbeel (2014). "Learning neural network policies with guided policy search under unknown dynamics". In: *Advances in Neural Information Processing Systems, Montreal*, pp. 1071–1079.

Levine, S. and V. Koltun (2013). "Guided policy search". In: *Int. Conf. Machine Learning (ICML), Atlanta*, pp. 1–9.

Levine, S., N. Wagener, and P. Abbeel (2015). "Learning contact-rich manipulation skills with guided policy search". In: *Robotics and Automation (ICRA), IEEE Int. Conf., Seattle*. IEEE, pp. 156–163.

Lindström, E., E. Ionides, J. Frydendall, and H. Madsen (2012). "Efficient iterated filtering". *IFAC Proceedings Volumes* **45**:16, pp. 1785–1790.

Ljung, L. (1987). *System identification: theory for the user*. Prentice-hall, Englewood Cliffs, NJ.

Ljung, L. and T. Söderström (1983). *Theory and practice of recursive identification*. MIT press, Cambridge, MA.

Manchester, I. R., M. M. Tobenkin, and A. Megretski (2012). "Stable nonlinear system identification: convexity, model class, and consistency". *IFAC Proceedings Volumes* **45**:16, pp. 328–333. DOI: 10.3182/20120711-3-BE-2027.00405.

Mayne, D. (1966). "A second-order gradient method for determining optimal trajectories of non-linear discrete-time systems". *International Journal of Control* **3**:1, pp. 85–95.

Merriënboer, B. van, O. Breuleux, A. Bergeron, and P. Lamblin (2018). *Automatic differentiation in ML: where we are and where we should be going.* eprint: `arXiv:1810.11530`.

Meta Vision Systems (2014). *Meta FSW*. URL: `http://www.meta-mvs.com/fsw` (visited on 2014-01-12).

Middleton, R. and G. Goodwin (1986). "Improved finite word length characteristics in digital control using delta operators". *IEEE Transactions on Automatic Control* **31**:11, pp. 1015–1021.

Midling, O. T., E. J. Morley, and A. Sandvik (1998). *Friction stir welding*. US Patent 5,813,592.

Mnih, V., K. Kavukcuoglu, D. Silver, A. A. Rusu, J. Veness, M. G. Bellemare, A. Graves, M. Riedmiller, A. K. Fidjeland, G. Ostrovski, et al. (2015). "Human-level control through deep reinforcement learning". *Nature* **518**:7540, p. 529.

Mooring, B., Z. Roth, and M. Driels (1991). *Fundamentals of manipulator calibration.* J. Wiley, New York. ISBN: 9780471508649.

Mortari, D., F. L. Markley, and P. Singla (2007). "Optimal linear attitude estimator". *Journal of Guidance, Control, and Dynamics* **30**:6, pp. 1619–1627.

Murphy, K. P. (2012). *Machine learning: a probabilistic perspective*. MIT press, Cambridge, MA.

Murray, R. M., Z. Li, and S. S. Sastry (1994). *A mathematical introduction to robotic manipulation*. CRC Press, Boca Raton, Florida.

Nagarajaiah, S. and Z. Li (2004). "Time segmented least squares identification of base isolated buildings". *Soil Dynamics and Earthquake Engineering* **24**:8, pp. 577–586.

Nayak, N. R. and A. Ray (2013). *Intelligent seam tracking for robotic welding*. Springer Science & Business Media, London, UK.

Nguyen, Q., M. C. Mukkamala, and M. Hein (2018). *On the loss landscape of a class of deep neural networks with no bad local valleys.* eprint: `arXiv:1809.10749`.

Nocedal, J. and S. Wright (1999). *Numerical optimization*. Springer-Verlag New York, Inc.

Nubiola, A. and I. A. Bonev (2013). "Absolute calibration of an ABB IRB 1600 robot using a laser tracker". *Robotics and Computer-Integrated Manufacturing* **29**:1, pp. 236–245. ISSN: 0736-5845.

Oberman, A. M. and J. Calder (2018). *Lipschitz regularized deep neural networks converge and generalize.* eprint: `arXiv:1808.09540`.

Ohlsson, H. (2010). *Regularization for Sparseness and Smoothness: Applications in System Identification and Signal Processing*. PhD thesis 1351. Linköping University Electronic Press, Linköping, Sweden.

Olofsson, B. (2015). *Topics in Machining with Industrial Robot Manipulators and Optimal Motion Control*. PhD thesis. ISRN TFRT–1108–SE, Lund University, Lund, Sweden.

Olsson, H., K. J. Åström, C. C. de Wit, M. Gäfvert, and P. Lischinsky (1998). "Friction models and friction compensation". *European Journal of Control* **4**:3, pp. 176–195.

Parikh, N. and S. Boyd (2014). "Proximal algorithms". *Foundations and Trends in Optimization* **1**:3, pp. 127–239.

Paris, J. F. (2011). "A note on the sum of correlated gamma random variables". *CoRR* **abs/1103.0505**. URL: http://arxiv.org/abs/1103.0505.

Park, J. and I. W. Sandberg (1991). "Universal approximation using radial-basis-function networks". *Neural computation* **3**:2, pp. 246–257.

Pascanu, R., T. Mikolov, and Y. Bengio (2013). "On the difficulty of training recurrent neural networks". In: *International Conference on Machine Learning, Atlanta*, pp. 1310–1318.

Pearson, K. (1901). "On lines and planes of closest fit to systems of points in space". *The London, Edinburgh, and Dublin Philosophical Magazine and Journal of Science* **2**:11, pp. 559–572.

Picinbono, B. (1996). "Second-order complex random vectors and normal distributions". *IEEE Transactions on Signal Processing* **44**:10, pp. 2637–2640.

Puryear, C. I., O. N. Portniaguine, C. M. Cobos, and J. P. Castagna (2012). "Constrained least-squares spectral analysis: application to seismic data". *Geophysics* **77**:5, pp. V143–V167.

Ramachandran, P., B. Zoph, and Q. V. Le (2017). *Searching for activation functions*. eprint: arXiv:1710.05941.

Rasmussen, C. E. (2004). "Gaussian processes in machine learning". In: *Advanced lectures on machine learning*. Springer, New York, pp. 63–71.

Rauch, H. E., F. Tung, C. T. Striebel, et al. (1965). "Maximum likelihood estimates of linear dynamic systems". *AIAA journal* **3**:8, pp. 1445–1450.

Rawlings, J. and D. Mayne (2009). *Model Predictive Control: Theory and Design*. Nob Hill Pub. Madison, Wisconsin. ISBN: 9780975937709.

Rummery, G. A. and M. Niranjan (1994). *On-line Q-learning using connectionist systems*. Vol. 37. University of Cambridge, Department of Engineering, Cambridge, England.

Saad, Y. and M. H. Schultz (1986). "Gmres: a generalized minimal residual algorithm for solving nonsymmetric linear systems". *Journal on scientific and statistical computing* **7**:3, pp. 856–869.

Schulman, J., S. Levine, P. Abbeel, M. Jordan, and P. Moritz (2015). "Trust region policy optimization". In: *International Conference on Machine Learning, Lille*, pp. 1889–1897.

SICK IVP (2011). *SICK Ranger*. URL: http : / / www . chronos – vision . de / downloads/FAQ_Summary_3D_Camera_V1.13.pdf (visited on 2018-08-29).

Silver, D., A. Huang, C. J. Maddison, A. Guez, L. Sifre, G. Van Den Driessche, J. Schrittwieser, I. Antonoglou, V. Panneershelvam, M. Lanctot, et al. (2016). "Mastering the game of go with deep neural networks and tree search". *Nature* **529**:7587, p. 484.

Silver, D., G. Lever, N. Heess, T. Degris, D. Wierstra, and M. Riedmiller (2014). "Deterministic policy gradient algorithms". In: *Proceedings of the 31st International Conference on International Conference on Machine Learning (ICML) Beijing*. JMLR.

Sjöberg, J., Q. Zhang, L. Ljung, A. Benveniste, B. Delyon, P.-Y. Glorennec, H. Hjalmarsson, and A. Juditsky (1995). "Nonlinear black-box modeling in system identification: a unified overview". *Automatica* **31**:12, pp. 1691–1724.

Song, A., J. Wu, G. Qin, and W. Huang (2007). "A novel self-decoupled four degree-of-freedom wrist force/torque sensor". *Measurement* **40**:9-10, pp. 883–891.

Sörnmo, O. (2015). *Adaptation and Learning for Manipulators and Machining*. PhD thesis. ISRN TFRT–1110–SE, Lund University, Lund, Sweden.

Spong, M. W., S. Hutchinson, and M. Vidyasagar (2006). *Robot modeling and control*. Vol. 3. Wiley, New York.

Stella, L., N. Antonello, and M. Fält (2016). *ProximalOperators.jl*. URL: https://github.com/kul-forbes/ProximalOperators.jl.

Stoer, J. and R. Bulirsch (2013). *Introduction to numerical analysis*. Vol. 12. Springer Science & Business Media, New York.

Stoica, P. and R. L. Moses (2005). *Spectral analysis of signals*. Pearson/Prentice Hall Upper Saddle River, NJ.

Sutton, R. S. (1991). "Dyna, an integrated architecture for learning, planning, and reacting". *ACM SIGART Bulletin* **2**:4, pp. 160–163.

Sutton, R. S., D. A. McAllester, S. P. Singh, and Y. Mansour (2000). "Policy gradient methods for reinforcement learning with function approximation". In: *Advances in neural information processing systems, Denver*, pp. 1057–1063.

Tassa, Y., N. Mansard, and E. Todorov (2014). "Control-limited differential dynamic programming". In: *Robotics and Automation (ICRA), 2014 IEEE International Conference on, Hong Kong*. DOI: 10.1109/ICRA.2014.6907001.

Thrun, S., W. Burgard, and D. Fox (2005). *Probabilistic Robotics*. Intelligent robotics and autonomous agents. MIT Press, Cambridge, MA. ISBN: 9780262201629.

Tibshirani, R. J. et al. (2014). "Adaptive piecewise polynomial estimation via trend filtering". *The Annals of Statistics* **42**:1, pp. 285–323.

Todorov, E. and W. Li (2005). "A generalized iterative lqg method for locally-optimal feedback control of constrained nonlinear stochastic systems". In: *American Control Conference*. IEEE, pp. 300–306.

Tsai, R. Y. and R. K. Lenz (1989). "A new technique for fully autonomous and efficient 3d robotics hand/eye calibration". *Robotics and Automation, IEEE Transactions on* **5**:3, pp. 345–358.

Tsiotras, P., J. L. Junkins, and H. Schaub (1997). "Higher-order cayley transforms with applications to attitude representations". *Journal of Guidance, Control, and Dynamics* **20**:3, pp. 528–534.

Ulyanov, D., A. Vedaldi, and V. Lempitsky (2017). "Deep image prior". *arXiv preprint arXiv:1711.10925*.

Van Overschee, P. and B. De Moor (1995). "A unifying theorem for three subspace system identification algorithms". *Automatica* **31**:12, pp. 1853–1864.

Verhaegen, M. and P. Dewilde (1992). "Subspace model identification part 1. the output-error state-space model identification class of algorithms". *International Journal of Control* **56**:5, pp. 1187–1210. DOI: 10.1080/00207179208934363. URL: https://doi.org/10.1080/00207179208934363.

Vidal, R., A. Chiuso, and S. Soatto (2002). "Observability and identifiability of jump linear systems". In: *IEEE Conf. Decision and Control (CDC), Las Vegas*. Vol. 4. IEEE, pp. 3614–3619.

Watkins, C. J. and P. Dayan (1992). "Q-learning". *Machine learning* **8**:3-4, pp. 279–292.

Wells, D. E., P. Vanícek, and S. D. Pagiatakis (1985). *Least squares spectral analysis revisited*. 84. Department of Surveying Engineering, University of New Brunswick Fredericton, Canada.

Wilk, M. B. and R. Gnanadesikan (1968). "Probability plotting methods for the analysis of data". *Biometrika* **55**:1, pp. 1–17. DOI: 10.1093/biomet/55.1.1. URL: https://doi.org/10.1093/biomet/55.1.1.

Williams, R. (1988). *Toward a Theory of Reinforcement-learning Connectionist Systems*. B00072BHM6. Northeastern University, Boston, MA.

Wilson, A. C., R. Roelofs, M. Stern, N. Srebro, and B. Recht (2017). "The marginal value of adaptive gradient methods in machine learning". In: *Advances in Neural Information Processing Systems*, pp. 4148–4158.

Wittenmark, B., K. J. Åström, and K.-E. Årzén (2002). "Computer control: an overview". *IFAC Professional Brief*. URL: https://www.ifac-control.org/publications/list-of-professional-briefs/pb_wittenmark_etal_final.pdf/view.

Xu, K., J. Ba, R. Kiros, K. Cho, A. C. Courville, R. Salakhutdinov, R. S. Zemel, and Y. Bengio (2015). "Show, attend and tell: neural image caption generation with visual attention". *CoRR* **abs/1502.03044**. arXiv: 1502.03044. URL: http://arxiv.org/abs/1502.03044.

Yongsheng, W., W. Tianqi, L. Liangyu, L. Jinzhong, and D. Boyu (2017). "Automatic path planning technology of stitching robot for composite fabric with curved surface (translated from Chinese)". *Materials Science and Technology (translated from Chinese)* **25**:2, pp. 16–21.

Yuan, M. and Y. Lin (2006). "Model selection and estimation in regression with grouped variables". *Journal of the Royal Statistical Society: Series B (Statistical Methodology)* **68**:1, pp. 49–67.

Zhang, Q. and R. Pless (2004). "Extrinsic calibration of a camera and laser range finder (improves camera calibration)". In: *Intelligent Robots and Systems, 2004. (IROS 2004). Proceedings. 2004 IEEE/RSJ International Conference on, Sendai, Japan*. Vol. 3, 2301–2306 vol.3. DOI: 10.1109/IROS.2004.1389752.

Zhang, W. (1999). *State-space search: Algorithms, complexity, extensions, and applications*. Springer Science & Business Media, New York.

Zhuang, H., S. Motaghedi, and Z. S. Roth (1999). "Robot calibration with planar constraints". In: *Robotics and Automation, 1999. Proceedings. 1999 IEEE International Conference on, Detroit, Michigan*. Vol. 1, 805–810 vol.1. DOI: 10.1109/ROBOT.1999.770073.